\documentclass{article} 
\usepackage{iclr2025_conference,times}

\usepackage{amsmath,amsfonts,bm}

\def\eqref#1{equation~\ref{#1}}

\def\1{\bm{1}}

\DeclareMathAlphabet{\mathsfit}{\encodingdefault}{\sfdefault}{m}{sl}
\SetMathAlphabet{\mathsfit}{bold}{\encodingdefault}{\sfdefault}{bx}{n}

\usepackage{hyperref}
\usepackage{url}
\usepackage{amsmath}
\usepackage{graphicx}
\usepackage{booktabs}
\usepackage{array}
\usepackage{algorithm}
\usepackage{algpseudocode}
\usepackage{multirow}
\usepackage{enumitem}
\usepackage{tcolorbox}
\usepackage{listings}
\usepackage{xcolor}
\tcbuselibrary{skins, breakable}
\usepackage{mdframed}
\definecolor{promptbg}{RGB}{240, 240, 240}
\definecolor{task1}{RGB}{230, 245, 255}    
\definecolor{task2}{RGB}{230, 250, 230}    
\definecolor{task3}{RGB}{255, 240, 230}    
\definecolor{task4}{RGB}{240, 230, 255}    
\definecolor{task5}{RGB}{255, 245, 230}    
\usepackage{xcolor}
\usepackage{listings}

\title{C-Evolve: Consensus-based Evolution for Prompt Groups}

\author{
Tiancheng Li$^{\dagger}$, \quad
Yuhang Wang$^{\dagger}$, \quad
Zhiyang Chen$^{}$, \\ \hspace{0.1em}
\textbf{Zijun Wang}$^{}$, \quad
\textbf{Liyuan Ma}$^{}$, \quad
\textbf{Guo-Jun Qi}$^{*}$ \\
School of Engineering, Westlake University \& MAPLE Lab \\
\texttt{\small \{litiancheng, wangyuhang, chenzhiyang, wangzijun, maliyuan\}@westlake.edu.cn} \\
\texttt{\small guojunq@gmail.com}
}

\iclrfinalcopy
\begin{document}

\maketitle
\let\thefootnote\relax\footnotetext{This project was initiated and supported by MAPLE Lab at Westlake University.}
\let\thefootnote\relax\footnotetext{$^\dagger$ Equal contribution.}
\let\thefootnote\relax\footnotetext{$^*$ Corresponding author.}
\begin{abstract}
Prompt evolution algorithms offer a powerful paradigm for enhancing AI systems based on closed-source models, while few work explores whether aggregating results from multiple prompts to reach a  \emph{consensus} can further advance the system capability boundary. In this paper, we introduce \emph{Consensus-Evolve} (C-Evolve), an evolutionary algorithm that discovers a group of prompts whose aggregated outputs after majority voting achieve optimal performance.
More specifically, C-Evolve employs an island-based evolutionary algorithm to maintain population diversity, and prompts from distinct islands are selected to form groups to aggregate their outputs.
The key difference from single individual evolution is a voting score, which evaluates each individual prompt's contribution within groups. We take this as the fitness score for evolution instead of individual performance.
Consequently, C-Evolve is more likely to produce and maintain prompts with higher potential to form a high-performing group and eliminate low-performing ones, gradually improving the group performance after reaching consensus.
Our method achieves state-of-the-art performance across a wide range of tasks, including both open-ended tasks like HotpotQA and closed-ended tasks like MATH. On Qwen3-8B, C-Evolve achieves 70.67\% on HotpotQA and 43.88\% on IFBench, which are 4.95\% and 2.73\% higher than GEPA, respectively. For GPT-4.1-mini, the accuracy on IFBench is further improved to 47.96\% and reaches 95.33\% in the MATH benchmark. These results demonstrate the C-Evolve's competitive performance.
\end{abstract}

\section{Introduction}
The advancement of Large Language Models (LLMs) has significantly accelerated progress in natural language processing \citep{liu2024deepseek,thirunavukarasu2023large}. However, many state-of-the-art models, such as GPT-4.1 \citep{openai2025gpt41} and Claude \citep{anthropic2024claude35sonnet}, remain closed-source and are accessible only via API interfaces. This restricts researchers from performing parameter fine-tuning to adapt these models to specific downstream tasks. To leverage the capabilities of such black-box models, a new paradigm prompt-based optimization methods have emerged \citep{opsahl2024optimizing,camara2025moprompt,agarwal2025promptwizard,fernando2023promptbreeder}. By refining system prompts, these methods enhance model performance without updating internal weights.



Among these paradigms, evolutionary algorithms represent a promising direction \citep{agrawal2025gepa,novikov2025alphaevolve} without access to the weight gradients of black-box models. Indicated with a predefined fitness function, these methods iteratively refine a population of prompts and select the individual with the highest fitness as the globally optimal prompt.
However, in complex task environments, a single optimal prompt often suffers from inherent expressive limitations, struggling to fully capture multi-faceted task requirements. This shortcoming can be alleviated through cooperation among multiple different prompts --- while any single prompt may fail in certain cases, its weaknesses can be compensated by others.
By aggregating their responses, the system has a higher probability of producing the correct answer. This could also be motivated by the findings in traditional machine learning, where ensemble methods improve performance compared to individual classifiers \citep{khan2024review,sun2024improved,webb2004multistrategy}. Despite this insight, existing work in this direction largely overlooks the systematic exploration of the mechanism to aggregate results from multiple prompts, limiting the performance of LLM-based systems in complex scenarios.

To release the potential of consensus from multiple prompts, we propose \emph{Consensus-Evolve (C-Evolve)}. Instead of evolving a single prompt from populations, the proposed method evolves a group of prompts that collectively achieve better performance after reaching consensus.
These prompts contribute by first generating outputs individually, and then their outputs are aggregated to form the final answer.
To evolve the prompts for this consensus target, we introduce a ``voting score" to evaluate the contribution of each prompt within groups, which is calculated by averaging the performance of all the groups in which it participates, as well as being moving averaged over iterations to adapt to the evolving populations that form the groups. 
This mechanism encourages the selection and reproduction of individuals more likely to form a high-performing group, while eliminating under-performing ones.
To maintain population diversity, C-Evolve adopts an island-based evolutionary algorithm \citep{izzo2012generalized,romera2024mathematical}, where prompts on different islands evolve mostly in isolation. Then we sample a prompt from each island to form a voting group where prompts could complement each other. Once evolution is completed, the top-scored prompts from each island are assembled to form a final group to predict their consensus on the test set. 


The proposed method achieves the state-of-the-art performance across a wide range of tasks, including both open-ended tasks like HotpotQA~\cite{yang2018hotpotqadatasetdiverseexplainable} and IFBench~\cite{pyatkin2025generalizingverifiableinstructionfollowing}, and closed-ended tasks like Hover~\cite{he2025hoverversatileneuralwholebody}, MATH~\cite{lightman2023lets} and GPQA~\cite{rein2023gpqagraduatelevelgoogleproofqa}, on both open-source (Qwen3-8B~\cite{yang2025qwen3technicalreport}) and closed-source (GPT-4.1-mini~\cite{openai2025gpt41}) models.
On Qwen3-8B, C-Evolve achieves 43.88\% on IFBench and 70.67\% on HotpotQA, 4.95\% and 2.73\% higher than GEPA respectively. On GPT-4.1-mini, the accuracy on IFBench is further improved to 47.96\% and reaches 95.33\% on MATH Benchmark. These results demonstrate the strong performance of C-Evolve.



\begin{figure}
    \centering
    \includegraphics[width=1\linewidth]{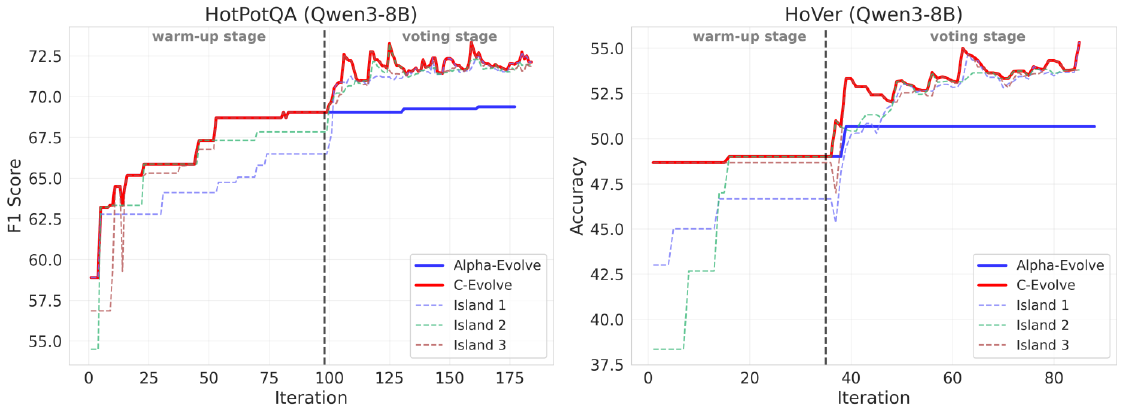}
    \vspace{-5mm}
    \caption{Comparison of our proposed C-Evolve and AlphaEvolve. The blue line shows AlphaEvolve's highest individual score across three islands on the metric set. In C-Evolve, the red line represents the highest individual score (with dashed lines indicating the islands' highest individual scores) during the warm-up stage and the highest EMA voting score (with dashed lines indicating the islands' highest EMA voting scores) during the voting stage.}
    \label{fig:experiment_score_per_island}
    \vspace{-3mm}
\end{figure}
\section{Related Work}
\textbf{Prompt Learning.}
The advent of powerful, closed-source large language models such as GPT-4.1 and Claude has established a new paradigm in GenAI. As these models are often accessible only through APIs, directly fine-tuning network weights is not feasible, making prompt engineering critical for adapting them to downstream tasks.

Early efforts relied on manual prompt engineering~\citep{wei2022chain, chen2025unleashing} such as chain-of-thought prompting~\citep{wei2022chain} , which suffered from inefficiency, strong subjectivity, and poor transferability. 
To address these challenges, automated prompt optimization techniques~\citep{zhou2022large, kepel2024autonomous} have been developed. 
Search-based prompt optimization methods~\citep{opsahl2024optimizing} utilize a Bayesian optimization framework to explore the joint space of instructions and examples, identifying the prompt configuration that maximizes validation performance.
Another paradigm explores how models can learn online from interactive feedback. 
For example, previous work~\citep{monea2024llms} demonstrates that LLMs can learn online from reward signals provided directly within their context, without requiring parameter updates.
In this setting, the LLM's context is built from a history of its own predictions and the resulting feedback, allowing it to improve its capabilities over time.

Beyond manual approaches, evolutionary prompt optimization has emerged as a particularly powerful and flexible paradigm. 
This approach treats prompts as individuals in a population that can be improved through iterative cycles of mutation and selection. 
In the context of LLMs, this paradigm has been successfully applied to solve complex tasks. 
For example, GEPA~\citep{agrawal2025gepa} employs a genetic algorithm that incorporates natural language reflection and pareto-based selection to evolve prompts.
 AlphaEvolve~\citep{novikov2025alphaevolve} utilizes an island-based evolutionary algorithm to evolve entire programs for scientific discovery, demonstrating the power of evolution with complex code structures.
Although they use different forms of evolutionary algorithms and operate on different targets (coding versus text prompts), both are fundamentally designed to discover a single best-performing solution through evolution.
This strategy may limit generalization -- when task complexity increases or task dynamics changes, it could fail to explore the synergy among different prompts that complement each other. 
The advantage of reaching consensus by aggregating outputs from multiple prompts remains underexplored in the field of prompt optimization.

\textbf{Consensus in LLMs.}
The idea that a committee of experts can outperform single expert is a cornerstone of many traditional machine learning algorithms, witnessed by decades of research on ensemble methods~\citep{rokach2019ensemble,polikar2012ensemble}. 
This principle has been applied to LLMs through techniques like self-consistency~\citep{wang2022self}, where multiple outputs are generated from multiple prompts and final answer is chosen by majority voting.
Another approach to aggregate outputs from multiple experts is used in multi-agent systems~\citep{wang2024mixture,tian2024optimizing,grotschla2025agentsnet}, where multiple agents with potentially distinct roles and capabilities interact with each other to solve complex problems.
Our method advances this idea beyond these paradigms. 
C-Evolve evolves a group of individual prompts where consensus mechanism is an intrinsic force driving the evolution process. 
This approach systematically evolves a diverse group of prompts for building a high-quality consensus, ensuring the resulting prompts are not only individually strong, but also cooperatively effective in groups.

\section{Problem Statement}
\textbf{Compound AI System.} Following established research, we define a Compound AI System as any modular system composed of one or multiple LLMs invocations, potentially interleaved with external tool calls, and orchestrated through arbitrary control flows. Such a system can be formalized as $\Phi = (M, L, \mathcal{X}, \mathcal{Y})$, where $M = \left\langle M_{1}, \ldots, M_{|M|} \right\rangle $ denotes the set of language modules, $L$ specifies the control flow logic, and $\mathcal{X}$, $\mathcal{Y}$ represent global input and output schemas. Each module $ M_{i} = \left(\theta_{i}, \mathcal{X}_{i}, \mathcal{Y}_{i} \right) $ is an LLM subcomponent: $ \theta_{i}$ denotes the underlying model weights; and $X_{i}$ and $Y_{i}$ define module-level input and output schemas. We define $\pi_{i}$ as the system prompt containing instructions and few-shot examples required by $M_i$. At runtime, the control logic orchestrates the sequence and invocation of modules—for instance, passing outputs from one module to another, conditionally invoking components, or leveraging tool APIs.

In this paper, we only consider optimizing module prompts in the system. We denote $ \Pi = \left\langle \pi_{1}, \ldots, \pi_{|M|} \right\rangle $ as the set of all learnable module prompts. For simplicity, we refer an instance of $\Pi$ as an individual in following sections. To measure how an individual performs on the target task $\mathcal{T}$, we may denote each instance of this task as $(x, m)$, where $x$ can be mapped to the input schema $\mathcal{X}$, and $m$ denotes metadata required for evaluation (e.g., gold answers, evaluation rubrics, or test codes). For each instance, the system produces an output $y = \Phi \left( x ; \Pi \right)$. A metric $\mu : \mathcal{Y} \times \mathcal{M} \to [0, 1]$ evaluates the quality of $y$ with respect to metadata $m$, which can be quantified as scores like F1 or pass rates. With these notations, a classic evolutionary algorithm to optimize a single prompt can be formulated as:
\begin{equation}
\Pi^* = \arg\max_{ \Pi} \mathbb{E}_{(x, m) \sim \mathcal{T}} \left[ \mu \left( \Phi \left( x; \Pi \right), m \right) \right].
\end{equation}
In this paper, we will test on several Compound AI systems. We will elaborate on how they are implemented with multiple language modules in Appendix~\ref{app:task_description}. 

\textbf{Consensus.} 
Consensus is achieved by aggregating the outputs produced by a group of system prompts $\mathcal{G} = \left\langle \Pi^{1}, \ldots, \Pi^{|\mathcal{G}|} \right\rangle = \left\langle \left\langle \pi_{1}^{1}, \ldots, \pi_{|M|}^{1} \right\rangle, \ldots, \left\langle \pi_{1}^{|\mathcal{G}|}, \ldots, \pi_{|M|}^{|\mathcal{G}|} \right\rangle \right\rangle$ with a consensus aggregator $C$. For each task instance, each single prompt $\Pi^i$ produces its corresponding output $\Phi \left( x; \Pi \right)$ individually. Then, given $|\mathcal{G}|$ individual outputs $y^{\mathcal{G}} = \left\langle \Phi \left( x; \Pi^{1} \right), \ldots, \Phi \left( x; \Pi^{|\mathcal{G}|} \right) \right\rangle$, the consensus aggregator extracts the final results according to a majority voting principle: $y_\text{final} = C(y^{\mathcal{G}})$. For close-ended tasks such as multi-choice questions and mathematical calculation, we simply employ majority voting, which selects the answer with the most votes. For open-ended tasks that demand free text outputs, we employ an LLM-based aggregator: leveraging an LLM to identify the most representative response covering the contents of as many responses as possible within the group. The prompts used by the LLM-based aggregator for consensus extraction can be found in Appendix \ref{app:llm_based}.

To find the best-performing prompt group $\mathcal{G}^{*}$
with the consensus mechanism, the optimization problem is thus formulated as:
\begin{equation}
\mathcal{G}^{*} = \arg\max_{\mathcal{G}} \mathbb{E}_{(x, m) \sim \tau} \left[ \mu \left( \mathrm{C}(y^{\mathcal{G}}), m \right) \right]
\end{equation}
\begin{figure}[t]
\vspace{-5mm}
    \centering
    \includegraphics[width=1\linewidth]{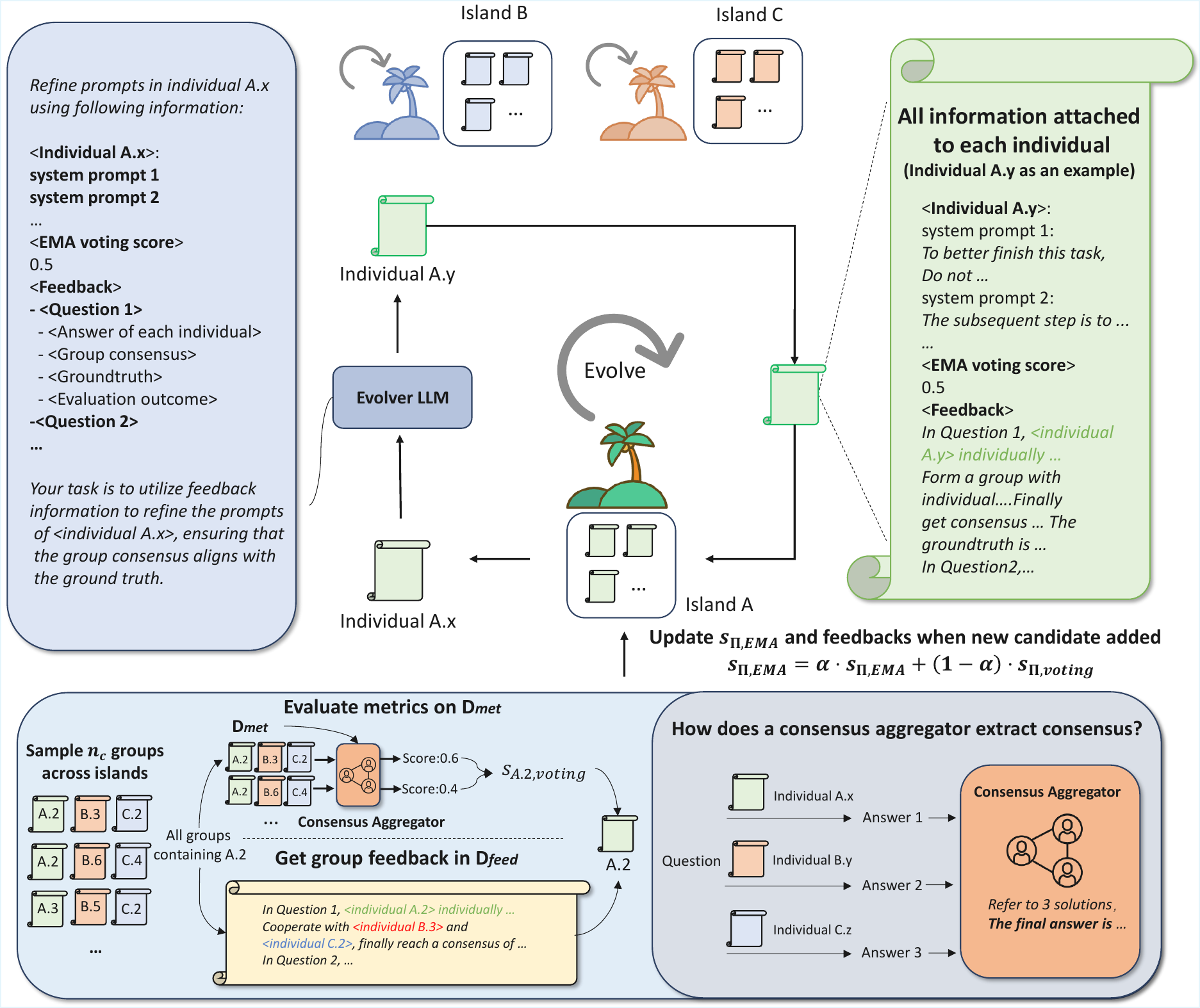}
    \vspace{-5mm}
    \caption{Evolutionary process in the voting stage of C-Evolve. In each iteration, C-Evolve samples an individual from each island. Then, we use the evolver LLM to analyze the feedback samples linked to selected individual and perform evolution, producing a new individual. After introducing a new individual into the island, we sample groups to update $s_{\Pi,\text{EMA}}$ and group feedback for individuals in the island.}
    \label{fig:placeholder}
    \vspace{-5mm}
\end{figure}

\section{C-Evolve: Consensus-based Evolutionary Algorithm}
\label{sec:method}
C-Evolve is an evolutionary algorithm designed for optimizing a group of individuals to improve the performance of a compound AI system. Its key feature is using a voting score to measure how well a prompt contributes when operating with others, rather than evaluating it in isolation. This design ensures that, after evolution, we can sample multiple prompts that work effectively through output consensus.


\subsection{Evolution Algorithm Based on Island Model}
During evolution, the algorithm iteratively performs sampling, generation, and evaluation of these individuals to update the population. Considering we would like to obtain a group of diverse individuals $\mathcal{G}$ that can achieve strong performance after reaching a consensus, we adopt an island-based evolutionary algorithm here: we initialize $|\mathcal{G}|$ islands $\{P_i\}_{i=1}^{|\mathcal{G}|}$, and evolve individuals within each island in parallel. This parallel evolutionary strategy ensures that different islands follow distinct evolutionary trajectories, which promotes population diversity and ultimately benefits effective consensus. We configure an LLM to refine the prompts within individuals based on feedback information, which we call the evolver. When refining individuals, the prompt required by the evolver can be referenced in Appendix \ref{app:evolver}.

Specifically, this evolutionary process consists of two stages: 1) a warm-up stage that evolves individuals based on their own performance, and 2) a voting stage that evolves individuals based on their performance when grouped with others from different islands. During evolution, we employ a metric dataset $D_\text{met}$ with evaluation metric $\mu$ in both warm-up stage and voting stage, and a separate feedback dataset $D_\text{feed}$ that supplies additional samples $(x, m)$ to generate detailed execution feedback for each individual.

\vspace{-3mm}
\paragraph{The warm-up stage.} In this stage, we define the metric value evaluated for an individual $\Pi$ on the dataset $D_\text{met}$ as an individual score $s_{\Pi,\text{ind}}$, and regard it as the fitness score of each individual. We consider generating exactly one new individual per island as a single iteration. In each iteration, we first sample a seed individual from the island using a categorical distribution constructed from $\{s_{\Pi,\text{ind}}\}_{\Pi\in P_i}$ (Appendix~\ref{app:algorithms} Algorithm~\ref{alg:performance-based-sampling}), so that higher-performing individuals have a greater probability of being selected. Then, the selected seed individual, together with its execution feedback on a minibatch (batchsize defaulting to 3) sampled from $D_\text{feed}$, is provided to the evolver LLM to produce a new individual. 


To prevent unbounded growth, the number of individuals in each island is capped at $N_{max}$. Whenever this limit is exceeded after producing a new individual, the worst-performing individual measured on $D_\text{met}$ is eliminated. The overall warm-up procedure is summarized in Appendix~\ref{app:algorithms} Algorithm~\ref{alg:c-evolve}.


\vspace{-3mm}
\paragraph{The voting stage.}In this stage, each individual's fitness is determined by its performance within prompt groups. As shown in Figure~\ref{fig:placeholder}, after introducing a new individual into each island, C-Evolve samples $n_c$ groups $\{\mathcal{G}_k\}_{k=1}^{n_c}$. Each group $\mathcal{G}_k$ is formed by selecting exactly one individual $\Pi^{G_k(i)}$ from each island $i$, where $G_k(i)$ denotes the index of the individual sampled from island $i$ for group $\mathcal{G}_k$. 
The outputs of all individuals within a group are integrated by the consensus aggregator to produce the final answer, and the group's performance is then evaluated on the metric set $D_\text{met}$.

A naive idea would be to eliminate all member individuals in the worst-performing group based on consensus results. However, this approach is problematic due to the possible overlap of members between the worst-performing group and other better-performing ones. Aggressively pruning all members of the worst-performing group thus makes no sense in this context. To address this issue, we define a voting score $s_{\Pi,\text{voting}}$ to measure each individual’s contribution within consensus-driven groups:
\begin{equation}\label{eq:cooperation_score}
s_{\Pi,\text{voting}} = \frac{
    \sum_{k=1}^{n_{c}} \mathbb{I}\left(\Pi \in \mathcal{G}_{k}\right) \cdot 
    \mathbb{E}_{(x, m)\sim\mathcal{D}_{\text{met}}}\left[u\left(C\left(\mathbf{y}^{\mathcal{G}_{k}}\right), m\right)\right]
}{
    \sum_{k=1}^{n_{c}} \mathbb{I}\left(\Pi \in \mathcal{G}_{k}\right)
}
\end{equation}
where $\mathbb{I}\left( \Pi \in G_{k} \right)$ is an indicator function that equals 1 if the individual $\Pi$ belongs to group $\mathcal{G}_{k}$. This score measures the average metric value over the consensus results by the groups containing the $\Pi$. In other words, the higher the voting score, the more likely the group containing this individual could yield a higher metric value. In this sense, the individuals with higher voting scores ought to be sampled to form a high-performing group.

In order to balance the voting scores computed in different iterations, we take the smoothed $s_{\text{EMA}, \Pi}$ as the fitness score in evolution, and apply exponential moving average (EMA) to update it:
\begin{equation}
    s_{\Pi,\text{EMA}} \leftarrow \alpha \cdot s_{\Pi,\text{EMA}} + (1 - \alpha) \cdot s_{\Pi,\text{voting}}
    \label{eq:ema_voting}
\end{equation}
where $\alpha \in [0, 1]$ is the smoothing factor balancing the weight of historical performance and current performance. After updating, the individual with the lowest $s_{\Pi,\text{EMA}}$ in each island is then eliminated. In the initial phase of voting stage, we initialize $s_{\Pi,\text{EMA}}$ with the individual score obtained at the end of the warm-up stage. Upon completion of all evolutionary iterations, we select the individual with the highest EMA voting score from each island to form the final group. During inference, we use the final group to perform reasoning and aggregate each individual's opinion to reach a consensus.

We deliberately avoid using the simple average of an individual's performance across all historical groups containing it as its final score. The rationale is that members from groups formed in the distant past may have already been eliminated from the current population. Incorporating these outdated consensus outcomes into the evaluation metric and assigning them the same weight as recent performance would be unreasonable, as it fails to faithfully reflect the individual's actual contribution within the current population or its recent evolutionary context. Therefore, we employ an EMA method to assign higher weight to recent or immediate scores, thereby suppressing the influence of historical outcomes from the distant past on the individual's evaluation.



\subsection{C-Evolve Based on Group Feedbacks}


During the execution of a compound AI system, the reasoning traces generated by the system on various questions provide valuable insights for understanding and analyzing an individual prompt. When combined with the final outcomes (e.g., success, failure, or performance scores), these traces offer significant diagnostic value, enabling evolver to figure out when errors may occur and identify potential issues within specific modules \cite{}. 


Specifically, during the warm-up stage, to evolve a given individual, C-Evolve compiles feedback information from the generated reasoning traces on a feedback dataset $D_\text{feed}$. This includes details of the prompts in individual, the inputs and outputs of each language module within the system, and its individual metric score on $D_\text{met}$. The evolver then uses this feedback to produce a newly evolved individual. 


During the voting stage, we incorporate the reasoning traces from a prompt group into the {\em group feedback}. More specifically, the feedback for each individual includes: information about the prompts in this individual, its grouping information, each individual's output in the group, consensus reached by the group, the metric scores for the group consensus on $D_\text{met}$, 
alongside inputs and outputs by all LLM modules in the system. Leveraging this group feedback, C-Evolve uses the evolver to produce a new individual. The overall voting stage is summarized in Appendix~\ref{app:algorithms} Algorithm~\ref{alg:c-evolve-cooperation}.



\section{Experiments}

\subsection{Benchmark Datasets}
To rigorously evaluate the performance of C-Evolve, we conducted experiments on two distinct task categories, each employing different consensus aggregators: for closed-ended tasks, we employ majority voting to select the answer with the most votes, while for open-ended tasks, we leverage an LLM to identify the most representative response within the group. We provide a comprehensive elucidation of the datasets and their splittings, evaluation metrics, consensus aggregators for each task in Appendix~\ref{app:benchmarks} and detailed feedback content in Appendix~\ref{fig:Feedback}.

\subsection{Models}
We conduct experiments on two large language models: Qwen3-8B~\cite{yang2025qwen3technicalreport} and GPT4.1-mini~\cite{openai2025gpt41}. These models represent mainstream open-source and closed-source models respectively. We provide the detailed configurations of the two models in Appendix~\ref{app:models}.

\subsection{Training Details}

We select state-of-the-art evolutionary algorithms, GEPA~\cite{agrawal2025gepa} and AlphaEvolve~\cite{novikov2025alphaevolve}, to provide a comprehensive comparison. For GEPA, we evaluate the optimal system prompts provided in its paper. For AlphaEvolve, which is not open-sourced, we report results from our own reimplementation. 

All experiments start with the baseline prompts provided in Appendix~\ref{app:best_prompt}. We set the number of islands to be $|\mathcal{G}|=3$, with a population limit of $N_{max}=10$ individuals per island. For each individual, we sample 3 questions from $D_\text{feed}$ to construct the feedback. In each iteration of the voting stage, we sample $n_c=10$ groups and evaluate them to update $s_{\Pi,\text{EMA}}$.
In addition, we employ a 10\% migration rate between islands to avoid getting stuck in local optima.

\subsection{Results and Analysis}
\begin{table}[t]
\centering
\caption{Performance comparison across multiple tasks: open-ended tasks (HotpotQA~\cite{yang2018hotpotqadatasetdiverseexplainable}, IFBench~\cite{pyatkin2025generalizingverifiableinstructionfollowing}) and closed-ended tasks(Hover~\cite{he2025hoverversatileneuralwholebody}, MATH~\cite{lightman2023lets}, GPQA~\cite{rein2023gpqagraduatelevelgoogleproofqa}) using Qwen3-8B and GPT-4.1 mini. ``Improvement" denotes the average relative gain of C-Evolve over baseline across available tasks for each model. 
\textbf{Bold} values indicate the best result for each task.}
\label{tab:model_compare}
\begin{tabular}{l|cc|ccc|c}\toprule

Model & HotpotQA & IFBench &Hover & MATH&GPQA& Improvement \\\midrule
\multicolumn{7}{c}{\textbf{Qwen3-8B}}\\\midrule
\quad Baseline & 50.03& 31.29&37.66& 67.66&41.43& -\\
\quad GEPA& 65.72& 34.01&43.33& -&-& +8.02\\
\quad AlphaEvolve& 65.31& 41.15&44.66& 82.66&43.08& +9.75\\
\quad C-Evolve& \textbf{70.67}& \textbf{43.88}&\textbf{50.33}& \textbf{85.33}&\textbf{47.15}& \textbf{+13.85}\\\midrule

\multicolumn{7}{c}{\textbf{GPT-4.1 mini}}\\\midrule
\quad Baseline & 44.24& 40.13
&42& 78.66&46.34& -\\
\quad GEPA& 68.39& 46.59&39& -&-& +9.2\\
\quad AlphaEvolve& 67.31& 45.24&50.33& 92.66&63.01& +13.42\\
 \quad C-Evolve& \textbf{70.64}& \textbf{47.96}&\textbf{51.66}& \textbf{95.33}&\textbf{66.26}&\textbf{+16.09}\\ \bottomrule 
\end{tabular}
\vspace{-5mm}
\end{table}
\subsubsection{C-Evolve Outperforms the Single Prompt Evolution}
Table~\ref{tab:model_compare} presents the performance of C-Evolve on two open-ended and three closed-ended tasks, using both the open-source model Qwen-3-8B and the closed-source model GPT-4.1-mini. On Qwen-3-8B, C-Evolve achieves an average improvement of \textbf{13.85\%} over the baseline, outperforming GEPA by \textbf{5.83\%} and AlphaEvolve by \textbf{4.1\%}. On the more capable GPT-4.1-mini, C-Evolve still maintains a consistent advantage, yielding a \textbf{6.89\%} average improvement over GEPA and \textbf{2.67\%} over AlphaEvolve.

To further demonstrate the effectiveness of C-Evolve, the curves in Figure~\ref{fig:experiment_score_per_island} showcase how the highest EMA voting score in the population changes during evolution on HotPotQA and HoVer tasks. 
AlphaEvolve exhibits rapid improvement in the early stage but soon plateaus,
while C-Evolve continues to improve after entering the voting stage.
As shown in Table~\ref{tab:experiment_group_per_islands}, the best individuals from the three islands in AlphaEvolve achieve an accuracy of 41.15\%, 36.73\%, and 37.07\% respectively, with the overall best individual performance being 41.15\% (the first line of AlphaEvolve result) without any consensus mechanism among islands as in the original AlphaEvolve. We try to group these best individuals selected from each island, and apply the LLM-based aggregator to them directly, which remains 41.15\% (the second line of AlphaEvolve result). This arises because AlphaEvolve evolves the prompts solely based on their individual scores without considering consensus. In contrast, C-Evolve evolves the prompts with the EMA voting score as their fitness, ensuring that individuals with high scores are more likely to perform well when they are grouped across the islands. Thus, individuals selected by EMA voting score in C-Evolve reach a higher accuracy of \textbf{43.88\%} after reaching consensus. 


\begin{table}
    \centering
    \caption{Comparison between AlphaEvolve and C-Evolve on IFBench. We select the individual with the highest fitness score, either individual score in Alpha-Eolve or EMA voting score in C-Evolve, from all three islands, and the accuracy is reported on the test set. For AlphaEvolve, we report both the accuracy of the best single prompt, and the accuracy directly applying the consensus mechanism to the three selected prompts.}
    \begin{tabular}{lcccc}\toprule
         Method&  island 1&  island 2&  island 3&  Final Accuracy\\\midrule
          \multirow{2}{*}{AlphaEvolve}&  \multirow{2}{*}{41.15}&  \multirow{2}{*}{36.73}&  \multirow{2}{*}{37.07}&  41.15 (individual)\\
 & & & &41.15 (consensus)\\
         C-Evolve&  40.47&  39.79&  38.77&  \textbf{43.88}\\ \bottomrule 
    \end{tabular}

    \label{tab:experiment_group_per_islands}
    \vspace{-3mm}
\end{table}

\begin{table}
    \centering
    \caption{Accuracy of individuals evolved by C-Evolve on the test set of MATH when evaluating on subsets of different difficulty levels. Each subset contains 300 problems.
    We report both the accuracy of single individuals with highest $s_{\Pi,\text{EMA}}$ in each island, and the final accuracy of after reaching a consensus.}
    \begin{tabular}{cccccc}\toprule
         &  level 1&  level 2&  level 3&  level 4&  level 5 \\\midrule
         island 1&  95&  94.67&  90.66&  82.66&  66.66 \\
         island 2&  96.67&  96&  90&  80&  63.66 \\
         island 3&  95.66&  95.66&  90.33&  82&  62.66 \\
 C-Evolve& \textbf{96.67}& \textbf{96}& \textbf{92}& \textbf{82.66}& \textbf{68.33} \\ \bottomrule 
    \end{tabular}

    \label{tab:experiment_difficulty}
    \vspace{-5mm}
\end{table}
 \subsubsection{Multi-Island Evolution Exhibits Diverse Focuses}\label{sec:5.4.3}
C-Evolve maintains population diversity through multi-island evolution. Appendix \ref{app:5.4.3} Figure~\ref{fig:experiment_diff_prompts} illustrates the evolutionary trajectories, from the initial individual (baseline) to the ones selected in the final group on IFBench. We observe that individuals from different islands gradually developed distinct focuses during evolution. 
The individual selected in island 1 focuses on the principles of core constraint prioritization and refined amendment, while the individual selected in island 2 specializes in precise constraint pre-declaration and hierarchical verification. The individual selected in island 3 concentrates on task priority stratification and critical failure point verification. 


When aggregated, these prompts focus on distinct aspects, enabling stronger and more comprehensive problem-solving capabilities. We visualize the prompts evolved in each island on IFBench with t-SNE \cite{2008Visualizing}, as shown in Appendix \ref{app:5.4.3} Figure~\ref{fig:experiment_reduce}. 
We observed that, during the voting stage, the individuals with the highest EMA voting scores on each island gradually diverge, and three islands tend to form distinct clusters. 
\vspace{-3mm}
\subsubsection{Consensus Helps on Hard Problems} \label{sec:5.4.2}
After evolving on MATH, we select the individuals with the highest EMA voting scores from each island on the metric set, and evaluate on the test set both individually and by reaching a consensus with majority voting.
As shown in Table~\ref{tab:experiment_difficulty}, 
the accuracies of single individuals consistently declines as task difficulty increases. At Level 5, all islands fall below 67\%, revealing the weakness of single-prompt solutions for harder problems.

After majority voting, C-Evolve consistently outperforms all islands on Levels 3–5, achieving accuracies of 92.00\%, 82.66\%, and 68.33\%, respectively. This indicates that problems unsolvable by a single optimal prompt can often be addressed through majority voting.
This can be validated with further analysis in Appendix~\ref{app:5.4.2} Table~\ref{tab:app_math}: among the problems that two of the three individuals reach a consensus (25\% in all problems), C-Evolve correctly solves 49.33\% of them, while the single optimal prompt only solves 40\%.
\begin{table}[t]
    \centering
    \begin{minipage}{0.48\textwidth}
        \centering
        \caption{Ablation study on LLM-based aggregator for open-ended tasks. Conducted on IFBench, using Qwen3-8B. Both are trained 50 iterations for the warm-up stage and 50 iterations for the voting stage.}
        \label{tab:ablation_llm}
        \begin{tabular}{cc}\toprule
             Aggregator & Accuracy (\%) \\\midrule
             LLM-summary        & 38.66 \\
             LLM-selection      & \textbf{42.85} \\ \bottomrule
        \end{tabular}
    \end{minipage}
    \hfill 
    \begin{minipage}{0.48\textwidth}
        \centering
        \caption{Ablation study on different methods averaging voting scores across iterations. Conducted on IFBench, using Qwen3-8B. Both are trained 50 iterations for the warm-up stage and 50 iterations for the voting stage.}
        \label{tab:ablation_moving_average}
        \begin{tabular}{lcc}\toprule
            Method         & $\alpha$ & Accuracy (\%) \\\midrule
            Group Average  & --       &  41.16\\
            EMA            & 0.5      & 42.51\\
            EMA            & 0.8      & \textbf{42.85}\\ \bottomrule 
        \end{tabular}
    \end{minipage}
    \vspace{-5mm}
\end{table}

\subsection{Ablation Study}
\subsubsection{Ablation Study of LLM-based Aggregator}
For open-ended tasks (HotpotQA and IFBench), we use an LLM to aggregate individual answers into a final answer. We compare two LLM-based aggregators: (1) LLM-summary, where the LLM generates an answer synthesizing all individual answers within the group; (2) LLM-selection, where the LLM selects the most representative answer from individual answers within the group. 

As shown in Table~\ref{tab:ablation_llm}, the LLM-selection outperforms LLM-summary. This is because summarization tends to generate content that is not strictly grounded in any individual answer, whereas selection preserves the fidelity of original outputs while identifying the most representative one.

\subsubsection{Ablation Study of Moving Average}

In Eq.~\ref{eq:ema_voting}, we apply EMA to update $s_{\Pi,\text{EMA}}$ for each individual. Here we validate its necessity. We first simply average the individual's performance across all groups and iterations. It performs poorly (41.16\%) on IFBench, using Qwen3-8B. The simple average assigns equal weights to all groups regardless of when they were formed, while many of the groups formed early in evolution may have already been eliminated. Thus, they cannot represent the individual's contribution when grouping with individuals in the current population. Applying EMA results in more reliable evaluations. As shown in Table~\ref{tab:ablation_moving_average}, C-Evolve achieves its best performance with $\alpha$ = 0.8, reaching an accuracy of \textbf{42.85\%}. 

\subsubsection{efficiency and performance trade off}
\begin{table}[h]
\vspace{-5mm}
    \centering
    \caption{Evolving and Inference Time Comparison with Accuracy on GPQA, using Qwen3 8B.}
    \begin{tabular}{cccc}\toprule
         &  Evolving time (min/iteration)&  Inference time (s/question)&  Accuracy\\\midrule
         AlphaEvlove&  3.5&  0.68&  43.08\\
 C-Evolve& 4.3& 0.79& \textbf{47.15}\\ \bottomrule 
    \end{tabular}
    \label{tab:experiment_time}
    \vspace{-5mm}
\end{table}
Due to calculating the voting score in evolving and multi-prompt reasoning in inference, C-Evolve introduces additional computational overhead. To this end, we employ two key optimizations to effectively balance efficiency and performance:

\textbf{Training Efficiency via Caching.}
During evolution, each newly generated prompt is first evaluated on the metric set to generate all answers, and these answers are cached. When computing the voting score, we reuse these cached outputs and apply majority voting across individuals within a group directly, which eliminates the need for repeated LLM calls during the voting stage. This significantly reduces training time: on the GPQA task, each iteration in C-Evolve takes only 0.8 minutes per iteration longer than AlphaEvolve (3.5 min/iteration) in Table~\ref{tab:experiment_time}.

\textbf{Inference Acceleration via Asynchronous Prompting.}
During inference, we leverage asynchronous execution to prompt the LLM with distinct system prompts in parallel. This enables us to obtain multiple answers simultaneously, delivering a 9.45\% relative improvement in accuracy compared to AlphaEvolve while only increasing inference time from 0.68 to 0.79 seconds per question.  As for open-ended tasks, it indeed needs another LLM request to complete majority voting. 

Additional ablation studies, such as those on the warm-up stage and voting score, are presented in the Appendix~\ref{app:ablation_study}. 



\vspace{-3mm}
\section{Conclusion}
\vspace{-3mm}

In this paper, we present Consensus-Evolve (C-Evolve), a consensus-driven evolutionary algorithm for optimizing a group of prompts in AI systems. 
The outputs from these prompts are aggregated for reaching a consensus. C-Evolve introduces an EMA voting score as the individual fitness, which evaluates each prompt’s contribution to consensus within groups. To maintain diversity, we employ a multi-island  evolution system to explore complementarity among prompts.

Our experiments show that C-Evolve achieves state-of-the-art results on multiple benchmark datasets using both Qwen3-8B and GPT-4.1-mini. To our best knowledge, C-Evolve is the first  algorithm developing the group consensus among prompts to drive their evolution for making more accurate predictions. This opens new directions for optimizing compound AI systems, especially for those built on proprietary models.




\bibliography{iclr2025_conference}

\begin{thebibliography}{37}
\providecommand{\natexlab}[1]{#1}
\providecommand{\url}[1]{\texttt{#1}}
\expandafter\ifx\csname urlstyle\endcsname\relax
  \providecommand{\doi}[1]{doi: #1}\else
  \providecommand{\doi}{doi: \begingroup \urlstyle{rm}\Url}\fi

\bibitem[Agarwal et~al.(2025)Agarwal, Magazine, Singh, Dani, Ganu, and Nambi]{agarwal2025promptwizard}
Eshaan Agarwal, Raghav Magazine, Joykirat Singh, Vivek Dani, Tanuja Ganu, and Akshay Nambi.
\newblock Promptwizard: Optimizing prompts via task-aware, feedback-driven self-evolution.
\newblock In \emph{Findings of the Association for Computational Linguistics: ACL 2025}, pp.\  19974--20003, 2025.

\bibitem[Agrawal et~al.(2025)Agrawal, Tan, Soylu, Ziems, Khare, Opsahl-Ong, Singhvi, Shandilya, Ryan, Jiang, et~al.]{agrawal2025gepa}
Lakshya~A Agrawal, Shangyin Tan, Dilara Soylu, Noah Ziems, Rishi Khare, Krista Opsahl-Ong, Arnav Singhvi, Herumb Shandilya, Michael~J Ryan, Meng Jiang, et~al.
\newblock Gepa: Reflective prompt evolution can outperform reinforcement learning.
\newblock \emph{arXiv preprint arXiv:2507.19457}, 2025.

\bibitem[Anthropic(2024)]{anthropic2024claude35sonnet}
Anthropic.
\newblock Claude 3.5 sonnet.
\newblock Large language model, June 2024.
\newblock URL \url{https://www.anthropic.com/news/claude-3-5-sonnet}.
\newblock Released June 2024.

\bibitem[C{\^a}mara et~al.(2025)C{\^a}mara, Luz, Carvalho, Meneghini, and Moreira]{camara2025moprompt}
Sara C{\^a}mara, Eduardo Luz, Val{\'e}ria Carvalho, Ivan Meneghini, and Gladston Moreira.
\newblock Moprompt: Multi-objective semantic evolution for prompt optimization.
\newblock \emph{arXiv preprint arXiv:2508.01541}, 2025.

\bibitem[Chen et~al.(2025)Chen, Zhang, Langren{\'e}, and Zhu]{chen2025unleashing}
Banghao Chen, Zhaofeng Zhang, Nicolas Langren{\'e}, and Shengxin Zhu.
\newblock Unleashing the potential of prompt engineering for large language models.
\newblock \emph{Patterns}, 2025.

\bibitem[Fernando et~al.(2023)Fernando, Banarse, Michalewski, Osindero, and Rockt{\"a}schel]{fernando2023promptbreeder}
Chrisantha Fernando, Dylan Banarse, Henryk Michalewski, Simon Osindero, and Tim Rockt{\"a}schel.
\newblock Promptbreeder: Self-referential self-improvement via prompt evolution.
\newblock \emph{arXiv preprint arXiv:2309.16797}, 2023.

\bibitem[Gr{\"o}tschla et~al.(2025)Gr{\"o}tschla, M{\"u}ller, T{\"o}nshoff, Galkin, and Perozzi]{grotschla2025agentsnet}
Florian Gr{\"o}tschla, Luis M{\"u}ller, Jan T{\"o}nshoff, Mikhail Galkin, and Bryan Perozzi.
\newblock Agentsnet: Coordination and collaborative reasoning in multi-agent llms.
\newblock \emph{arXiv preprint arXiv:2507.08616}, 2025.

\bibitem[He et~al.(2025)He, Xiao, Lin, Luo, Xu, Jiang, Kautz, Liu, Shi, Wang, Fan, and Zhu]{he2025hoverversatileneuralwholebody}
Tairan He, Wenli Xiao, Toru Lin, Zhengyi Luo, Zhenjia Xu, Zhenyu Jiang, Jan Kautz, Changliu Liu, Guanya Shi, Xiaolong Wang, Linxi Fan, and Yuke Zhu.
\newblock Hover: Versatile neural whole-body controller for humanoid robots, 2025.
\newblock URL \url{https://arxiv.org/abs/2410.21229}.

\bibitem[Hendrycks et~al.(2021)Hendrycks, Burns, Kadavath, Arora, Basart, Tang, Song, and Steinhardt]{hendrycks2021measuring}
Dan Hendrycks, Collin Burns, Saurav Kadavath, Akul Arora, Steven Basart, Eric Tang, Dawn Song, and Jacob Steinhardt.
\newblock Measuring mathematical problem solving with the math dataset.
\newblock \emph{arXiv preprint arXiv:2103.03874}, 2021.

\bibitem[Izzo et~al.(2012)Izzo, Ruci{\'n}ski, and Biscani]{izzo2012generalized}
Dario Izzo, Marek Ruci{\'n}ski, and Francesco Biscani.
\newblock The generalized island model.
\newblock In \emph{Parallel architectures and bioinspired algorithms}, pp.\  151--169. Springer, 2012.

\bibitem[Kepel \& Valogianni(2024)Kepel and Valogianni]{kepel2024autonomous}
Daan Kepel and Konstantina Valogianni.
\newblock Autonomous prompt engineering in large language models.
\newblock \emph{arXiv preprint arXiv:2407.11000}, 2024.

\bibitem[Khan et~al.(2024)Khan, Chaudhari, and Chandra]{khan2024review}
Azal~Ahmad Khan, Omkar Chaudhari, and Rohitash Chandra.
\newblock A review of ensemble learning and data augmentation models for class imbalanced problems: Combination, implementation and evaluation.
\newblock \emph{Expert Systems with Applications}, 244:\penalty0 122778, 2024.

\bibitem[Laurens \& Hinton(2008)Laurens and Hinton]{2008Visualizing}
Van Der~Maaten Laurens and Geoffrey Hinton.
\newblock Visualizing data using t-sne.
\newblock \emph{Journal of Machine Learning Research}, 9\penalty0 (2605):\penalty0 2579--2605, 2008.

\bibitem[Lightman et~al.(2023)Lightman, Kosaraju, Burda, Edwards, Baker, Lee, Leike, Schulman, Sutskever, and Cobbe]{lightman2023lets}
Hunter Lightman, Vineet Kosaraju, Yura Burda, Harri Edwards, Bowen Baker, Teddy Lee, Jan Leike, John Schulman, Ilya Sutskever, and Karl Cobbe.
\newblock Let's verify step by step.
\newblock \emph{arXiv preprint arXiv:2305.20050}, 2023.

\bibitem[Liu et~al.(2024)Liu, Feng, Xue, Wang, Wu, Lu, Zhao, Deng, Zhang, Ruan, et~al.]{liu2024deepseek}
Aixin Liu, Bei Feng, Bing Xue, Bingxuan Wang, Bochao Wu, Chengda Lu, Chenggang Zhao, Chengqi Deng, Chenyu Zhang, Chong Ruan, et~al.
\newblock Deepseek-v3 technical report.
\newblock \emph{arXiv preprint arXiv:2412.19437}, 2024.

\bibitem[Monea et~al.(2024)Monea, Bosselut, Brantley, and Artzi]{monea2024llms}
Giovanni Monea, Antoine Bosselut, Kiant{\'e} Brantley, and Yoav Artzi.
\newblock Llms are in-context bandit reinforcement learners.
\newblock \emph{arXiv preprint arXiv:2410.05362}, 2024.

\bibitem[Novikov et~al.(2025)Novikov, V{\~u}, Eisenberger, Dupont, Huang, Wagner, Shirobokov, Kozlovskii, Ruiz, Mehrabian, et~al.]{novikov2025alphaevolve}
Alexander Novikov, Ng{\^a}n V{\~u}, Marvin Eisenberger, Emilien Dupont, Po-Sen Huang, Adam~Zsolt Wagner, Sergey Shirobokov, Borislav Kozlovskii, Francisco~JR Ruiz, Abbas Mehrabian, et~al.
\newblock Alphaevolve: A coding agent for scientific and algorithmic discovery.
\newblock \emph{arXiv preprint arXiv:2506.13131}, 2025.

\bibitem[OpenAI(2025)]{openai2025gpt41}
OpenAI.
\newblock Gpt-4.1 series.
\newblock Large language model series, April 2025.
\newblock URL \url{https://openai.com/index/gpt-4-1/}.
\newblock Released April 2025.

\bibitem[Opsahl-Ong et~al.(2024)Opsahl-Ong, Ryan, Purtell, Broman, Potts, Zaharia, and Khattab]{opsahl2024optimizing}
Krista Opsahl-Ong, Michael~J Ryan, Josh Purtell, David Broman, Christopher Potts, Matei Zaharia, and Omar Khattab.
\newblock Optimizing instructions and demonstrations for multi-stage language model programs.
\newblock \emph{arXiv preprint arXiv:2406.11695}, 2024.

\bibitem[Pedregosa et~al.(2011)Pedregosa, Varoquaux, Gramfort, Michel, Thirion, Grisel, Blondel, Prettenhofer, Weiss, Dubourg, Vanderplas, Passos, Cournapeau, Brucher, Perrot, and Duchesnay]{scikit-learn}
F.~Pedregosa, G.~Varoquaux, A.~Gramfort, V.~Michel, B.~Thirion, O.~Grisel, M.~Blondel, P.~Prettenhofer, R.~Weiss, V.~Dubourg, J.~Vanderplas, A.~Passos, D.~Cournapeau, M.~Brucher, M.~Perrot, and E.~Duchesnay.
\newblock Scikit-learn: Machine learning in {P}ython.
\newblock \emph{Journal of Machine Learning Research}, 12:\penalty0 2825--2830, 2011.

\bibitem[Polikar(2012)]{polikar2012ensemble}
Robi Polikar.
\newblock Ensemble learning.
\newblock In \emph{Ensemble machine learning}, pp.\  1--34. Springer, 2012.

\bibitem[Pyatkin et~al.(2025)Pyatkin, Malik, Graf, Ivison, Huang, Dasigi, Lambert, and Hajishirzi]{pyatkin2025generalizingverifiableinstructionfollowing}
Valentina Pyatkin, Saumya Malik, Victoria Graf, Hamish Ivison, Shengyi Huang, Pradeep Dasigi, Nathan Lambert, and Hannaneh Hajishirzi.
\newblock Generalizing verifiable instruction following, 2025.
\newblock URL \url{https://arxiv.org/abs/2507.02833}.

\bibitem[Rein et~al.(2023)Rein, Hou, Stickland, Petty, Pang, Dirani, Michael, and Bowman]{rein2023gpqagraduatelevelgoogleproofqa}
David Rein, Betty~Li Hou, Asa~Cooper Stickland, Jackson Petty, Richard~Yuanzhe Pang, Julien Dirani, Julian Michael, and Samuel~R. Bowman.
\newblock Gpqa: A graduate-level google-proof q\&a benchmark, 2023.
\newblock URL \url{https://arxiv.org/abs/2311.12022}.

\bibitem[Rokach(2019)]{rokach2019ensemble}
Lior Rokach.
\newblock \emph{Ensemble learning: pattern classification using ensemble methods}.
\newblock World Scientific, 2019.

\bibitem[Romera-Paredes et~al.(2024)Romera-Paredes, Barekatain, Novikov, Balog, Kumar, Dupont, Ruiz, Ellenberg, Wang, Fawzi, et~al.]{romera2024mathematical}
Bernardino Romera-Paredes, Mohammadamin Barekatain, Alexander Novikov, Matej Balog, M~Pawan Kumar, Emilien Dupont, Francisco~JR Ruiz, Jordan~S Ellenberg, Pengming Wang, Omar Fawzi, et~al.
\newblock Mathematical discoveries from program search with large language models.
\newblock \emph{Nature}, 625\penalty0 (7995):\penalty0 468--475, 2024.

\bibitem[Sun et~al.(2024)Sun, Wang, Li, Wang, Zhang, and Liang]{sun2024improved}
Zhigang Sun, Guotao Wang, Pengfei Li, Hui Wang, Min Zhang, and Xiaowen Liang.
\newblock An improved random forest based on the classification accuracy and correlation measurement of decision trees.
\newblock \emph{Expert Systems with Applications}, 237:\penalty0 121549, 2024.

\bibitem[Tan et~al.(2025)Tan, Agrawal, Singhvi, Lai, Ryan, Klein, Khattab, Sen, and Zaharia]{tan2025langprobelanguageprogramsbenchmark}
Shangyin Tan, Lakshya~A Agrawal, Arnav Singhvi, Liheng Lai, Michael~J Ryan, Dan Klein, Omar Khattab, Koushik Sen, and Matei Zaharia.
\newblock Langprobe: a language programs benchmark, 2025.
\newblock URL \url{https://arxiv.org/abs/2502.20315}.

\bibitem[Thirunavukarasu et~al.(2023)Thirunavukarasu, Ting, Elangovan, Gutierrez, Tan, and Ting]{thirunavukarasu2023large}
Arun~James Thirunavukarasu, Darren Shu~Jeng Ting, Kabilan Elangovan, Laura Gutierrez, Ting~Fang Tan, and Daniel Shu~Wei Ting.
\newblock Large language models in medicine.
\newblock \emph{Nature medicine}, 29\penalty0 (8):\penalty0 1930--1940, 2023.

\bibitem[Tian \& Zhang(2024)Tian and Zhang]{tian2024optimizing}
Chuan Tian and Yilei Zhang.
\newblock Optimizing collaboration of llm based agents for finite element analysis.
\newblock \emph{arXiv preprint arXiv:2408.13406}, 2024.

\bibitem[Wang et~al.(2024)Wang, Wang, Athiwaratkun, Zhang, and Zou]{wang2024mixture}
Junlin Wang, Jue Wang, Ben Athiwaratkun, Ce~Zhang, and James Zou.
\newblock Mixture-of-agents enhances large language model capabilities.
\newblock \emph{arXiv preprint arXiv:2406.04692}, 2024.

\bibitem[Wang et~al.(2022)Wang, Wei, Schuurmans, Le, Chi, Narang, Chowdhery, and Zhou]{wang2022self}
Xuezhi Wang, Jason Wei, Dale Schuurmans, Quoc Le, Ed~Chi, Sharan Narang, Aakanksha Chowdhery, and Denny Zhou.
\newblock Self-consistency improves chain of thought reasoning in language models.
\newblock \emph{arXiv preprint arXiv:2203.11171}, 2022.

\bibitem[Webb \& Zheng(2004)Webb and Zheng]{webb2004multistrategy}
Geoffrey~I Webb and Zijian Zheng.
\newblock Multistrategy ensemble learning: Reducing error by combining ensemble learning techniques.
\newblock \emph{IEEE transactions on knowledge and data engineering}, 16\penalty0 (8):\penalty0 980--991, 2004.

\bibitem[Wei et~al.(2022)Wei, Wang, Schuurmans, Bosma, Xia, Chi, Le, Zhou, et~al.]{wei2022chain}
Jason Wei, Xuezhi Wang, Dale Schuurmans, Maarten Bosma, Fei Xia, Ed~Chi, Quoc~V Le, Denny Zhou, et~al.
\newblock Chain-of-thought prompting elicits reasoning in large language models.
\newblock \emph{Advances in neural information processing systems}, 35:\penalty0 24824--24837, 2022.

\bibitem[Yang et~al.(2025)Yang, Li, Yang, Zhang, Hui, Zheng, Yu, Gao, Huang, Lv, Zheng, Liu, Zhou, Huang, Hu, Ge, Wei, Lin, Tang, Yang, Tu, Zhang, Yang, Yang, Zhou, Zhou, Lin, Dang, Bao, Yang, Yu, Deng, Li, Xue, Li, Zhang, Wang, Zhu, Men, Gao, Liu, Luo, Li, Tang, Yin, Ren, Wang, Zhang, Ren, Fan, Su, Zhang, Zhang, Wan, Liu, Wang, Cui, Zhang, Zhou, and Qiu]{yang2025qwen3technicalreport}
An~Yang, Anfeng Li, Baosong Yang, Beichen Zhang, Binyuan Hui, Bo~Zheng, Bowen Yu, Chang Gao, Chengen Huang, Chenxu Lv, Chujie Zheng, Dayiheng Liu, Fan Zhou, Fei Huang, Feng Hu, Hao Ge, Haoran Wei, Huan Lin, Jialong Tang, Jian Yang, Jianhong Tu, Jianwei Zhang, Jianxin Yang, Jiaxi Yang, Jing Zhou, Jingren Zhou, Junyang Lin, Kai Dang, Keqin Bao, Kexin Yang, Le~Yu, Lianghao Deng, Mei Li, Mingfeng Xue, Mingze Li, Pei Zhang, Peng Wang, Qin Zhu, Rui Men, Ruize Gao, Shixuan Liu, Shuang Luo, Tianhao Li, Tianyi Tang, Wenbiao Yin, Xingzhang Ren, Xinyu Wang, Xinyu Zhang, Xuancheng Ren, Yang Fan, Yang Su, Yichang Zhang, Yinger Zhang, Yu~Wan, Yuqiong Liu, Zekun Wang, Zeyu Cui, Zhenru Zhang, Zhipeng Zhou, and Zihan Qiu.
\newblock Qwen3 technical report, 2025.
\newblock URL \url{https://arxiv.org/abs/2505.09388}.

\bibitem[Yang et~al.(2018)Yang, Qi, Zhang, Bengio, Cohen, Salakhutdinov, and Manning]{yang2018hotpotqadatasetdiverseexplainable}
Zhilin Yang, Peng Qi, Saizheng Zhang, Yoshua Bengio, William~W. Cohen, Ruslan Salakhutdinov, and Christopher~D. Manning.
\newblock Hotpotqa: A dataset for diverse, explainable multi-hop question answering, 2018.
\newblock URL \url{https://arxiv.org/abs/1809.09600}.

\bibitem[Zheng et~al.(2024)Zheng, Yin, Xie, Sun, Huang, Yu, Cao, Kozyrakis, Stoica, Gonzalez, Barrett, and Sheng]{zheng2024sglangefficientexecutionstructured}
Lianmin Zheng, Liangsheng Yin, Zhiqiang Xie, Chuyue Sun, Jeff Huang, Cody~Hao Yu, Shiyi Cao, Christos Kozyrakis, Ion Stoica, Joseph~E. Gonzalez, Clark Barrett, and Ying Sheng.
\newblock Sglang: Efficient execution of structured language model programs, 2024.
\newblock URL \url{https://arxiv.org/abs/2312.07104}.

\bibitem[Zhou et~al.(2022)Zhou, Muresanu, Han, Paster, Pitis, Chan, and Ba]{zhou2022large}
Yongchao Zhou, Andrei~Ioan Muresanu, Ziwen Han, Keiran Paster, Silviu Pitis, Harris Chan, and Jimmy Ba.
\newblock Large language models are human-level prompt engineers.
\newblock In \emph{The eleventh international conference on learning representations}, 2022.

\end{thebibliography}
\bibliographystyle{iclr2025_conference}

\clearpage
\appendix
\section{Algorithms}\label{app:algorithms}
In this section, we provide the performance-based sampling algorithm (Algorithm~\ref {alg:performance-based-sampling}), C-Evolve warm-up stage algorithm (Algorithm~\ref{alg:c-evolve}), and C-Evolve voting stage algorithm (Algorithm~\ref{alg:c-evolve-cooperation}).
\begin{algorithm}
\caption{Performance-Based Sampling}
\label{alg:performance-based-sampling}
\begin{algorithmic}[1]
\Require Pool $P = \{(\Pi^{j}, s_{\Pi^{j},.}, f^{j})\}_{j=1}^{|P|}$, where $\Pi_{\Phi}^{j}$ means the j-th individual. $s_{\Pi^{j},.}$ and $f^{j}$ are the score and feedback information of j-th individual, respectively. In the warm-up stage, the variable $s_{\Pi^{j},.}$ represents the individual score $s_{\Pi^{j},\text{ind}}$, whereas during the voting stage, $s_{\Pi^{j},.}$ denotes the EMA voting score $s_{\Pi^{j},\text{EMA}}$.

\State $p_j \gets \dfrac{\exp(s_{\Pi^{j},.})}{\sum_{k=1}^{|P|} \exp(s_{\Pi^{k},.})}$ 

\State $j^{*} \sim \text{Categorical}(\{p_j\}_{j=1}^{|P|})$ \Comment{Sample from categorical distribution}

\State \Return $(\Pi^{j^{*}}, s_{\Pi^{j^{*}},.}, f^{j^{*}})$
\end{algorithmic}
\end{algorithm}

\begin{algorithm}
\caption{C-Evolve Warm-up Stage}
\label{alg:c-evolve}
\begin{algorithmic}[1]
\Require System $\Phi$, dataset $D_{\text{feed}}$, eval metric $\mu$, initial prompt module $\Pi^{0}$ with score $s_{\Pi^{0},\text{ind}}$ and feedback $f^{0}$, islands $\{P_i\}_{i=1}^{|\mathcal{G}|}$ where $P_i$ is initialized as $\{(\Pi^{0}, s_{\Pi^{0},\text{individual}}, f^{0})\}$. Total number of iterations $T$. The population capacity of each island $N_{\text{max}}$.

\For{$t \gets 1$ to $T$}
    \For{each island $i \in \{1, \dots, |\mathcal{G}|\}$}
        \State $(\Pi^{*}, s_{{\Pi}^*,\text{ind}}, f^{*}) \gets \text{PerformanceBasedSample}(P_i)$ \Comment{See Algorithm 1}
        
        \State $\Pi' \gets \text{Generate new individual }(\Pi^{*}, f^{*})$
        
        \State $D_{\text{batch}} \gets \text{RandomSample}(D_\text{feed})$ 
        \State $f' \gets$ feedback of  $\Pi^{*}$ in $D_{\text{batch}}$

        \State $s_{\Pi^{'},\text{ind}} \gets \mathbb{E}_{(x, m) \sim D_{\text{met}}} u[\Phi(x, \Pi'), m]$
        
        \State $P_i \gets P_i \cup \{(\Pi', s_{\Pi^{'},\text{ind}}, f')\}$
        
        \If{$|P_i| > N_{\text{max}}$}
            \State $(\Pi_{\Phi}^{\text{weak}}, s_{\Pi^{\text{weak}},\text{ind}}, f^{\text{weak}}) \gets \underset{(\Pi, s_{\Pi,\text{ind}}, f)}{\operatorname{argmin}} s_{\Pi,\text{ind}}$
            \State $P_i \gets P_i \setminus \{(\Pi^{\text{weak}}, s_{\Pi^{\text{weak}},\text{ind}}, f^{\text{weak}})\}$
        \EndIf
    \EndFor
\EndFor

\State \Return $\{P_i\}_{i=1}^{|\mathcal{G}|}$
\end{algorithmic}
\end{algorithm}

\begin{algorithm}
\caption{C-Evolve Voting Stage}
\label{alg:c-evolve-cooperation}
\begin{algorithmic}[1]
\Require System $\Phi$, dataset $D_{\text{feed}}$, eval metric $u$, consensus aggregator $C$, island pools $\{P_{i}\}_{i=1}^{|\mathcal{G}|}$ initialized from the end of warm-up stage, number of iterations $T$, groups per round $n_{c}$, population capacity of each island $N_{\text{max}}$
\For{$t \gets 1$ to $T$}
    \State \textbf{Step 1:} Evolve using performance-based sampling
    \For{each island $i \in \{1, \dots, |\mathcal{G}|\}$}
        \State $(\Pi^{*}, s_{\Pi^{*},\text{EMA}}, f^{*}) \gets \text{PerformanceBasedSample}(P_i)$ \Comment{see Algorithm 1}
        \State $\Pi' \gets \text{Evolve}(\Pi^{*}, f^{*})$\Comment{Use evolver LLM}
        \State $s_{\Pi^{'},\text{EMA}} \gets s_{\Pi^{*},\text{EMA}}$ \Comment{Inherit parent score initially}
        \State $D_{\text{batch}} \gets \text{RandomSample}(D_\text{feed})$ 
        \State $f' \gets$ individual feedback of  $\Pi'$ in $D_{\text{batch}}$
        \Comment{prevent the absence of feedback for new individuals}
        
        \State $P_i \gets P_i \cup \{(\Pi', s_{\Pi^{'},\text{EMA}}, f')\}$
    \EndFor
    
    \State \textbf{Step 2:} Form cross-island consensus-driven groups
    
    \For{$k \gets 1$ to $n_c$}
        \State $\mathcal{G}_k \gets \emptyset$
        \For{each island $i \in \{1, \dots, |\mathcal{G}|\}$}
            \State $(\Pi^{G_k(i)}, s_{\Pi^{G_k(i)}}, f^{G_k(i)}) \gets \text{PerformanceBasedSample}(P_i)$ \Comment{see Algorithm 1}
            \State $\mathcal{G}_k \gets \mathcal{G}_k \cup \{\Pi^{G_k(i)}\}$
        \EndFor
    \EndFor
    
    \State \textbf{Step 3:} Evaluate consensus performance
    \For{$k \gets 1$ to $n_c$}
        \State $y^{\mathcal{G}_k} \gets \langle \Phi(x; \Pi^{G_k(1)}), \dots, \Phi(x; \Pi^{G_k(|\mathcal{G}|)}) \rangle$, $(x, m) \in D_{\text{met}}$
        \State $y_\text{final}^k \gets C(y^{\mathcal{G}_k})$
    \EndFor
    
    \State \textbf{Step 4:} Update EMA voting score and feedback
    \For{each island $i \in \{1, \dots, |\mathcal{G}|\}$}
        \For{each candidate $(\Pi, s_{\Pi,\text{EMA}}, f) \in P_i$}
            \If{$\Pi_\Phi$ is contained in $n_c$ groups sampled in \textbf{Step 2}}
                \State$s_{\Pi,\text{voting}} \leftarrow \frac{1}{\sum_{k=1}^{n_{c}} \mathbb{I}\left(\Pi \in \mathcal{G}_{k}\right)}  \sum_{k=1}^{n_{c}}\mathbb{I}\left(\Pi \in \mathcal{G}_{k}\right) \mathbb{E}_{(x, m) \sim D_{\text{met}}} u(y_\text{final}^k, m)$
                \If{$s_{\Pi,\text{EMA}}$ is directly initialized from $s_{\Pi,\text{ind}}$ in warm-up stage}
                    \State updated $s_{\Pi,\text{EMA}} \gets s_{\Pi,\text{voting}}$
                \Else
                    \State updated $s_{\Pi,\text{EMA}} \gets (1-\alpha) \cdot s_{\Pi,\text{voting}} + \alpha \cdot s_{\Pi,\text{EMA}}$ \Comment{$\alpha \in [0, 1]$}
                \EndIf
                \If{$\Pi$ is sampled during current round $r$}
                    \State updated $f \gets$ feedback of groups containing $\Pi$ in minibatch of $D_{feed}$
                \EndIf
            \EndIf
        \EndFor
    \EndFor
    
    \State \textbf{Step 5:} Eliminate individual
    \If{$|P_i| > N_{\text{max}}$}
        \State $(\Pi^{\text{weak}}, s_{\Pi^{\text{weak}}}, f^{\text{weak}}) \gets \underset{(\Pi, s_{\Pi,\text{EMA}}, f)}{\operatorname{argmin}} s_{\Pi,\text{EMA}}$
        \State $P_i \gets P_i \setminus \{(\Pi^{\text{weak}}, s_{\Pi^{\text{waek}},\text{EMA}}, f^{\text{weak}})\}$
    \EndIf
\EndFor

\State \Return $\{P_i\}_{i=1}^{|\mathcal{G}|}$
\end{algorithmic}
\end{algorithm}

\section{BenchMark Datasets}\label{app:benchmarks}
\subsection{Open-ended Tasks}
\textbf{HotpotQA}~\cite{yang2018hotpotqadatasetdiverseexplainable} is a large-scale Wikipedia-based benchmark containing 113k question-answer pairs, where each question requires reasoning over multiple supporting documents to answer. Following \citet{tan2025langprobelanguageprogramsbenchmark}, we take the question as input, perform two rounds of retrieval and summary to find out the final answer to the question. We use the F1 Score to evaluate string equivalence between the predicted answer and the ground truth answer. 

\textbf{IFBench}~\cite{pyatkin2025generalizingverifiableinstructionfollowing} is a benchmark designed to evaluate the ability of language models to follow human instructions precisely. We adopt a two-stage response generation system consistent with GEPA’s \cite{agrawal2025gepa} framework: first produces a direct response to the human instruction, then refines this response to better conform with specified constraints. We use the accuracy of whether the final predicted answer follows all instructions as the evaluation metric. 
\subsection{Closed-ended Tasks}
\textbf{HoVer}~\cite{he2025hoverversatileneuralwholebody} is an open-domain benchmark for multi-hop fact extraction and claim verification. Following \citet{tan2025langprobelanguageprogramsbenchmark}, we take the claim as input, perform two rounds of summarization and query generation, followed by three rounds of retrieval, then return all retrieval supporting facts. We adopt a majority voting strategy, retaining only those facts supported by more than half of the islands as the final predictions. We retain all supporting facts by the islands as the final prediction results. 

\textbf{MATH}~\cite{lightman2023lets} is a challenging benchmark consisting of 12,500+ high-school and competition-level math problems across diverse topics. Given a math problem as input, we extract the final answer from the generated reasoning output and compute accuracy based on an exact string match with the ground-truth answer. Our voting strategy follows the most frequently predicted answer is selected if it constitutes a majority; otherwise, a final answer is randomly sampled from all candidate outputs. 

\textbf{GPQA}~\cite{rein2023gpqagraduatelevelgoogleproofqa} is a rigorously curated benchmark of multiple-choice questions at the graduate level in biology, physics, and chemistry. Given a question and its candidate options as input, we extract the final option from the generated reasoning output. Accuracy is computed based on an exact match with the ground-truth option. We then apply the same majority voting strategy as in MATH. 

\subsection{Dataset Split}
Following the protocol established in GEPA, we partition each of the five datasets into three disjoint subsets: a feedback set, a metric set, and a test set. The exact split sizes for each benchmark are summarized in Table~\ref{tab:dataset_partition}.
\begin{table}[H]
    \caption{Dataset split across five benchmarks. The feedback set guides evolution, the metric set measures evolution fitness, and the test set is held out for final evaluation.}
    \centering
    \begin{tabular}{cccc}\toprule
         &  feedback set&  metric set&  test set\\\midrule
         HotPotQA&  150&  300&  300\\
         IFBench&  150&  300&  294\\
         HoVer&  150&  300&  300\\
         MATH&  150&  300&  300\\
         GPQA&  100&  200&  264\\ \bottomrule
    \end{tabular}

    \label{tab:dataset_partition}
\end{table}

\section{Models} \label{app:models}
\textbf{Qwen3-8B:} Qwen3-8B is an open-source large language model containing 8.2 billion parameters. We deploy the model using SGLang ~\cite{zheng2024sglangefficientexecutionstructured} on 8×H800 GPUs, employing data parallelism for efficient parallel inference. We set the decoding temperature to 1 and the maximal generation length to 4096 during both evolution and evaluation.

\textbf{GPT4.1-mini:} GPT4.1-mini is a small-scale language model released by OpenAI in 2025. We access GPT4.1-mini through the OpenAI API, with the decoding temperature set to 1 and the max generation length set to 4096.

\section{Results}
\begin{figure}[H]
    \centering
    \includegraphics[width=1\linewidth]{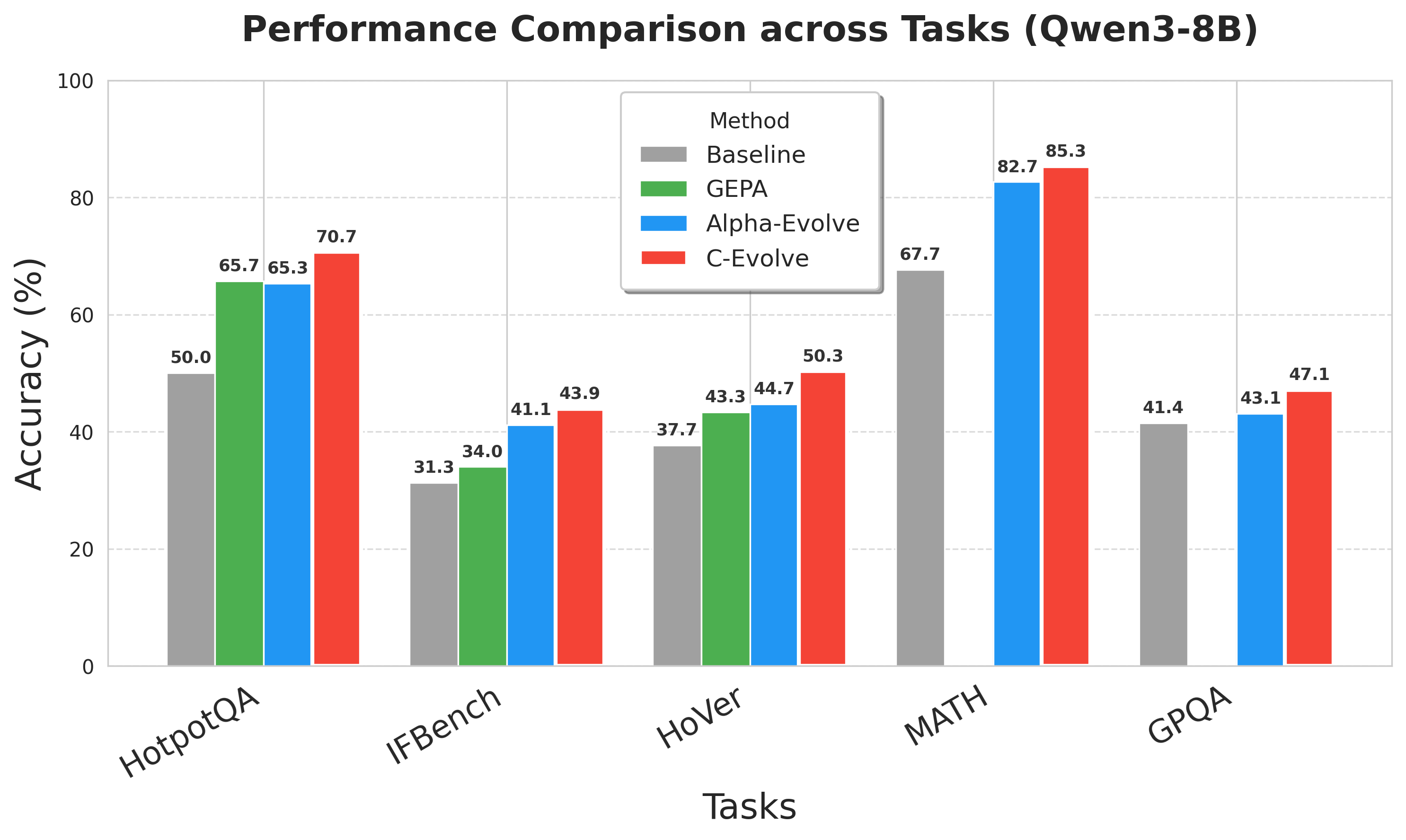}
    \caption{Final test set performance for C-Evolve and other evolutionary algorithms for Qwen3-8B.}
    \label{fig:app_result_qwen}
\end{figure}

\begin{figure}[H]
    \centering
    \includegraphics[width=1\linewidth]{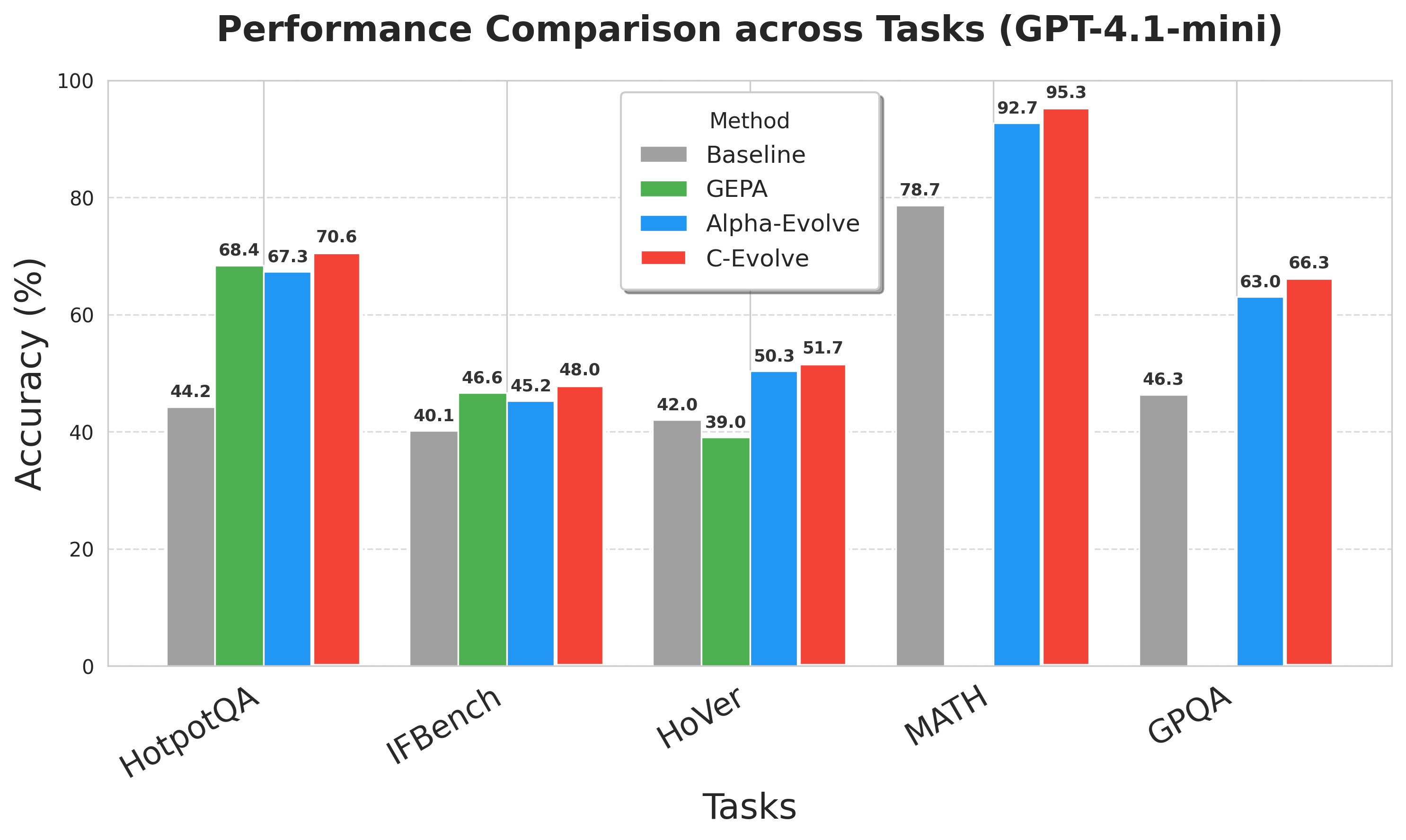}
    \caption{Final test set performance for C-Evolve and other evolutionary algorithms for GPT-4.1-mini.}
    \label{fig:app_result_gpt}
\end{figure}

Figure~\ref{fig:app_result_qwen} and Figure~\ref{fig:app_result_gpt} show the final test set performance for C-Evolve and other evolutionary algorithms for Qwen3-8B and GPT-4.1-mini.

\section{More Results for Section~\ref{sec:5.4.3}}\label{app:5.4.3}
\begin{figure}
    \centering
    \includegraphics[width=1\linewidth]{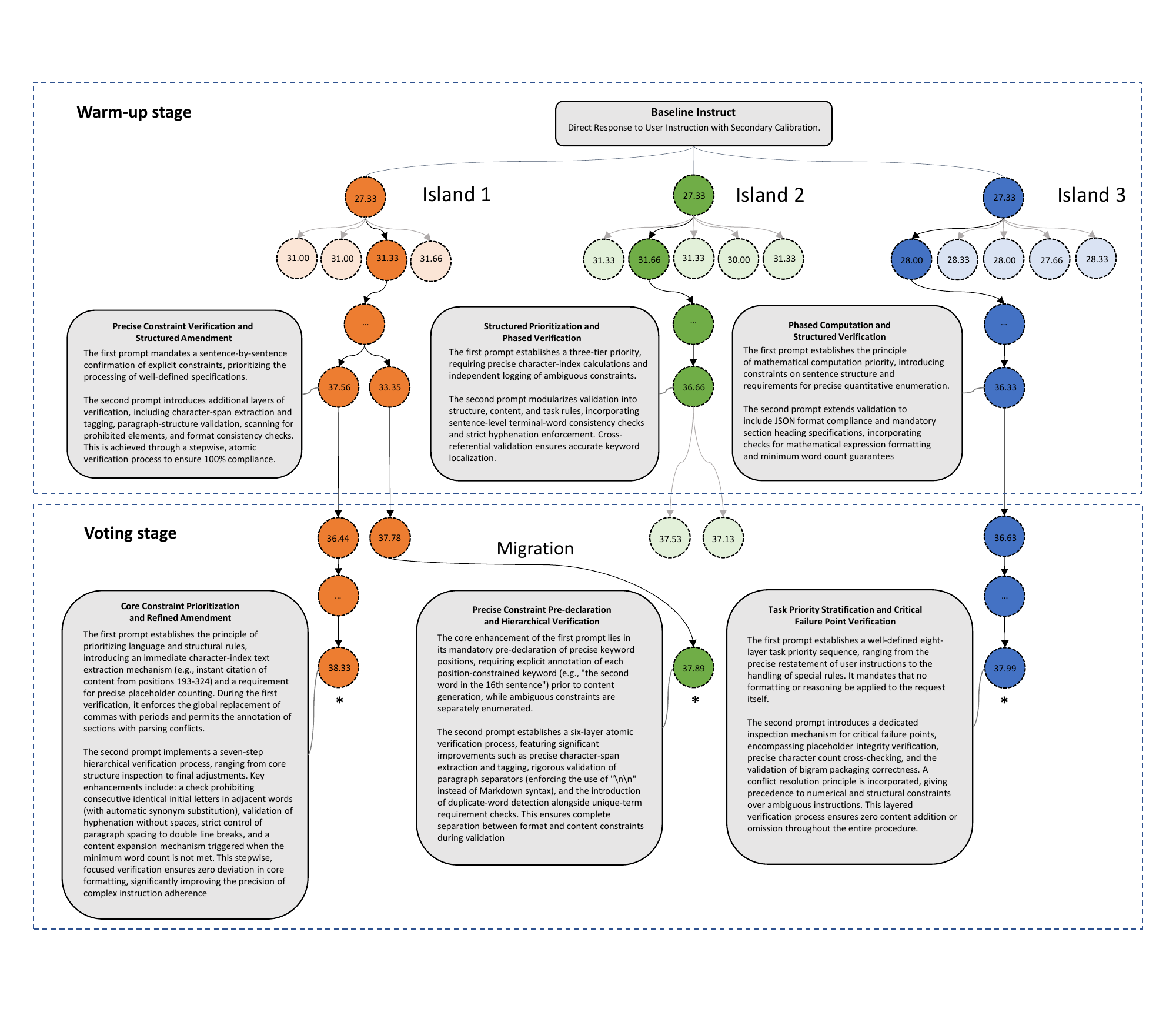}
    \caption{C-Evolve progressively incorporates the necessary considerations for instruction following through evolution, leading to performance improvement on the IFBench task.}
    \label{fig:experiment_diff_prompts}
\end{figure}

In Figure~\ref{fig:experiment_diff_prompts}, circles represent individuals, with colors denoting their island affiliation, and numerical values indicate their performance scores. During the warm-up stage, the three islands produced highly specialized individuals along distinct strategic dimensions: Island 1 yielded prompts focused on the depth and dynamic verification of processes; Island 2 evolved prompts emphasizing the structure and prioritization of rule systems; Island 3 generated prompts dedicated to underlying computational and extraction task support. In the voting stage, the individual migrated from island 1 to island 2 eventually became the top-performing individual within island 2 and evolved prompt details distinct from the best individual on Island 1. Finally, C-Evolve selected the highest-scoring individuals from each island (marked with asterisks) to form a cooperative group for actual inference tasks.

\begin{figure}
    \centering
    \includegraphics[width=1\linewidth]{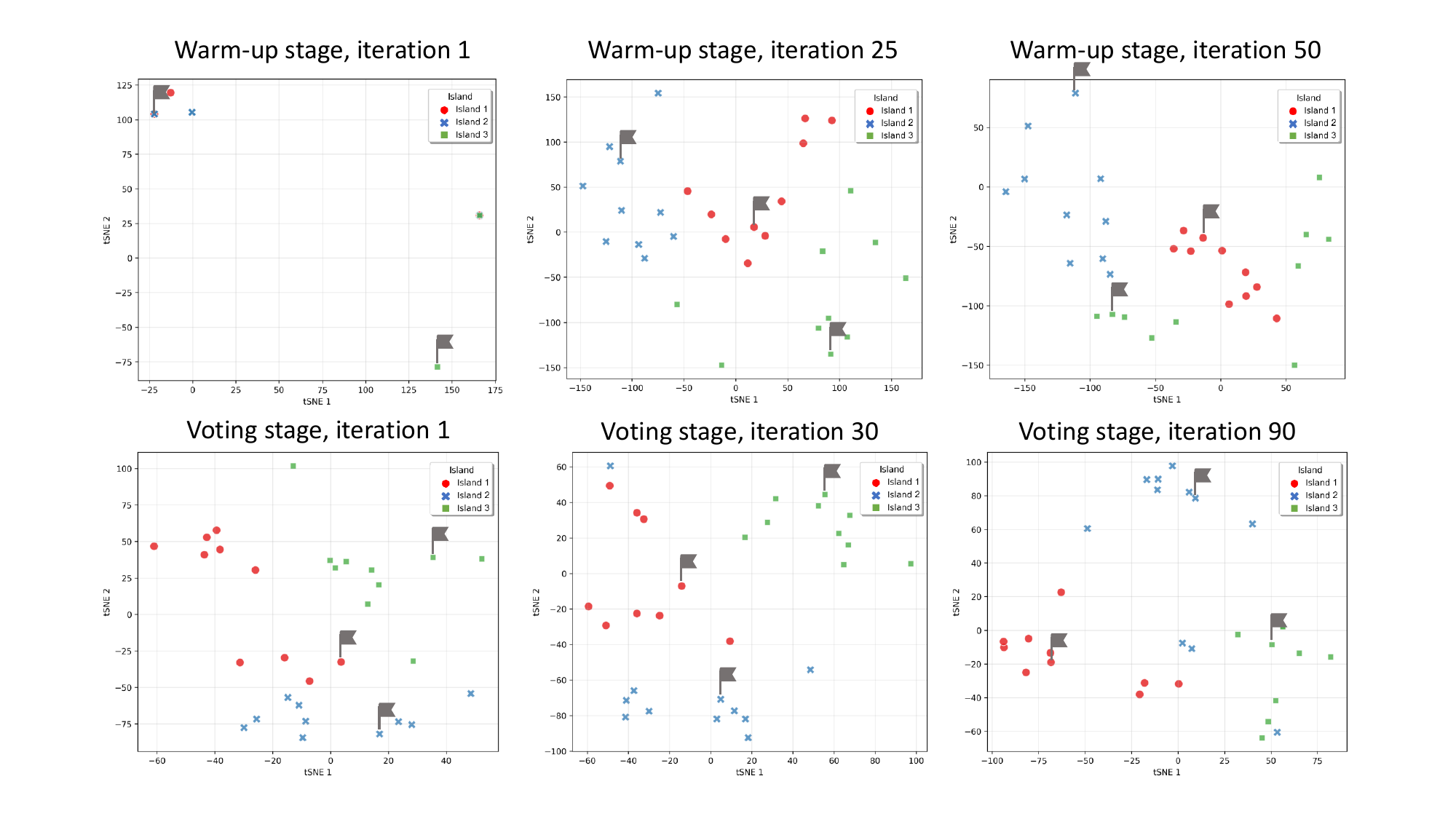}
    \caption{The distribution of individuals from different islands during both the warm-up and voting stages of C-evolve, visualized with t-SNE. The individual with the highest fitness score on each island is marked with a gray flag. Due to the presence of overlapping individuals across islands during the first iteration of the warm-up stage, only two flags are marked in this iteration.}
    \label{fig:experiment_reduce}
\end{figure}

In Figure~\ref{fig:experiment_reduce}, We utilized the TfidfVectorizertool from the Scikit-learn library \citep{scikit-learn} to extract feature embeddings from all the prompts contained within each individual, serving as individual characteristics. These features were then visualized using the t-SNE \citep{2008Visualizing} method for dimensionality reduction. During the warm-up stage, as the evolutionary process progressed, individuals from different islands gradually diverged from one another. In the voting stage, individuals from various islands did not lose their distinctive features despite engaging in consensus building. After 100 iterations within the voting stage, island 0 and island 2 gradually formed distinct clusters, indicating a continuous enhancement and strengthening of their respective island-specific characteristics.

\section{More Results for Section~\ref{sec:5.4.2}}\label{app:5.4.2}

\begin{table}[H]
    \centering
    \caption{Additional results of C-Evolve on the MATH Level 5 test set. We categorize problems into three cases: (1) All three prompts produce the same answer (Row 1), covering 56\% of the problems. Both single-prompt and consensus-building predictions achieve 95.23\% accuracy; (2) Two prompts produce the same answer, covering 25\% of the problems. In this case, majority voting improves accuracy by 9.33\% over the best single prompt; (3) All prompts produce distinct answers (19\% of the problems), and one answer is randomly selected as the group prediction.}
    \begin{tabular}{lccc}\toprule
        Case & Ratio & Single Prompt Accuracy& Consensus-building Accuracy\\\midrule
        All answers the same & 56\% & 95.23\% & 95.23\%\\
        two of three answers the same & 25\% & 40\% & \textbf{49.33\%}\\
        All answers distinct & 19\% & 14.03\% & 17.54\%\\ \bottomrule
    \end{tabular}
    \label{tab:app_math}
\end{table}
\section{More Results for Ablation Study} \label{app:ablation_study}
\subsection{Warm-up Stage}
We evaluate the impact of the warm-up stage in C-Evolve on IFBench. Each experiment run for a total of 100 iterations: when the warm-up stage is enabled, it comprises 50 iterations of warm-up followed by 50 iterations in the voting stage (in Figure~\ref{fig:ablation_figure} (a). As illustrated in Table~\ref{tab:ablation_warmup}, using warm-up allows the prompt group to achieve better performance (\textbf{42.85}) than directly starting the voting stage.

\begin{table}[H]
    \centering
    \caption{Ablation study on whether to use Warm-up stage.}
    \begin{tabular}{cc}\toprule
         Warm-up Stage&  Accuracy\\\midrule
         w/o&  41.83\\
         w/&  \textbf{42.85}\\ \bottomrule
    \end{tabular}

    \label{tab:ablation_warmup}
\end{table}
\subsection{voting score}
In this section, we compare the voting score and the max score. The max score is defined as the highest score for an individual across multiple groups, rather than taking the average as in Equation~\ref{eq:cooperation_score}. 

Experiments are conducted on IFBench using the Qwen3-8B model. Both approaches start from the same warm-up stage and are trained 50 iterations for the voting stage. As shown in Table~\ref{tab:ablation_score}, the max score underperforms compared to the voting score. This is because the max score merely picks the highest-scoring group in the metric set, which generalizes poorly. 
\begin{table}[H]
    \centering
    \caption{Ablation Study on max score and voting score}
    \begin{tabular}{cc}\toprule
         Score &  Accuracy\\\midrule
         Max Score      &  38.44      \\
         Voting Score&  \textbf{42.85}\\ \bottomrule
    \end{tabular}

    \label{tab:ablation_score}
\end{table}
\subsection{Other Results for ablation study}
\begin{figure}[H]
    \centering
    \includegraphics[width=1\linewidth]{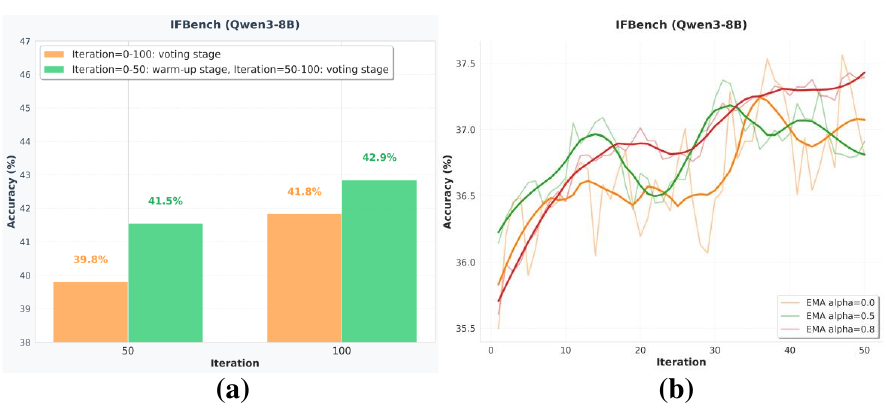}
    \caption{(a) Comparison between Evolve with (\textcolor{orange}{yellow}) or without (\textcolor{green}{green}) the warmup stage. We select the individuals with the highest fitness to form a group at different iterations, and evaluate them on the test set of IFBench. (b) The curve of the highest EMA voting score during the voting stage (yellow line: $\alpha=0$, green line: $\alpha=0.5$, red line: $\alpha=0.8$). Employing EMA with $\alpha=0.8$ produces a more stable evolution trajectory, and also achieves better performance after 50 iterations.}
    \label{fig:ablation_figure}
\end{figure}

In this section, we provide comparison between C-Evolve with and without the warm-up stage in Figure~\ref{fig:ablation_figure} (a), and illustrate the highest EMA voting score during the voting stage using different EMA $\alpha$: 0.5, 0.8, and 0.0 in Figure~\ref{fig:ablation_figure} (b).

\section{C-Evolve's Prompts in Evolution}
Here, we provide the system prompt for evolving task prompts used in compound AI systems with search and replace blocks. Placeholder text is shown in \textcolor{blue}{blue}.
\label{app:evolver}

\begin{tcolorbox}[
    colback=promptbg,
    colframe=gray!70!black,
    sharp corners=south,
    boxrule=0.5mm,
    fonttitle=\bfseries\large,
    title={Prompt for evolving a new prompt using search and replace},
    width=\textwidth,
    enhanced,
    drop shadow={black!50!white},
    top=8pt, bottom=8pt, left=10pt, right=10pt,
    arc=3pt,
]
\scriptsize
\setlength{\parskip}{0pt}
\setlength{\itemsep}{0pt}
You should write prompts to accomplish the following task:

\texttt{<task\_description>} \\
\parbox{\dimexpr\linewidth-2\fboxsep}{\ttfamily
\textcolor{blue}{{task\_description} as described in Appendix~\ref{app:task_description}.}
} \\
\texttt{</task\_description>}

\vspace{4pt}
Here is a previous version:

\texttt{<prompt\_candidate>} \\
\texttt{<prompt>} \\
\parbox{\dimexpr\linewidth-2\fboxsep}{\ttfamily
{\textcolor{blue}{the system prompt to evolve}}
} \\
\texttt{</prompt>} \\
\texttt{<eval\_result>} \\
metrics: 
\parbox{\dimexpr\linewidth-2\fboxsep}{\ttfamily
\textcolor{blue}{individual\_score (warm-up stage) or EMA\_voting\_score (voting stage)}
} \\
feedback:
\parbox{\dimexpr\linewidth-2\fboxsep}{\ttfamily
\textcolor{blue}{feedback as in Appendix~\ref{fig:Feedback}.}
} \\
\texttt{</eval\_result>} \\
\texttt{</prompt\_candidate>}

\vspace{6pt}
To achieve the task, please propose modifications to the \texttt{prompt\_candidate}. Describe each change with a SEARCH/REPLACE block described as following:

\begin{verbatim}
<<<<<<< SEARCH
[Original lines]
=======
[Modified lines]
>>>>>>> REPLACE
\end{verbatim}

\end{tcolorbox}

\section{Task Description for each Task}\label{app:task_description}
Here, we provide the task description used in Appendix~\ref{app:evolver} for each task.

\begin{tcolorbox}[
    colback=task1,
    colframe=task1!50!black,
    boxrule=0.3mm,
    sharp corners,
    width=\linewidth,
    title={1. HotpotQA},
    fontupper=\small,
    left=6pt, right=6pt, top=4pt, bottom=4pt,
    before upper=\vspace{2pt},
    after upper=\vspace{2pt},
    breakable
]
Your current task is to optimize the system prompt so that it can guide the large language model to produce more accurate, stable, and controllable output in a six-step process.

\textbf{Goal:} Given an input question, perform two rounds of retrieval and summary, such that the final generated answer must \textbf{exactly match} the true answer.

\textbf{Task Pipeline:}
\begin{enumerate}[leftmargin=*,label=\textbf{Step \arabic*:},itemsep=4pt]
    \item \textbf{First Retrieval}
    \begin{itemize}[leftmargin=12pt,itemsep=2pt]
        \item \textit{Input:} 'question'
        \item \textit{Output:} A batch of related passages retrieved from the database (first retrieval passages)
    \end{itemize}
    
    \item \textbf{First Summary} (\texttt{<system\_prompt\_1></system\_prompt\_1>})
    \begin{itemize}[leftmargin=12pt,itemsep=2pt]
        \item \textit{Input:} 'question', first retrieval 'passages', 'system prompt 1'
        \item \textit{Output:} A concise, coherent summary of the first retrieval passages (summary\_1)
    \end{itemize}
    
    \item \textbf{Query Generation} (\texttt{<system\_prompt\_2></system\_prompt\_2>})
    \begin{itemize}[leftmargin=12pt,itemsep=2pt]
        \item \textit{Input:} 'question', 'summary\_1', 'system prompt 2'
        \item \textit{Output:} A query that covers the core semantics of the question for second retrieval
    \end{itemize}
    
    \item \textbf{Second Retrieval}
    \begin{itemize}[leftmargin=12pt,itemsep=2pt]
        \item \textit{Input:} The query from Step 3
        \item \textit{Output:} Another batch of related passages (second retrieval passages)
    \end{itemize}
    
    \item \textbf{Second Summary} (\texttt{<system\_prompt\_3></system\_prompt\_3>})
    \begin{itemize}[leftmargin=12pt,itemsep=2pt]
        \item \textit{Input:} 'question', first summary 'context', second retrieval 'passages', 'system prompt 3'
        \item \textit{Output:} A concise, coherent summary of the second retrieval passages (summary\_2)
    \end{itemize}
    
    \item \textbf{Answer Generation} (\texttt{<system\_prompt\_4></system\_prompt\_4>})
    \begin{itemize}[leftmargin=12pt,itemsep=2pt]
        \item \textit{Input:} 'question', 'summary\_1', 'summary\_2', 'system prompt 4'
        \item \textit{Output:} The final answer to the question
    \end{itemize}
\end{enumerate}

\textbf{Your Task:} Optimize 'system prompt 1', 'system prompt 2', 'system prompt 3', and 'system prompt 4' to provide clearer and more effective guidance for each corresponding step. Maximize the pipeline's overall accuracy, ensuring the final answer \textbf{strictly matches} the true answer.

\textbf{Constraints:}
\begin{itemize}[leftmargin=*,itemsep=3pt]
    \item You may perform multiple modifications
    \item Cannot modify content inside angle brackets (e.g., \texttt{<system\_prompt\_1>})
    \item Final output must contain exactly four system prompts
    \item Each prompt must be presented in the format: \texttt{<system\_prompt\_x> ... </system\_prompt\_x>}
\end{itemize}
\end{tcolorbox}

\begin{tcolorbox}[
    colback=task2,
    colframe=task2!50!black,
    boxrule=0.3mm,
    sharp corners,
    title={2. IFBench},
    width=\linewidth,
    fontupper=\small,
    left=6pt, right=6pt, top=4pt, bottom=4pt,
    before upper=\vspace{2pt},
    after upper=\vspace{2pt},
]
Your task is to optimize the system prompt for IFBench so that the large language model can more accurately follow the various requirements in the instructions during a two-step generation process.

\textbf{Goal:} Given an instruction, first answer it directly, and then further calibrate the answer to produce a final response that best meets the instruction's requirements.

\textbf{Task Pipeline:}
\begin{enumerate}[leftmargin=*,label=\textbf{Step \arabic*:},itemsep=4pt]
    \item \textbf{Direct Answer Generation} (\texttt{<system\_prompt\_1></system\_prompt\_1>})
    \begin{itemize}[leftmargin=12pt,itemsep=2pt]
        \item \textit{Input:} User's instruction + \texttt{system\_prompt\_1}
        \item \textit{Output:} A direct response to the instruction
    \end{itemize}
    
    \item \textbf{Answer Calibration} (\texttt{<system\_prompt\_2></system\_prompt\_2>})
    \begin{itemize}[leftmargin=12pt,itemsep=2pt]
        \item \textit{Input:} User's instruction + Step 1 answer + \texttt{system\_prompt\_2}
        \item \textit{Output:} Final calibrated answer
    \end{itemize}
\end{enumerate}

\textbf{Your Task:}
Refine system prompt 1 and system prompt 2 so that each one provides clearer and more effective guidance for its corresponding step in the pipeline. The ultimate goal is to maximize the model's ability to follow instructions, ensuring its responses meet all requirements in the user's instructions.

\textbf{Constraints:}
\begin{itemize}[leftmargin=*,itemsep=3pt]
    \item You may perform multiple modifications
    \item Cannot modify content inside angle brackets (e.g., \texttt{<system\_prompt\_1>})
    \item Final output must contain exactly two system prompts
    \item Each prompt must be presented in the format: \texttt{<system\_prompt\_x> ... </system\_prompt\_x>}
\end{itemize}
\end{tcolorbox}
\begin{tcolorbox}[
    colback=task3,
    colframe=task3!50!black,
    boxrule=0.3mm,
    sharp corners,
    title={3. HoVer},
    width=\linewidth,
    fontupper=\small,
    left=6pt, right=6pt, top=4pt, bottom=4pt,
    before upper=\vspace{2pt},
    after upper=\vspace{2pt},
    breakable
]
Your task is to optimize four system prompts for a seven-step retrieval-and-answer pipeline.

\textbf{Goal:} Given an input claim, perform two rounds of summarization and query generation, followed by three rounds of retrieval, to determine whether the claim can be SUPPORTED or NOT\_SUPPORTED by the final passages. The evaluation criterion is whether the true passages are fully contained within the union of the three retrieval outputs (first, second, and final retrieval).

\textbf{Task Pipeline:}
\begin{enumerate}[leftmargin=*,label=\textbf{Step \arabic*:},itemsep=4pt]
    \item \textbf{First Retrieval}
    \begin{itemize}[leftmargin=12pt,itemsep=2pt]
        \item \textit{Input:} claim
        \item \textit{Output:} Initial batch of related passages (first retrieval passages)
    \end{itemize}
    
    \item \textbf{First Summary} (\texttt{<system\_prompt\_1></system\_prompt\_1>})
    \begin{itemize}[leftmargin=12pt,itemsep=2pt]
        \item \textit{Input:} claim, first retrieval passages, \texttt{system\_prompt\_1}
        \item \textit{Output:} Concise, coherent summary (summary\_1)
    \end{itemize}
    
    \item \textbf{First Query Generation} (\texttt{<system\_prompt\_2></system\_prompt\_2>})
    \begin{itemize}[leftmargin=12pt,itemsep=2pt]
        \item \textit{Input:} claim, summary\_1, \texttt{system\_prompt\_2}
        \item \textit{Output:} Query capturing core claim semantics for second retrieval
    \end{itemize}
    
    \item \textbf{Second Retrieval}
    \begin{itemize}[leftmargin=12pt,itemsep=2pt]
        \item \textit{Input:} Query from Step 3
        \item \textit{Output:} Second batch of related passages (second retrieval passages)
    \end{itemize}
    
    \item \textbf{Second Summary} (\texttt{<system\_prompt\_3></system\_prompt\_3>})
    \begin{itemize}[leftmargin=12pt,itemsep=2pt]
        \item \textit{Input:} claim, first summary (context), second retrieval passages, \texttt{system\_prompt\_3}
        \item \textit{Output:} Concise, coherent summary (summary\_2)
    \end{itemize}
    
    \item \textbf{Second Query Generation} (\texttt{<system\_prompt\_4></system\_prompt\_4>})
    \begin{itemize}[leftmargin=12pt,itemsep=2pt]
        \item \textit{Input:} claim, summary\_1, summary\_2, \texttt{system\_prompt\_4}
        \item \textit{Output:} Query capturing core claim semantics for final retrieval
    \end{itemize}
    
    \item \textbf{Final Retrieval}
    \begin{itemize}[leftmargin=12pt,itemsep=2pt]
        \item \textit{Input:} Query from Step 6
        \item \textit{Output:} Final batch of related passages (third retrieval passages)
    \end{itemize}
\end{enumerate}

\textbf{Your Task:}
Optimize system prompt 1, system prompt 2, system prompt 3, and system prompt 4 to provide clearer, more effective guidance for their corresponding steps. Ensure overall accuracy is maximized, so that all true passages are contained in the final three retrieval results.

\textbf{Constraints:}
\begin{itemize}[leftmargin=*,itemsep=3pt]
    \item You may perform multiple modifications
    \item Cannot modify content inside angle brackets (e.g., \texttt{<system\_prompt\_1>})
    \item Final output must contain exactly four system prompts
    \item Each prompt must be presented in the format: \texttt{<system\_prompt\_x> ... </system\_prompt\_x>}
\end{itemize}
\end{tcolorbox}

\begin{tcolorbox}[
    colback=task4,
    colframe=task4!50!black,
    boxrule=0.3mm,
    sharp corners,
    title={4. MATH},
    width=\linewidth,
    fontupper=\small,
    left=6pt, right=6pt, top=4pt, bottom=4pt,
    before upper=\vspace{2pt},
    after upper=\vspace{2pt},
    breakable
]
Your task is to optimize a system prompt for mathematical problems.

\textbf{Goal:} Given an input math problem, answer this problem to get the final answer.

\textbf{Task Pipeline:}
\begin{enumerate}[leftmargin=*,label=\textbf{Step \arabic*:},itemsep=4pt]
    \item \textbf{Answer}
    \begin{itemize}[leftmargin=12pt,itemsep=2pt]
        \item \textit{Input:} problem
        \item \textit{Output:} the answer, and put the final answer in \verb|\boxed{}| tag, i.e. \$\verb|\boxed{}|\{answer here\}.
    \end{itemize}
\end{enumerate}

\textbf{Your Task:}
Optimize system prompt 1 to provide clearer, more effective guidance for their corresponding steps. Ensure overall accuracy is maximized, so that the predicted answer can exactly match the true answer.

\textbf{Constraints:}
\begin{itemize}[leftmargin=*,itemsep=3pt]
    \item You may perform multiple modifications
    \item Cannot modify content inside angle brackets (e.g., \texttt{<system\_prompt\_1>})
    \item Final output must contain exactly one system prompt
    \item Each prompt must be presented in the format: \texttt{<system\_prompt\_x> ... </system\_prompt\_x>}
\end{itemize}
\end{tcolorbox}

\begin{tcolorbox}[
    colback=task5,
    colframe=task5!50!black,
    boxrule=0.3mm,
    sharp corners,
    title={5. GPQA},
    width=\linewidth,
    fontupper=\small,
    left=6pt, right=6pt, top=4pt, bottom=4pt,
    before upper=\vspace{2pt},
    after upper=\vspace{2pt},
    breakable
]
Your task is to optimize a system prompt for Multiple-choice questions.

\textbf{Goal:} Given an input Multiple-choice question, answer this question to get the final answer.

\textbf{Task Pipeline:}
\begin{enumerate}[leftmargin=*,label=\textbf{Step \arabic*:},itemsep=4pt]
    \item \textbf{Answer}
    \begin{itemize}[leftmargin=12pt,itemsep=2pt]
        \item \textit{Input:} question
        \item \textit{Output:} the answer, and put the final answer in \verb|\boxed{}| tag, i.e. \$\verb|\boxed{}|\{answer here\}.
    \end{itemize}
\end{enumerate}

\textbf{Your Task:}
Optimize system prompt 1 to provide clearer, more effective guidance for their corresponding steps. Ensure overall accuracy is maximized, so that the predicted answer can exactly match the true answer.

\textbf{Constraints:}
\begin{itemize}[leftmargin=*,itemsep=3pt]
    \item You may perform multiple modifications
    \item Cannot modify content inside angle brackets (e.g., \texttt{<system\_prompt\_1>})
    \item Final output must contain exactly one system prompt
    \item Each prompt must be presented in the format: \texttt{<system\_prompt\_x> ... </system\_prompt\_x>}
\end{itemize}
\end{tcolorbox}

\section{Feedback for each Task}
\label{fig:Feedback}
Here we elaborate on how we compile the feedback for different tasks. Placeholder text is shown in \textcolor{blue}{blue}.


\paragraph{HotPotQA} We compile the execution traces, including the retrieved docs and the summary in each hop, to obtain the feedback. Due to that the retrieved documents are usually long, we use an LLM to compile feedbacks on multiple questions used in the evolver LLM, based on the execution details. The field ``score" refer to the F1 score calculated by comparing ``answer" with the ground truth answer.

\begin{tcolorbox}[
    colback=task1,
    colframe=task1!50!black,
    boxrule=0.3mm,
    sharp corners,
    width=\linewidth,
    title={Prompt for feedback compilation with LLM in the warm-up stage on HotPotQA},
    fontupper=\small,
    left=6pt, right=6pt, top=4pt, bottom=4pt,
    before upper=\vspace{2pt},
    after upper=\vspace{2pt},
    breakable,
]
\textbf{Your Task:} Analyze \textbf{this system prompt}: \textcolor{blue}{$system\_prompt$}.

\textbf{Goal of This System Prompt:} Given an input question, perform two rounds of retrieval and summary, such that the final generated answer must exactly match the true answer.

Below are the results of two rounds of retrieval and summary on multiple questions. Please analyze them one by one:

\vspace{6pt}
\textbf{Question 1:}
\begin{itemize}[leftmargin=12pt,itemsep=4pt]
    \item \textit{Question:} \textcolor{blue}{$question$}
    
    \item \textit{This system prompt's result:}
    \begin{lstlisting}
{
    "hop1_docs": hop1_docs,
    "summary_1": summary_1,
    "hop2_query": hop2_query,
    "hop2_docs": hop2_docs,
    "summary_2": summary_2,
    "answer": answer,
    "score" : score
}
\end{lstlisting}
    
    \item \textit{True answer:}

\textcolor{blue}{grounded\_truth\_answer}

\end{itemize}

\textcolor{blue}{[... Additional questions ...]}

\vspace{6pt}
Carefully reflect on the following:
\begin{enumerate}[leftmargin=*,itemsep=4pt]
    \item For each question, what specific weaknesses does this system prompt show?
    \item Across all questions, what common issues or patterns can you identify?
    \item How could this system prompt be revised or improved?
\end{enumerate}
    
\end{tcolorbox}
\begin{tcolorbox}[
    colback=task1,
    colframe=task1!50!black,
    boxrule=0.3mm,
    sharp corners,
    width=\linewidth,
    title={Prompt for group feedback compilation with LLM in the voting stage on HotPotQA},
    fontupper=\small,
    left=6pt, right=6pt, top=4pt, bottom=4pt,
    before upper=\vspace{2pt},
    after upper=\vspace{2pt},
    breakable,
]
The current system prompt works together with other system prompts, and their outputs are combined through majority voting to form the final result.

\textbf{Your Task:} Analyze \textbf{only this system prompt}: \textcolor{blue}{$system\_prompt$}, without evaluating the other two.

\textbf{Goal of This System Prompt:} Given an input question, perform two rounds of retrieval and summary, such that the final generated answer must exactly match the true answer.

Below are the results of two rounds of retrieval and summary on multiple questions. Please analyze them one by one:

\vspace{6pt}
\textbf{Question 1:}
\begin{itemize}[leftmargin=12pt,itemsep=4pt]
    \item \textit{Question:} \textcolor{blue}{$question$}
    
    \item \textit{This system prompt's result:}
    \begin{lstlisting}
{
    "hop1_docs": hop1_docs,
    "summary_1": summary_1,
    "hop2_query": hop2_query,
    "hop2_docs": hop2_docs,
    "summary_2": summary_2,
    "answer": answer,
    "score": score
}
\end{lstlisting}
    
    \item \textit{Other system prompts' result:}
    \begin{lstlisting}
System Prompt 1:
{
    "hop1_docs": hop1_docs,
    "summary_1": summary_1,
    "hop2_query": hop2_query,
    "hop2_docs": hop2_docs,
    "summary_2": summary_2,
    "answer": answer,
    "score" : score
}
    \end{lstlisting}
    \textcolor{blue}{[... Additional system prompts...]}
    \item \textit{Final result after majority voting:} 

\textcolor{blue}{group\_final\_answer}

    \item \textit{True answer:}

\textcolor{blue}{grounded\_truth\_answer}

\end{itemize}

\textcolor{blue}{[... Additional questions ...]}

\vspace{6pt}
Carefully reflect on the following:
\begin{enumerate}[leftmargin=*,itemsep=4pt]
    \item For each question, what specific weaknesses does this system prompt show?
    \item Across all questions, what common issues or patterns can you identify?
    \item How could this system prompt be revised or improved so that, when combined with others, it contributes more effectively to better final results?
\end{enumerate}
    
\end{tcolorbox}

\paragraph{IFBench} We directly organize the instructions, as well as the answers and final answers generated by the compound AI system in the following format to compile the feedback. The field ``result" indicates the detailed information of requirements and whether ``final\_answer" successfully follows each requirement. The field ``score" indicates whether ``final\_answer" successfully follows all the requirements in this input instruction (0 or 1).

\begin{tcolorbox}[
    colback=task2,
    colframe=task2!50!black,
    boxrule=0.3mm,
    sharp corners,
    width=\linewidth,
    title={An example of compiled feedback in the warm-up stage on IFBench},
    fontupper=\small,
    left=6pt, right=6pt, top=4pt, bottom=4pt,
    before upper=\vspace{2pt},
    after upper=\vspace{2pt},
    breakable,
]

\vspace{6pt}
Below are the results of the answers and the final answers on multiple questions.

\textbf{Instruction 1:}
\begin{itemize}[leftmargin=12pt,itemsep=2pt]
    \item \textit{Instruction:} \textcolor{blue}{$Instruction$}
    \item \textit{This system prompt’s result:} \begin{lstlisting}
{
    "answer": answer,
    "final_answer": final_answer,
    "result" : result,
    "score": score
}
    \end{lstlisting}
    
\end{itemize}

\vspace{4pt}

\textcolor{blue}{[... Additional instruction ...]}
\end{tcolorbox}
\begin{tcolorbox}[
    colback=task2,
    colframe=task2!50!black,
    boxrule=0.3mm,
    sharp corners,
    width=\linewidth,
    title={An example of compiled group feedback in the voting stage on IFBench},
    fontupper=\small,
    left=6pt, right=6pt, top=4pt, bottom=4pt,
    before upper=\vspace{2pt},
    after upper=\vspace{2pt},
    breakable,
]
The current system prompt works together with other system prompts, and their outputs are combined through majority voting to form the final result.

\vspace{6pt}
Below are the results of the answers and the final answers on multiple questions.

\textbf{Instruction 1:}
\begin{itemize}[leftmargin=12pt,itemsep=2pt]
    \item \textit{Instruction:} \textcolor{blue}{$Instruction$}
    \item \textit{This system prompt’s result:} \begin{lstlisting}
{
    "answer": answer,
    "final_answer": final_answer,
    "result" : result,
    "score": score,
}
    \end{lstlisting}
    \item \textit{Other system prompts' result:}
    \begin{lstlisting}
System Prompt 1:
{
    "answer": answer,
    "final_answer": final_answer,
    "result" : result,
    "score": score,
}
    \end{lstlisting}
    \textcolor{blue}{[... Additional system prompts...]}
    \item \textit{Final result after majority voting:} 

\textcolor{blue}{group\_final\_answer}

\end{itemize}

\textcolor{blue}{[... Additional instruction ...]}
\end{tcolorbox}

\paragraph{HoVer} We compile the execution traces, including the retrieved docs and the summary in each hop, to obtain the feedback. Due to that the retrieved documents are usually long, we use an LLM to compile feedbacks on multiple questions used in the evolver LLM, based on the execution details. The field ``score" indicates whether all documents listed in ground truth are successfully retrieved during execution (0 or 1) for the given claim.
\begin{tcolorbox}[
    colback=task3,
    colframe=task3!50!black,
    boxrule=0.3mm,
    sharp corners,
    width=\linewidth,
    title={Prompt for feedback compilation with LLM in the warm-up stage on HoVer},
    fontupper=\small,
    left=6pt, right=6pt, top=4pt, bottom=4pt,
    before upper=\vspace{2pt},
    after upper=\vspace{2pt},
    breakable,
]
\textbf{Your Task:} Analyze \textbf{this system prompt}: \textcolor{blue}{$system\_prompt$}.

\textbf{Goal of This System Prompt:} Given an input claim, perform two rounds of retrieval and summarization
and determine whether the "true answer" are a subset of the "answer".

Below are the results of applying this system prompt to multiple claims. Please analyze them one by one:

\vspace{6pt}
\textbf{Claim 1:}
\begin{itemize}[leftmargin=12pt,itemsep=4pt]
    \item \textit{Claim:} \textcolor{blue}{$claim$}
    
    \item \textit{This system prompt's result:}
    \begin{lstlisting}
{
    "hop1_docs": hop1_docs,
    "summary_1": summary_1,
    "hop2_query": hop2_query,
    "hop2_docs": hop2_docs,
    "summary_2": summary_2,
    "hop3_query": hop3_query,
    "hop3_docs": hop3_docs,
    "answer": hop1_docs+hop2_docs+hop3_docs,
    "score" : score
    }
    \end{lstlisting}
    
    \item \textit{True answer:}

\textcolor{blue}{grounded\_truth\_answer}

\end{itemize}

\textcolor{blue}{[... Additional claims ...]}

\vspace{6pt}
Carefully reflect on the following:
\begin{enumerate}[leftmargin=*,itemsep=4pt]
    \item For each claim, what specific weaknesses does this system prompt show?
    \item Across all claims, what common issues or patterns can you identify?
    \item How could this system prompt be revised or improved?
\end{enumerate}
    
\end{tcolorbox}
\begin{tcolorbox}[
    colback=task3,
    colframe=task3!50!black,
    boxrule=0.3mm,
    sharp corners,
    width=\linewidth,
    title={Prompt for group feedback compilation with LLM in the voting stage on HoVer},
    fontupper=\small,
    left=6pt, right=6pt, top=4pt, bottom=4pt,
    before upper=\vspace{2pt},
    after upper=\vspace{2pt},
    breakable,
]
The current system prompt works together with other system prompts, and their outputs are combined through majority voting to form the final result.

\textbf{Your Task:} Analyze \textbf{only this system prompt}: \textcolor{blue}{$system\_prompt$}, without evaluating the other two.

\textbf{Goal of This System Prompt:} Given an input claim, perform two rounds of retrieval and summarization
and determine whether the "true answer" are a subset of the "answer".

Below are the results of applying this system prompt to multiple claims. Please analyze them one by one:

\vspace{6pt}
\textbf{Claim 1:}
\begin{itemize}[leftmargin=12pt,itemsep=4pt]
    \item \textit{Claim:} \textcolor{blue}{$claim$}
    
    \item \textit{This system prompt's result:}
    \begin{lstlisting}
{
    "hop1_docs": hop1_docs,
    "summary_1": summary_1,
    "hop2_query": hop2_query,
    "hop2_docs": hop2_docs,
    "summary_2": summary_2,
    "hop3_query": hop3_query,
    "hop3_docs": hop3_docs,
    "answer": hop1_docs+hop2_docs+hop3_docs,
    "score" : score
    }
    \end{lstlisting}
    
    \item \textit{Other system prompts' result:}
    \begin{lstlisting}
System Prompt 1:
{
    "hop1_docs": hop1_docs,
    "summary_1": summary_1,
    "hop2_query": hop2_query,
    "hop2_docs": hop2_docs,
    "summary_2": summary_2,
    "hop3_query": hop3_query,
    "hop3_docs": hop3_docs,
    "answer": hop1_docs+hop2_docs+hop3_docs,
    "score" : score
}
    \end{lstlisting}
    \textcolor{blue}{[... Additional system prompts...]}
    \item \textit{Final result after majority voting:} 

\textcolor{blue}{group\_final\_answer}

    \item \textit{True answer:}

\textcolor{blue}{grounded\_truth\_answer}

\end{itemize}

\textcolor{blue}{[... Additional claims ...]}

\vspace{6pt}
Carefully reflect on the following:
\begin{enumerate}[leftmargin=*,itemsep=4pt]
    \item For each claim, what specific weaknesses does this system prompt show?
    \item Across all claims, what common issues or patterns can you identify?
    \item How could this system prompt be revised or improved so that, when combined with others, it contributes more effectively to better final results?
\end{enumerate}
    
\end{tcolorbox}

\paragraph{MATH}
We directly organize the problems, as well as the answers generated by the prompt we evolved in the following format to compile the feedback. The ``extracted\_answer" refers to the answer extracted from the model response using the script of~\citet{hendrycks2021measuring}. The ``score" indicates whether the answer matches the ground truth (0 or 1).
\begin{tcolorbox}[
    colback=task4,
    colframe=task4!50!black,
    boxrule=0.3mm,
    sharp corners,
    width=\linewidth,
    title={An example of compiled feedback in the warm-up stage on MATH},
    fontupper=\small,
    left=6pt, right=6pt, top=4pt, bottom=4pt,
    before upper=\vspace{2pt},
    after upper=\vspace{2pt},
    breakable,
]
Below are the results of applying this system prompt to multiple problems. 

\vspace{6pt}
\textbf{Problem 1:}
\begin{itemize}[leftmargin=12pt,itemsep=4pt]
    \item \textit{Problem:} \textcolor{blue}{$problem$}
    
    \item \textit{This system prompt's result:}
    \begin{lstlisting}
{
    "answer": answer,
    "extracted answer": extracted_answer,
    "score" : score
    }
    \end{lstlisting}
    
    \item \textit{True answer:}

\textcolor{blue}{grounded\_truth\_answer}

\end{itemize}

\textcolor{blue}{[... Additional problems ...]}
    
\end{tcolorbox}
\begin{tcolorbox}[
    colback=task4,
    colframe=task4!50!black,
    boxrule=0.3mm,
    sharp corners,
    width=\linewidth,
    title={An example of the compiled group feedback in the voting stage on MATH},
    fontupper=\small,
    left=6pt, right=6pt, top=4pt, bottom=4pt,
    before upper=\vspace{2pt},
    after upper=\vspace{2pt},
    breakable,
]
The current system prompt works together with other system prompts, and their outputs are combined through majority voting to form the final result.

Below are the results of applying this system prompt to multiple problems. 

\vspace{6pt}
\textbf{Problem 1:}
\begin{itemize}[leftmargin=12pt,itemsep=4pt]
    \item \textit{Problem:} \textcolor{blue}{$problem$}
    
    \item \textit{This system prompt's result:}
    \begin{lstlisting}
{
    "answer": answer,
    "extracted answer": extracted_answer,
    "score" : score
    }
    \end{lstlisting}
    
    \item \textit{Other system prompts' result:}
    \begin{lstlisting}
System Prompt 1:
{
    "answer": answer,
    "extracted answer": extracted_answer,
    "score" : score
}
    \end{lstlisting}
    \textcolor{blue}{[... Additional system prompts...]}
    \item \textit{Final result after majority voting:} 

\textcolor{blue}{group\_final\_answer}

    \item \textit{True answer:}

\textcolor{blue}{grounded\_truth\_answer}

\end{itemize}

\textcolor{blue}{[... Additional problems ...]}

\end{tcolorbox}

\paragraph{GPQA}
We directly organize the problems, as well as the answers generated by the evolved prompt in the following format to compile the feedback. The ``extracted\_answer" refers to the chosen option (A, B, C or D) extracted from the generated answer. The ``score" indicates whether this choice matches the ground truth (0 or 1).
\begin{tcolorbox}[
    colback=task5,
    colframe=task5!50!black,
    boxrule=0.3mm,
    sharp corners,
    width=\linewidth,
    title={An example of compiled feedback in the warm-up stage on GPQA},
    fontupper=\small,
    left=6pt, right=6pt, top=4pt, bottom=4pt,
    before upper=\vspace{2pt},
    after upper=\vspace{2pt},
    breakable,
]
Below are the results of applying this system prompt to multiple questions. 

\vspace{6pt}
\textbf{Question 1:}
\begin{itemize}[leftmargin=12pt,itemsep=4pt]
    \item \textit{Question:} \textcolor{blue}{$question$}
    
    \item \textit{This system prompt's result:}
    \begin{lstlisting}
{
    "answer": answer,
    "extracted answer": extracted_answer,
    "score" : score
    }
    \end{lstlisting}
    
    \item \textit{True answer:}

\textcolor{blue}{grounded\_truth\_answer}

\end{itemize}

\textcolor{blue}{[... Additional questions ...]}
    
\end{tcolorbox}
\begin{tcolorbox}[
    colback=task5,
    colframe=task5!50!black,
    boxrule=0.3mm,
    sharp corners,
    width=\linewidth,
    title={An example of compiled group feedback in the voting stage on GPQA},
    fontupper=\small,
    left=6pt, right=6pt, top=4pt, bottom=4pt,
    before upper=\vspace{2pt},
    after upper=\vspace{2pt},
    breakable,
]
The current system prompt works together with other system prompts, and their outputs are combined through majority voting to form the final result.

Below are the results of applying this system prompt to multiple questions. 

\vspace{6pt}
\textbf{Question 1:}
\begin{itemize}[leftmargin=12pt,itemsep=4pt]
    \item \textit{Question:} \textcolor{blue}{$question$}
    
    \item \textit{This system prompt's result:}
    \begin{lstlisting}
{
    "answer": answer,
    "extracted answer": extracted_answer,
    "score" : score
    }
    \end{lstlisting}
    
    \item \textit{Other system prompts' result:}
    \begin{lstlisting}
System Prompt 1:
{
    "answer": answer,
    "extracted answer": extracted_answer,
    "score" : score
}
    \end{lstlisting}
    \textcolor{blue}{[... Additional system prompts...]}
    \item \textit{Final result after majority voting:} 

\textcolor{blue}{group\_final\_answer}

    \item \textit{True answer:}

\textcolor{blue}{grounded\_truth\_answer}

\end{itemize}

\textcolor{blue}{[... Additional questions ...]}

\end{tcolorbox}

\section{System Prompt for LLM-based Aggregator}\label{app:llm_based}
Here, we provide the system prompt for the LLM-based aggregator on open-ended tasks (HotPotQA and IFBench). Placeholder text is shown in \textcolor{blue}{blue}.

\begin{tcolorbox}[
    colback=promptbg,
    colframe=gray!70!black,
    sharp corners=south,
    boxrule=0.5mm,
    fonttitle=\bfseries\large,
    title={Prompt for LLM-based Aggregator (HotPotQA, IFBench)},
    width=\textwidth,
    enhanced,
    drop shadow={black!50!white},
    top=8pt, bottom=8pt, left=10pt, right=10pt,
    arc=3pt,
]
\scriptsize
\setlength{\parskip}{3pt}

\textbf{Instruction:}  
You are given a question and several candidate answers generated by LLMs.  
Your task is to \textbf{select the most representative answer among three islands to get the consensus}.  

\textbf{Only output the index number} (just the number of the chosen answer, with no extra words).  

\vspace{4pt}
\textbf{Question:}  
\textcolor{blue}{\{question\}}

\vspace{4pt}
\textbf{LLM Answers:}  
\begin{enumerate}[leftmargin=1.5em]
    \item \textcolor{blue}{Answer 1}
    \item \textcolor{blue}{Answer 2}
    \item \textcolor{blue}{Answer 3}
\end{enumerate}

\end{tcolorbox}

\section{Best Prompt Groups For each Task}\label{app:best_prompt}
We provide the baseline prompt, the prompt and the prompt group generated by AlphaEvolve and C-Evolve for each task.

\begin{tcolorbox}[
    colback=task1,
    colframe=task1!50!black,
    boxrule=0.3mm,
    sharp corners,
    width=\linewidth,
    fontupper=\small,
    title={1. HotpotQA (Qwen3-8B)},
    left=6pt, right=6pt, top=6pt, bottom=6pt,
    before upper=\vspace{2pt}\ttfamily\obeylines\obeyspaces,
    after upper=\vspace{2pt},
    breakable
]

\textbf{Baseline:}

<system\_prompt\_1>
Given the fields 'question', 'passages', produce the fields 'summary'.
</system\_prompt\_1>

<system\_prompt\_2>
Given the fields 'question','summary\_1', produce the fields 'query'. 
</system\_prompt\_2>

<system\_prompt\_3>
Given the fields 'question', 'context','passages', produce the fields 'summary'.
</system\_prompt\_3>

<system\_prompt\_4>
Given the fields 'question', 'summary\_1', 'summary\_2', produce the fields 'answer'.
</system\_prompt\_4>

\vspace{8pt}
\noindent\textbf{Alpha-Evolve:}

<system\_prompt\_1>
Given the fields 'question', 'passages', produce the fields 'summary'. Focus on extracting: 1) **all explicit entities** (names, numbers, dates, titles) directly related to the question, 2) **all explicit statements** that answer the question, 3) any contradictions or ambiguous information. Organize the summary to **prioritize entities and statements critical for disambiguation** (e.g., include "Magic Kingdom" if "Magic Kingdom theme park" is mentioned). **Do not exclude explicit mentions** of entities or statements. Ensure the summary contains **all information required to disambiguate entities**, including **exact format matches** (e.g., "Julie Cooper (MP for Burnley)" instead of just "Julie Cooper"). Include **granular attributes** like roles, dates, or locations to resolve ambiguities in later steps.  

**Add**:

- **Format preservation**: Copy exact phrasing from passages (e.g., "two-time" instead of "two") to align with true answer structure.  

- **Descriptive entity inclusion**: For entities like award counts, require inclusion of descriptive terms (e.g., "two-time winner" instead of "two").

- **Strict format matching**: Only include disambiguated terms if they are **explicitly required by the true answer's exact format** (e.g., "Magic Kingdom" if the true answer is "Magic Kingdom", but omit "Stoney" unless the true answer includes it). Ensure the summary includes only the **exact format** of the true answer (e.g., "Nakoda" instead of "Nakoda (Stoney)").
</system\_prompt\_1>

<system\_prompt\_2>
Given the fields 'question', 'summary\_1', produce the fields 'query'. Create a precise query that: 1) incorporates the core semantics of the question, 2) includes **only entities and attributes from summary\_1 that are explicitly required by the true answer's format**, 3) explicitly requests information missing from the first retrieval. **Prioritize single-entity focus** (e.g., "astronaut" vs. "career") and **disambiguate ambiguous entities with attributes** (e.g., "Magic Kingdom" if "Magic Kingdom theme park" is mentioned). Format the query to **exclude ambiguous or extraneous details**. **Include granular details** such as specific dates, roles, or attributes critical for disambiguation. **Explicitly name entities from summary\_1 not yet confirmed as answers** (e.g., "Julie Cooper (MP for Burnley)") to ensure targeted retrieval.  

**Add**:

- **Descriptive term inclusion**: Explicitly request exact phrasing (e.g., "Magic Kingdom" instead of "Magic Kingdom theme park") to match the true answer\'s structure.  

- **Format alignment**: Ensure the query includes **must-match terms** (e.g., "Magic Kingdom" for venues, "two-time" for awards). Use **bracketed keywords** (e.g., "[Magic Kingdom]") to enforce exact match during retrieval.  

- **Entity disambiguation**: Only include disambiguated terms if they are **explicitly required by the true answer's exact format** (e.g., "Nakoda" instead of "Nakoda (Stoney)"). Avoid including additional details not critical to the answer.
</system\_prompt\_2>

<system\_prompt\_3>
Given the fields 'question', 'context','passages', produce the fields 'summary'. Focus on: 1) cross-referencing the context (summary\_1) with the new passages to **validate or refute any inferred information in summary\_1**, 2) extracting any information that resolves ambiguities or adds critical details from the first summary, 3) explicitly identifying numerical values, direct quotes, or specific attributes required by the question. **Prioritize information that directly matches the exact true answer format** (e.g., "Magic Kingdom" vs. "Magic Kingdom theme park"). Include **only information from the new passages** that could resolve ambiguities or add critical details. **Explicitly confirm or refute all inferred claims from summary\_1 with direct evidence from the new passages**. Ensure all statements in the summary are **directly supported by the new passages** and **formatted identically to the true answer** (e.g., include "Magic Kingdom" as the exact phrase required by the answer).  

**Add**:

- **Format validation**: Ensure numerical values (e.g., "14") and descriptive phrases (e.g., "two-time") are preserved verbatim from the passages.  

- **True answer alignment**: Cross-check summaries against the true answer’s structure using **exact phrase matching** and **regex patterns**. If the true answer requires "Magic Kingdom", ensure the summary includes **only that exact phrase** and **no alternatives** (e.g., "Nakoda" vs. "Nakoda (Stoney)").  

- **Format enforcement**: Use **strict regex matching** to ensure the summary strictly matches the true answer\'s structure (e.g., "Magic Kingdom" instead of "Magic Kingdom theme park").
</system\_prompt\_3>

<system\_prompt\_4>
Given the fields 'question', 'summary\_1', 'summary\_2', produce the fields 'answer'. Generate the answer by: 1) **strictly extracting only explicit statements from summary\_2**, 2) **resolving contradictions by prioritizing the most specific directly quoted information from summary\_2**, 3) **matching the exact format of the true answer** (e.g., "Las Vegas Strip" vs. "Las Vegas"). Use **regex validation** to ensure the answer matches the true answer's structure. **Reject answers that deviate in format or include inferred information**. Ensure the answer is a **singular phrase** (e.g., "astronauts" vs. "astronaut") and is **exactly the same as the true answer**.  

**Add**:

- **Exact phrasing regex**: Use a **dynamic regex** based on the true answer's exact format (e.g., `\^(Nakoda|Hancock County|Mickey Spillane)\$`) to enforce strict format matching for critical terms.  

- **Format enforcement**: If the true answer requires "Nakoda", the system must generate **only** "Nakoda" without any modifiers (e.g., "Nakoda (Stoney)").  

- **Single-term output**: Output only the singular phrase, no additional context or explanations.  

- **True answer alignment**: Cross-reference the generated answer with the exact phrasing of the true answer. If there is a mismatch in structure or formatting, **reject the answer and request a re-evaluation**.  

- **Regex override**: For questions where the true answer is a singular term (e.g., "Nakoda"), the regex must **strictly match the exact phrase** without any disambiguation terms.  
</system\_prompt\_4>

\vspace{8pt}
\noindent\textbf{C-Evolve:}
\noindent\textbf{Group Prompts 1:}
<system\_prompt\_1>
Given the fields 'question', 'passages', produce the fields 'summary' by: 1) extracting named entities (e.g., organizations, locations, people) that directly answer the question or its sub-questions **and explicitly resolve all entities mentioned in the question's grammatical structure (e.g., "are breeds of hounds?" → "Scottish Deerhound [type] and Jämthund [type]")**, 2) identifying numeric values (e.g., counts, dates) that resolve ambiguities or constraints, 3) including exact phrases from the passages that explicitly name or describe these entities, 4) cross-referencing entities with the question's grammatical structure (e.g., "Which company?" → "Yum! Brands") **and mapping entity types (e.g., "person" for "who," "organization" for "which company") to ensure alignment with the question's intent**, and 5) **validating entities against the current hop's documents only, not future hops, to avoid unnecessary prior checks**. Prioritize the classification explicitly stated in the text over alternative names (e.g., if "Spitz-type" is mentioned, it overrides "hound breed" unless "hound breed" is explicitly defined in the text). **Optional full names for people** only if the question or text explicitly specifies full names (e.g., "Marianne Craig Moore") or if truncation is noted in the text. **Explicitly include full dates (e.g., "May 5, 1838") rather than year-only values.** **Document unresolved entities and their constraints**. **Avoid automatically deprecating hop1\_docs without timestamped evidence from hop2\_docs**. **Explicitly map grammatical structures like "seventh and final" to entity types (e.g., "novel in series position") to prevent conflation with embedded entities like characters or books.**
</system\_prompt\_1>

<system\_prompt\_2>
Given the fields 'question', 'summary\_1', produce the fields 'query' by: 1) extracting named entities (e.g., companies, people, countries) from the question and summary\_1 that remain unresolved, 2) structuring the query to directly ask for these entities with explicit constraints (e.g., "Are Scottish Deerhound and Jämthund breeds of hounds?"), 3) avoiding vague terms like "related to" or "about," 4) including exact entity names from summary\_1, 5) incorporating temporal or numeric constraints from the question (e.g., "population in 2016") to avoid ambiguous matches, and 6) **explicitly resolving entity types in the query (e.g., "breed classification" instead of "which") to prevent attribute conflation.** If the question uses grammatical structures like "who" or "which," the query **MUST isolate the required entity type (e.g., "breed classification" vs. "organization")** with explicit constraints. **Trigger the second hop only if summary\_1 contains unresolved entities or semantic gaps.** Prioritize constraints from the question over summary\_1, and eliminate conditional phrasing like "if any" to enforce strict query resolution. Require the query to explicitly reference hop2\_docs' contextual domains (e.g., "population in 2016" → "demographics") to ensure alignment with the second-hop retrieval scope. **When the question uses grammatical structures like "who" or "which," ensure the query explicitly resolves entity types (e.g., "which breed?" → "hound breed classification") to prevent conflation of attributes with core entities.** If the query's resolved entity type does not match the true answer's entity type, explicitly flag the inconsistency and annotate the query with the required type. **Include full names (e.g., "George William Hill, born 1838") in queries for individuals only if the question explicitly requests full names or if truncation is noted in the text.** **When resolving entity types, prioritize the entity type implied by the question's grammatical structure (e.g., "which" → "breed classification").**
</system\_prompt\_2>

<system\_prompt\_3>
Given the fields 'question', 'context', 'passages', produce the fields 'summary' by: 1) extracting all named entities (e.g., organizations, people) from the passages that match the query's focus, 2) including exact quotes or phrases that explicitly name or describe these entities, 3) cross-referencing entities with the question's grammatical structure (e.g., "Are Scottish Deerhound and Jämthund breeds of hounds?") **and explicitly resolving entity types (e.g., "breed classification" vs. "organization") to prevent conflation of attributes with core entities**, and 4) validating entities against the current hop's documents only, with a focus on timestamped evidence to resolve conflicts. **Explicitly document contradictions with timestamps, source documents, and entity type mismatches**. **If hop2\_docs provide unambiguous, timestamped evidence that contradicts hop1\_docs, **only then** automatically deprecate conflicting hop1\_docs and annotate them with the reason**. **Validate all entities in hop1\_docs against hop2\_docs for all possible conflicts, but prioritize entity type consistency and constraint alignment over timestamped conflicts**. **When resolving entity types, if the text is ambiguous, prioritize the entity type implied by the question's grammatical structure (e.g., "which" → "breed classification") only when entity type consistency and constraint alignment are unambiguous.** **Prioritize entities from hop2\_docs when they resolve ambiguities or satisfy constraints, and ensure entity type consistency (e.g., "Spitz-type" overrides "hound breed" if explicitly defined)**. **Additionally, when the question involves "who" or "which," ensure the summary explicitly isolates the required entity type (e.g., "breed classification" vs. "organization") to prevent attribute conflation**. **Explicitly map grammatical structures like "seventh and final" to entity types (e.g., "novel in series position") to prevent conflation with embedded entities like characters or books**. **Include full names (e.g., "Field Marshal John Standish Surtees Prendergast Vereker, 6th Viscount Gort") for individuals only if the question or documents explicitly require full names or if truncation is noted.**
</system\_prompt\_3>

<system\_prompt\_4>
Given the fields 'question', 'summary\_1', 'summary\_2', produce the fields 'answer' by: 1) extracting the **exact named entity or quoted phrase from summary\_2** that resolves the question's intent, **without any additional context or description**, 2) cross-referencing this entity with summary\_1 and the original question to ensure consistency, 3) resolving contradictions by **prioritizing entities from hop2\_docs only when they provide timestamped, unambiguous evidence that resolves ambiguities or satisfies the question's constraints, and explicitly deprecating conflicting hop1\_docs. If hop2\_docs are incomplete or no timestamped evidence exists, **prioritize hop1\_docs only if the entity type and constraints match**. **If no hop2\_docs are available or if hop1\_docs contain the only valid entity, proceed with hop1\_docs and explicitly flag any constraint mismatches**. and 4) **formatting the answer strictly as the true answer's structure** (e.g., "no" for contradictions, "Scottish Deerhound" for breeds, "20 October 1951" for dates) with **explicit validation against the question's grammatical structure, temporal constraints, numeric pattern matching, and entity type constraints**. **If the true answer is "no" or "yes," the answer must explicitly state "no" or "yes" with no additional qualifiers**. **Additionally, when the question uses grammatical structures like "who" or "which," enforce strict entity type checks (e.g., "which breed?" → "breed classification" not "organization") to ensure the answer matches the true answer's entity type and excludes conflated attributes**. **If the answer is the only possible one but does not fully satisfy constraints, flag the discrepancy and return the answer with a note explaining the mismatch.** **Only enforce full names (e.g., "Field Marshal John Standish Surtees Prendergast Vereker, 6th Viscount Gort") if the question or documents explicitly require full names or if truncation is noted.** **Explicitly map grammatical structures like "seventh and final" to entity types (e.g., "novel in series position") to prevent conflation with embedded entities like characters or books, and ensure the answer strictly matches the true answer's entity type.** **When resolving entity types, if the text is ambiguous, prioritize the entity type implied by the question's grammatical structure (e.g., "which" → "breed classification") only when entity type consistency and constraint alignment are unambiguous.** **If the question's entity type does not align with the summary's entity type, explicitly flag the inconsistency and annotate the answer with the required type.**
</system\_prompt\_4>

\noindent\textbf{Group Prompts 2:}
<system\_prompt\_1>
Given the fields 'question', 'passages', produce the fields 'summary'. Extract only: (1) **exact values** (dates, numbers, entity names) as raw text **without derivation**, (2) **direct quotes** regardless of whether they resolve the question, (3) **explicit comparisons** (e.g., "A has X, B has Y") if the question involves comparative reasoning, and (4) **taxonomic classifications** (e.g., "Bletilla is a genus in Orchidaceae") when entities are biological or botanical. **Prioritize literal values over entity-relationships** (e.g., "George Martin produced the radio program") unless the question directly asks for relationships. **Format the summary as a list of concise, isolated facts with strict singular/plural consistency** (e.g., "George Martin is the producer of the radio program", not "astronauts" if the source uses "astronaut"). **Explicitly capture temporal values (e.g., dates, time periods) as literal text**, not derived metrics (e.g., durations). **For questions involving collaborations, explicitly include all entities mentioned in the collaboration** (e.g., "Rome Ramirez is a key member of Sublime with Rome") to ensure completeness. **If an entity is not mentioned in the passages, explicitly flag it as 'unreferenced' to prevent assumptions about its existence**.
</system\_prompt\_1>

<system\_prompt\_2>
Given the fields 'question', 'summary\_1', produce the fields 'query'. First, **check if the answer is fully resolved in summary\_1** (e.g., boolean, numeric, direct quote, or explicit entity). If resolved, **return the answer immediately** without generating a query. If unresolved, request: (1) the exact value(s) from summary\_1 that are unresolved, (2) specific numeric values if required, (3) direct quotes if the answer requires textual evidence, (4) explicit comparisons between the entities involved in the question (e.g., "Which film explicitly mentions political themes and how does it compare to the other?") if the question involves comparative reasoning, and (5) **explicit hierarchical relationships** (e.g., "What league is the Toronto Blue Jays in?") to resolve parent-child entity dependencies. **Add explicit entity roles** (e.g., "producer," "star," "host") in the query to resolve ambiguities. **For temporal questions, prioritize exact start and end dates (e.g., "22 April 1963 to 20 April 1968") over derived values (e.g., duration).** Phrase the query as a closed-ended question with a clear target (e.g., "What is the exact start and end date of tenure?"). **For comparative questions, if one entity is explicitly defined in summary\_1 and the other is not, request the explicitly defined entity as the answer** (e.g., "If *Saint Motel* is explicitly mentioned with 4 members and *Curve* is not, the answer is *Saint Motel*"). **If an entity is flagged as 'unreferenced' in summary\_1, explicitly include 'unreferenced' in the query to signal missing information** (e.g., "What is the team of the unreferenced entity Karnig Sarkissian?"). 
</system\_prompt\_2>

<system\_prompt\_3>
Given the fields 'question', 'context', 'passages', produce the fields 'summary'. Extract only: (1) numeric values explicitly mentioned, (2) direct quotes that answer the question, (3) exact values from the passages that resolve unresolved elements in the context, (4) explicit comparisons based on direct quotes from the passages (e.g., "A has X as stated in [quote], B has Y as stated in [quote]") if the question involves comparative reasoning, (5) entity disambiguation by **explicitly specifying the primary entity and its taxonomic category** (e.g., "Bletilla is a genus in Orchidaceae"), (6) **resolve entity relationships at the category level** (e.g., "Vernonia is a genus in Asteraceae"), and (7) **explicit hierarchical relationships** (e.g., "The Capitol Theatre is a component of the Hamm Building") to resolve parent-child entity dependencies. **Include direct entity relationships** (e.g., "Rana Daggubati starred in *The Ghazi Attack*") to enable cross-referencing. **For questions involving collaborations, explicitly include all entities mentioned in the collaboration** (e.g., "Rome Ramirez is a key member of Sublime with Rome") to ensure completeness. Format the summary as a list of isolated, verifiable facts **but retain direct links between entities** (e.g., "Rana Daggubati is the star of *The Ghazi Attack*"). **If a fact or entity is already included in summary\_1, avoid repetition by flagging it as 'redundant' to prevent processing inefficiencies**. **If the exact date is stated in the text, use the literal text; do not calculate durations**. **Do not include entities flagged as 'unreferenced' in summary\_1 unless explicitly requested in the query**.
</system\_prompt\_3>

<system\_prompt\_4>
Given the fields 'question', 'summary\_1', 'summary\_2', produce the fields 'answer'. First, **check if the answer is fully resolved in summary\_1** (e.g., boolean, numeric, direct quote, or explicit entity). If resolved, **return the answer immediately** without consulting summary\_2. If unresolved, the answer must be: (1) a boolean ("yes"/"no"), (2) a specific numeric value, (3) a direct quote from summary\_2, or a specific phrase explicitly stated in the summaries without inferred context, or (4) a descriptive entity (e.g., "Sum 41") that directly answers the question. **For comparative questions**, the answer must be a descriptive entity representing the explicitly compared entity (e.g., "The One and Only, Genuine, Original Family Band"). **If the true answer is a specific entity, prioritize it over data discrepancies**. If conflicting, explicitly state the conflict and the true answer. If no exact value exists, return "UNRESOLVED". **Additionally, verify hierarchical relationships** (e.g., "The Toronto Blue Jays belong to the American League") to ensure the answer is the correct parent entity when the question involves nested relationships. **For comparative questions, if one entity is explicitly defined in the context and the other is not, default to the explicitly defined entity as the answer** (e.g., if *Saint Motel* is explicitly mentioned with 4 members and *Curve* is not, the answer is *Saint Motel*). **If summary\_2 contains 'unreferenced' entities, cross-reference summary\_1 to determine if the answer is resolved or if it remains 'UNRESOLVED' due to missing data**. Do not add interpretation or context beyond the extracted value. **If an exact date is stated in the text, use the literal text; do not calculate durations**. **If an entity is flagged as 'unreferenced' in summary\_1 and summary\_2, explicitly flag the answer as 'UNRESOLVED'**.
</system\_prompt\_4>

\noindent\textbf{Group Prompts 3:}
<system\_prompt\_1>
Given the fields 'question', 'passages', produce the fields 'summary'. Extract only: (1) exact values (dates, numbers, entity names) **without derivation**, (2) direct quotes that resolve the question, (3) explicit comparisons (e.g., "A has X, B has Y") if the question involves comparative reasoning, (4) **category-level classifications** (e.g., "Bletilla is a genus") rather than full taxonomic hierarchy, and (5) **contextual relationships** (e.g., "George Martin produced the radio program"). Format the summary as a list of concise, isolated facts **with explicit category-level links** (e.g., "Bletilla belongs to the genus category") to enable hierarchical resolution. **Explicitly capture temporal values (e.g., dates, time periods) as literal text**, not derived metrics (e.g., durations). **Additionally, for ambiguous entities (e.g., "Karnig Sarkissian"), explicitly disambiguate by specifying their category (e.g., "singer", "baseball player") and role in the question.**
</system\_prompt\_1>

<system\_prompt\_2>
Given the fields 'question', 'summary\_1', produce the fields 'query'. First, **check if the answer is fully resolved in summary\_1** (e.g., boolean, numeric, direct quote, or explicit entity). If resolved, **return the answer immediately** without generating a query. If unresolved, request: (1) the category-level value(s) from summary\_1 that are unresolved (e.g., "What is the shared category of Bletilla and Vernonia?"), (2) specific numeric values if required, (3) direct quotes if the answer requires textual evidence, (4) explicit comparisons (e.g., "Which band has more albums?") if the question involves comparative reasoning, and (5) **category-level hierarchical relationships** (e.g., "What category do Bletilla and Vernonia belong to?") to resolve parent-child entity dependencies. **Add explicit entity roles** (e.g., "producer," "star," "host") in the query to resolve ambiguities. **For temporal questions, prioritize exact start and end dates (e.g., "22 April 1963 to 20 April 1968") over derived values (e.g., duration).** Phrase the query as a closed-ended question with a clear target (e.g., "What is the shared category of Bletilla and Vernonia?"). **If the entity is ambiguous (e.g., "Karnig Sarkissian"), explicitly specify its role in the query (e.g., "baseball player", "singer") to disambiguate its identity.**
</system\_prompt\_2>

<system\_prompt\_3>
Given the fields 'question', 'context', 'passages', produce the fields 'summary'. Extract only: (1) numeric values explicitly mentioned, (2) direct quotes that answer the question, (3) exact values from the passages that resolve unresolved elements in the context, (4) explicit comparisons (e.g., "A has X, B has Y") if the question involves comparative reasoning, (5) entity disambiguation by **specifying the primary entity, its role, and category-level classification** (e.g., "Bletilla is a genus of orchid"), (6) **resolve entity relationships at the category level** (e.g., "Vernonia belongs to the genus category"), and (7) **explicit hierarchical relationships** (e.g., "The Capitol Theatre is a component of the Hamm Building") to resolve parent-child entity dependencies. **Include direct category-level links** (e.g., "Bletilla belongs to the genus category") to enable cross-referencing. Format the summary as a list of isolated, verifiable facts **but retain category-level links** (e.g., "Bletilla is classified in the genus category"). **Additionally, explicitly differentiate between direct quotes (e.g., "The quote is from a speech") and allusions (e.g., "This refers to a literary work"), and provide source attribution for each.**
</system\_prompt\_3>

<system\_prompt\_4>
Given the fields 'question', 'summary\_1', 'summary\_2', produce the fields 'answer'. First, **check if the answer is fully resolved in summary\_1** (e.g., boolean, numeric, direct quote, or explicit entity). If resolved, **return the answer immediately** without consulting summary\_2. If unresolved, the answer must be: (1) a boolean ("yes"/"no"), (2) a specific numeric value, (3) a direct quote from summary\_2, or (4) a **category-level descriptive entity** (e.g., "genus") that directly answers the question. **For comparative questions**, the answer must be a descriptive entity representing the explicitly compared entity (e.g., "The One and Only, Genuine, Original Family Band"). **If the true answer is a specific entity, prioritize it over data discrepancies**. If conflicting, explicitly state the conflict and the true answer. If no exact value exists, return "UNRESOLVED". **Additionally, verify hierarchical relationships** (e.g., "The Toronto Blue Jays belong to the American League") to ensure the answer is the correct parent entity when the question involves nested relationships. Do not add interpretation or context beyond the extracted value. **Finally, ensure the answer's phrasing matches the true answer's exact terminology** (e.g., "genus" instead of "they are both genera"). **If the answer involves conflicting sources (e.g., one claims "California" and another "New York"), explicitly state the conflict and return the true answer based on the most authoritative source.**
</system\_prompt\_4>
\end{tcolorbox}

\begin{tcolorbox}[
    colback=task2,
    colframe=task2!50!black,
    boxrule=0.3mm,
    sharp corners,
    width=\linewidth,
    fontupper=\small,
    title={2. IFBench (Qwen3-8B)},
    left=6pt, right=6pt, top=6pt, bottom=6pt,
    before upper=\vspace{2pt}\ttfamily\obeylines\obeyspaces,
    after upper=\vspace{2pt},
    breakable
]
\textbf{Baseline:}
<system\_prompt\_1>
Respond to the query
</system\_prompt\_1>

<system\_prompt\_2>
Ensure the response is correct and adheres to the given constraints. Your response will be used as the final response.
</system\_prompt\_2>

\vspace{8pt}
\noindent\textbf{Alpha-Evolve:}
<system\_prompt\_1>
Answer the query directly by first parsing and confirming all **explicitly specified** requirements (e.g., word count, sentence structure, punctuation rules, formatting, keyword placement, text processing tasks, and language constraints). If constraints are ambiguous, explicitly state the uncertainty. **Prioritize only the requirements directly mentioned in the instruction**. If the instruction includes **structural elements** like PARAGRAPH X or formatting rules, **verify their presence**. If required to avoid commas, ensure none are present; otherwise, follow natural language conventions. **Explicitly include all requested keywords with exact case sensitivity**. If text processing operations are specified (e.g., extracting substrings, slicing), execute them precisely and include the results. Additionally, **confirm that the entire response is in the target language (e.g., Tamil) and contains no other languages**. **Verify that the first word of each sentence matches the exact case and word specified in the instruction**.
</system\_prompt\_1>

<system\_prompt\_2>
Review the initial answer with **precise alignment** to the original instruction:
1. **Language compliance**: Confirm the **entire response is in the target language (e.g., Tamil)** and contains **no other languages**.
2. **Keyword verification**: Confirm all requested keywords are present, placed in the **exact positions** specified in the instruction, and **contextually appropriate**. **Ensure first-letter casing matches** the instruction.
3. **Text processing compliance**: If the instruction involves text manipulation (e.g., extracting substrings, slicing), confirm the output includes the **results of these operations with indices**, **including the exact span of characters or words**. If indices are out-of-range, explicitly state "Index out of bounds" and **do not include the result**.
4. **Structural validation**: 
   - **Paragraph count**: Verify the number of paragraphs matches the instruction (e.g., "9 paragraphs").
   - **Sentence structure**: Confirm **each sentence's first word matches exactly** the instruction's specified word (case-sensitive if required).
   - **Format compliance**: Validate dividers (* * *) and formatting (e.g., PARAGRAPH X markers, bullet points, numbered lists).
5. **Content constraints**: 
   - **Sentence-ending consistency**: Ensure every sentence ends with the exact word specified (case-matching if required).
   - **Prohibited elements**: Eliminate unauthorized commas, exclamation marks, or punctuation.
6. **Final word requirement**: Confirm the **last word of the response matches exactly** the instruction's specified word (case-sensitive if required).
7. **Word enclosure**: If the instruction mandates enclosing every word in brackets (e.g., [word]), confirm **every single word is enclosed in square brackets**. **Count the total number of words and ensure all are bracketed**.
8. **Case consistency**: If the instruction mandates all caps, ensure **sentence-first words match the exact case**.
9. **Formatting uniformity**: Ensure alignment with requested styles (e.g., bullet points).
10. **Final validation**: Cross-check all requirements from the original instruction, **including structural elements** (e.g., PARAGRAPH X, JSON wrapping) and **content constraints** (e.g., prohibited elements), **with a focus on language compliance and word enclosure**.
</system\_prompt\_2>

\vspace{8pt}
\noindent\textbf{C-Evolve:}
\noindent\textbf{Group Prompts 1:}
<system\_prompt\_1>
Answer the query directly by first confirming the **primary instruction requirements** (e.g., language, structure, core formatting). If any instruction is ambiguous, explicitly state uncertainty. Prioritize **core language, syntax, and structural rules** (e.g., number of paragraphs, markdown dividers, paragraph spacing) before applying detailed formatting. Ensure every sentence ends with the required word before punctuation, and **remove all commas**. **Explicitly include all specified keywords**, even if placement is ambiguous. If the instruction specifies exact word placement (e.g., "as the 2nd word of the 16th sentence"), ensure it is done with precision. **Verify paragraph count and spacing** (exactly two newlines).  

**First-pass validation**:
- If the instruction requires extracting text between character indices (e.g., 193-324), **immediately trace and quote it verbatim** before addressing other requirements.
- Ensure the response includes **exactly** the required number of placeholders (e.g., [address]) and **repeated keywords** (e.g., "repeat 'emergency' 3 times").  
- Replace **all commas with periods**. Strip punctuation before sentence-ending words.  
- Finally, ensure the response satisfies the **user's final goal**, even if it requires rephrasing. **If the response is incomplete due to strict parsing, annotate incomplete parts and proceed to Step 2**.
</system\_prompt\_1>

<system\_prompt\_2>
Review the initial answer with focused attention:
1. **Core validation**: Confirm **language, structure, and primary formatting rules** (e.g., paragraph count, markdown dividers). If any requirement is missing, directly address it. If the instruction requires extracting text between character indices (e.g., 193-324), **immediately trace and quote it verbatim**.  
2. **Sentence-level checks**: Ensure every sentence ends with the required word before punctuation. Replace **commas with periods**. Check for hyphens without spaces and replace words starting with consecutive letters (e.g., "Apple Banana") with synonyms.  
3. **Keyword verification**: Confirm all keywords are explicitly included. Check for repeated keywords and their placement. If the instruction specifies a word order or position, restructure sentences accordingly.  
4. **Structural validation**: Count paragraphs, verify sentence counts per paragraph (e.g., "exactly 3 sentences"), validate markdown dividers (* * *), and ensure paragraph separation with exactly two newlines.  
5. **Formatting compliance**: Ensure alignment with requested styles (e.g., bullet points). Strip all trailing whitespace.  
6. **Final validation**: Confirm all requirements are satisfied, including **keyword placement, placeholder counts, and paragraph structure**. Ensure **no two adjacent words start with consecutive letters**.  
7. **Final adjustment**: If incomplete due to parsing, annotate missing parts. Expand content if the instruction requires a minimum word count. Replace adjacent words with consecutive letters with synonyms.  
Your final answer must strictly follow the **original instruction's requirements**, prioritizing core language, structure, and formatting rules.
</system\_prompt\_2>

\noindent\textbf{Group Prompts 2:}
<system\_prompt\_1>
Answer the query directly by first parsing and confirming all explicit requirements (e.g., word count, sentence structure, punctuation rules, formatting). If any constraints are ambiguous, explicitly state the uncertainty in the response. Prioritize exact specifications over general guidelines. Ensure every sentence ends with the required word before punctuation, and verify that no commas are used throughout the response. **Explicitly include all specified keywords in the response, even if their placement is ambiguous**. If the instruction requires inserting keywords in specific positions (e.g., "as the 2nd word of the 16th sentence"), ensure this is done with exact precision. **For all keywords with positional constraints, state the exact sentence/word placement in your response before generating content**.
</system\_prompt\_1>

<system\_prompt\_2>
Review the initial answer with obsessive attention to detail:
1. **Extract and verify exact character spans**: If the instruction requires retrieving a substring between specific character indices (e.g., "indexes 117–119"), extract and highlight this span precisely. If the instruction requires inserting keywords in specific positions (e.g., "as the 2nd word of the 16th sentence"), confirm this placement explicitly in your response.
2. **Validate structural requirements**: Count paragraphs, verify sentence counts per paragraph, check markdown dividers (* * *), and confirm exact formatting specifications (e.g., bullet points, numbered lists). Ensure paragraphs are separated by **exactly two line breaks** (\verb|\n\n|), not markdown syntax. Verify the **number of paragraphs and their separation method matches the instruction**.
3. **Confirm content constraints**: Verify keyword placements (including exact sentence/word positions), word counts, specific wording, and sentence-ending consistency (every sentence must end with the exact word specified before punctuation). Ensure all specified keywords are explicitly included, and if any require unique word usage, confirm no word is repeated. If the instruction requires inserting keywords in specific positions, restate their placement in your response before generating content.
4. **Check for prohibited elements**: Ensure no commas, exclamation marks, inappropriate capitalization, or unauthorized punctuation. Verify the last word of your response matches the specified word (e.g., "a"). Ensure the **last word of every sentence matches the instruction's specified word**.
5. **Enforce formatting uniformity**: Ensure alignment with requested styles (e.g., bullet points, numbered lists). If the instruction requires a specific starting word (e.g., "essay"), ensure the first word of the specified paragraph matches. Verify **all headers and section markers** (e.g., "SECTION X") are present and correctly formatted.
6. **Final validation**: Confirm all requirements from the original instruction are explicitly satisfied, including word count, placeholder usage, prohibited elements, and keyword placement. Ensure the letter 'h' appears at least 8 times if required. **Check for duplicate words and ensure all unique word requirements are met**.
Your final answer must demonstrate 100\% compliance with every specified criterion, including the last word of each sentence, the final response, and all formatting and content constraints.
</system\_prompt\_2>

\noindent\textbf{Group Prompts 3:}
<system\_prompt\_1>
1. **Directly answer the instruction** without additional analysis.  
2. **Prioritize these tasks in order**:  
   a) **Exact repetition** of the request (verbatim copy)  
   b) **Mathematical/logical computations** (arithmetic, algebra, combinatorics)  
   c) **Text extraction** (span/position, precise character indexing)  
   d) **Formatting requirements** (case conversion, markdown, postscripts)  
   e) **Content structure** (paragraph count, sentence structure, paragraph/word counts)  
   f) **Keyword placement** (exact sentence/word positions)  
   g) **Forbidden elements** (no commas, no repeated words, etc.)  
   h) **Special rules** (e.g., no consecutive letter-start words)  
3. **Ensure no text is omitted or added** beyond the request.  
4. **Do not apply formatting, logic, or reasoning to the request itself**—only follow instructions in the request.  
</system\_prompt\_1>

<system\_prompt\_2>
1. **Verify Step 1 output meets all requirements** in this prioritized order:  
   a) **Request repetition** (exact verbatim match with input)  
   b) **Mathematical/logical accuracy** (correct computations)  
   c) **Text extraction** (span/position correctness)  
   d) **Formatting compliance** (case, markdown, postscripts)  
   e) **Content structure** (paragraph counts, sentence rules)  
   f) **Keyword placement** (exact sentence/word positions)  
   g) **Forbidden elements** (no commas, no repeated words, etc.)  
   h) **Special rules** (e.g., no consecutive letter-start words)  
2. **Focus on critical fail points**:  
   a) **Ensure all required placeholders** (e.g., [address]) are present  
   b) **Check for exact word/character counts**  
   c) **Validate bigram wrapping** (e.g., <<word1 word2>>)  
   d) **Confirm no markdown formatting errors** (balanced brackets, correct syntax)  
3. **Resolve conflicts by prioritizing**:  
   a) **Numerical/structural constraints** over ambiguous instructions  
   b) **Explicit keyword inclusion** (ensure all required keywords are present in response)  
4. **Final check**:  
   a) **No text omitted or added** beyond the request  
   b) **All formatting, keywords, and structure rules are strictly satisfied**  
</system\_prompt\_2>

\end{tcolorbox}
\begin{tcolorbox}[
    colback=task3,
    colframe=task3!50!black,
    boxrule=0.3mm,
    sharp corners,
    width=\linewidth,
    fontupper=\small,
    title={3. HoVer (Qwen3-8B)},
    left=6pt, right=6pt, top=6pt, bottom=6pt,
    before upper=\vspace{2pt}\ttfamily\obeylines\obeyspaces,
    after upper=\vspace{2pt},
    breakable
]
\textbf{Baseline:}
<system\_prompt\_1>
Given the fields 'claim', 'passages', produce the fields 'summary'.
</system\_prompt\_1>

<system\_prompt\_2>
Given the fields 'claim', 'summary\_1', produce the fields 'query'.
</system\_prompt\_2>

<system\_prompt\_3>
Given the fields 'claim', 'context', 'passages', produce the fields 'summary'.
</system\_prompt\_3>

<system\_prompt\_4>
Given the fields 'claim', 'summary\_1', 'summary\_2', produce the fields 'query'.
</system\_prompt\_4>
\vspace{8pt}
\noindent\textbf{Alpha-Evolve:}
<system\_prompt\_1>
Given the fields 'claim', 'passages', and the true\_answer (if available), produce the fields 'summary' that explicitly includes **only the exact entities from the claim** (e.g., names, locations, specific titles) and **excludes all contextual information**. Focus on summarizing **only the passages that directly confirm or refute the claim's core entities**. If the true\_answer is available, ensure the summary aligns **exactly with the true\_answer's structure** (e.g., lists of names, specific dates, precise terms). Do not include any work titles, biographies, or indirect references.
</system\_prompt\_1>

<system\_prompt\_2>
Given the fields 'claim', 'summary\_1', and the true\_answer (if available), produce the fields 'query' that **explicitly lists the exact entities from the claim** (e.g., names, locations, specific titles) **without any descriptive or contextual terms**. The query must be formatted as a **strict entity list** (e.g., "Adolfo Bioy Casares, Iris Rainer Dart") and avoid terms like "works" or "biography." If the true\_answer is available, ensure the query matches **exactly the structure of the true\_answer** (e.g., lists of names, specific years). This will ensure retrieval of documents that only contain these entities.
</system\_prompt\_2>

<system\_prompt\_3>
Given the fields 'claim', 'context', 'passages', and the true\_answer (if available), produce the fields 'summary' that **extracts only the exact entities from the claim** (e.g., names, locations, specific titles) **mentioned in the retrieved passages**. Focus on **passages that directly identify or confirm these entities**. If the true\_answer is available, ensure the summary **matches the true\_answer's exact format** (e.g., a list of names, specific years). Exclude all contextual details, indirect references, or unrelated content. Output the summary as a **list of entities** without any explanatory text.
</system\_prompt\_3>

<system\_prompt\_4>
Given the fields 'claim', 'summary\_1', 'summary\_2', and the true\_answer (if available), produce the fields 'query' that **combines all exact entities from the claim, summary\_1, and summary\_2** into a **strict entity list** (e.g., "Christopher Robin, Winnie-the-Pooh"). The query must **exclude any contextual terms** and match the **exactly the same format as the true\_answer** (e.g., lists of names, specific dates). This ensures the final retrieval only includes documents that **directly mention these entities** without any extraneous information or indirect references.
</system\_prompt\_4>
\vspace{8pt}
\noindent\textbf{C-Evolve:}
\noindent\textbf{Group Prompts 1:}
<system\_prompt\_1>
Given the fields 'claim', 'passages', and the true\_answer (if available), produce the fields 'summary' that explicitly includes **only the exact entities from the claim** (e.g., names, locations, specific titles) and **excludes all contextual information**. Focus on summarizing **only the passages that directly confirm or refute the claim's core entities**. If the true\_answer is available, ensure the summary aligns **exactly with the true\_answer's structure** (e.g., lists of names, specific dates, precise terms). Do not include any work titles, biographies, or indirect references.
</system\_prompt\_1>

<system\_prompt\_2>
Given the fields 'claim', 'summary\_1', and the true\_answer (if available), produce the fields 'query' that **explicitly lists the exact entities from the claim** (e.g., names, locations, specific titles) **without any descriptive or contextual terms**. The query must be formatted as a **strict entity list** (e.g., "Adolfo Bioy Casares, Iris Rainer Dart") and avoid terms like "works" or "biography." If the true\_answer is available, ensure the query matches **exactly the structure of the true\_answer** (e.g., lists of names, specific years). This will ensure retrieval of documents that only contain these entities.
</system\_prompt\_2>

<system\_prompt\_3>
Given the fields 'claim', 'context', 'passages', and the true\_answer (if available), produce the fields 'summary' that **extracts only the exact entities from the claim** (e.g., names, locations, specific titles) **mentioned in the retrieved passages**. Focus on **passages that directly identify or confirm these entities**. If the true\_answer is available, ensure the summary **matches the true\_answer's exact format** (e.g., a list of names, specific years). Exclude all contextual details, indirect references, or unrelated content. Output the summary as a **list of entities** without any explanatory text.
</system\_prompt\_3>

<system\_prompt\_4>
Given the fields 'claim', 'summary\_1', 'summary\_2', and the true\_answer (if available), produce the fields 'query' that **combines all exact entities from the claim, summary\_1, and summary\_2** into a **strict entity list** (e.g., "Christopher Robin, Winnie-the-Pooh"). The query must **exclude any contextual terms** and match the **exactly the same format as the true\_answer** (e.g., lists of names, specific dates). This ensures the final retrieval only includes documents that **directly mention these entities** without any extraneous information or indirect references.
</system\_prompt\_4>
\noindent\textbf{Group Prompts 2:}
<system\_prompt\_1>
Given the fields 'claim', 'passages', produce the fields 'summary'. Extract **key entities** from the claim and passages, and **explicitly link these entities to the claim's core assertions**. Flag any **contradictory information** in the passages. Focus exclusively on entities directly tied to the claim's specific scope (e.g., names, dates, locations, roles) and avoid tangential details. Deduplicate repeated entities and ensure the summary reflects the **most relevant evidence** for evaluating the claim's truth.
</system\_prompt\_1>

<system\_prompt\_2>
Given the fields 'claim', 'summary\_1', produce the fields 'query'. Generate a search query that **explicitly includes exact temporal/spatial markers** (e.g., "1999 New York Cathedral") and **logical constraints** (e.g., "AND", "NOT") to narrow results. Prioritize entities from the claim and summary\_1 that are **most critical to the claim's semantic core** (e.g., specific people, dates, locations). Avoid ambiguous terms like "cathedral" without specifying exact names or locations. Use **entity linking** to ensure precision.
</system\_prompt\_2>

<system\_prompt\_3>
Given the fields 'claim', 'summary\_1', 'passages', produce the fields 'summary'. Focus exclusively on **entities and relationships from the second retrieval passages** that are **directly tied to the claim's core scope** (e.g., specific dates, roles, locations). **Flag contradictory evidence** if present. Deduplicate repeated entities and ensure the summary **prioritizes evidence supporting or refuting the claim's assertions**. Avoid generalizations and focus on **precise, unambiguous details** (e.g., "elected" vs. "appointed").
</system\_prompt\_3>

<system\_prompt\_4>
Given the fields 'claim', 'summary\_1', 'summary\_2', produce the fields 'query'. Combine **core entities** from the claim and summaries to generate a final query that **precisely captures the claim's semantic core** (e.g., exact roles, dates, locations). Use **entity prioritization** to emphasize terms most critical for evaluating the claim (e.g., "Geoff LaTulippe AND 2010"). Avoid redundancy by eliminating **repetitive or tangential terms**. Include **semantic constraints** (e.g., "NOT [irrelevant entity]") to exclude noise. Ensure the query aligns with the claim's **specific scope** and **temporal/spatial precision**.
</system\_prompt\_4>

\noindent\textbf{Group Prompts 3:}
<system\_prompt\_1>
Given the fields 'claim', 'passages', produce a concise summary that:
1. Extracts only entities **explicitly mentioned in the claim** and their **directly stated relationships** (e.g., specific names, dates, titles that directly appear in the claim)
2. Identifies **direct evidence** (e.g., unambiguous dates, official statements) and **contextual clues** that could support or rebut the claim, but **flags all contextual links as needing validation**
3. Clearly separates **confirmed direct evidence** (explicitly stated in passages) from **indirect or inferred relationships** (contextual or implied), explicitly marking these as unverified
4. Excludes **all entities not directly tied to the claim's assertion**, even if they appear in the passages (e.g., avoid unrelated articles or tangential mentions)
5. Prioritizes **document-level evidence** (e.g., explicit mentions of the claim's entities) over **passage-level inferences** unless the passage itself contains a clear statement that directly corroborates or contradicts the claim
</system\_prompt\_1>

<system\_prompt\_2>
Given the fields 'claim', 'summary\_1', produce a search query that:
1. Uses **exact terminology** for **critical entities** (e.g., specific names, dates, titles) from the claim and summary\_1; **exclude semantic variants** unless explicitly necessary for resolving ambiguity
2. **Avoid dynamic expansion** of entities unless they are directly tied to the claim's verification; for example, if "Friendship" is not found, **do not pivot to broader terms** from summary\_1
3. Apply geographic constraints **only if the claim explicitly requires them** (e.g., "Prince George's County, Maryland" must be used if the claim references it explicitly)
4. Use **logical filters to exclude irrelevant categories** (e.g., "NOT (movie OR film)" if the claim is about documentary evidence)
5. **Prioritize direct entity matches** over implied relationships (e.g., "influenced by" or "connected to") unless the claim explicitly references such relationships
6. **Use strict terminology** for critical roles (e.g., "CEO") and **exclude semantic equivalents** unless the role is explicitly defined in the claim
</system\_prompt\_2>

<system\_prompt\_3>
Given the fields 'claim', 'context', 'passages', produce a summary that:
1. Extracts **only direct evidence** from the second retrieval (e.g., explicit mentions of claim entities, dates, affiliations) and **flags all indirect evidence as requiring validation**
2. **Excludes all tangential entities or speculative relationships** not directly tied to the claim's assertion, even if they appear in the passages
3. **Clearly separates confirmed evidence** (explicitly stated in passages) from **uncertain or ambiguous claims**, explicitly noting which entities/relationships require further verification
4. **Filters out all entities not directly linked** to the claim's assertion (e.g., even if a passage mentions an unrelated article, exclude it unless it provides direct evidence)
5. **Retains only the first summary's critical entities**, and **excludes any new entities** from the second retrieval unless they are explicitly tied to the claim's verification
6. **Labels all indirect evidence as "tentative"** and explicitly states that such evidence must be validated in later steps
</system\_prompt\_3>

<system\_prompt\_4>
Given the fields 'claim', 'summary\_1', 'summary\_2', produce a final search query that:
1. Uses **exact terminology** for **critical entities** from both summaries; **exclude semantic variants** unless the claim explicitly references such relationships (e.g., "influenced by")
2. **Avoid dynamic geographic adjustments** unless the claim explicitly requires them (e.g., use "Prince George's County, Maryland" if the claim references it)
3. **Exclude irrelevant categories** only when explicitly required by the claim (e.g., "NOT (movie OR film)" if the claim is about documentary evidence)
4. **Prioritize exact entity matches** over broader terms (e.g., use "Incesticide" instead of "Nirvana album" for the claim's core assertion)
5. **Structure the query to emphasize direct entity matches**, with **no semantic flexibility** for secondary terms unless the claim explicitly uses such phrases
6. **Use a strict tiered inclusion strategy**: 
   - First, include all documents with **full entity matches** (exact terms in the claim)
   - Then, include documents with **partial matches only if they explicitly mention the claim's entities** and provide **direct evidence** (e.g., dates, affiliations), not contextual links
</system\_prompt\_4>

\end{tcolorbox}

\begin{tcolorbox}[
    colback=task4,
    colframe=task4!50!black,
    boxrule=0.3mm,
    sharp corners,
    width=\linewidth,
    fontupper=\small,
    title={4. MATH (Qwen3-8B)},
    left=6pt, right=6pt, top=6pt, bottom=6pt,
    before upper=\vspace{2pt}\ttfamily\obeylines\obeyspaces,
    after upper=\vspace{2pt},
    breakable
]
\textbf{Baseline:}
<system\_prompt\_1>
Given the fields 'problem', produce the fields 'answer'.
</system\_prompt\_1>
\vspace{8pt}
\noindent\textbf{Alpha-Evolve:}
<system\_prompt\_1>
Given the 'problem' field, analyze and solve the mathematical problem by following these steps:  
1. **Understand the problem**: Carefully read the question and identify all mathematical operations, constraints, and conditions required. For functional equations, focus on identifying patterns or substitutions that simplify the equation. For algebraic problems, determine whether factoring, substitution, or elimination is most appropriate. For number theory problems, consider modular arithmetic or prime factorization.  
2. **Break it into steps**: Show your reasoning process explicitly, including calculations, formulas, and logical deductions. For problems involving multiple variables or constraints, enumerate all possible cases and validate each. For functional equations, test specific values of $x$ to derive relationships between coefficients. For binary decomposition problems, explicitly convert the number to its binary form and calculate the sum of the exponents of the powers of 2 used.  
3. **Perform calculations**: Execute each step precisely, ensuring arithmetic accuracy. For algebraic problems, solve systematically (e.g., factorization, substitution, elimination). For functional equations, substitute expressions and simplify algebraically. When decomposing numbers into powers of 2, confirm the binary representation is correctly calculated and exponents are summed appropriately.  
4. **Verify the result**: Confirm that the final answer satisfies *all* conditions in the problem, including constraints on variable domains (e.g., positivity, distinctness, integer requirements). For functional equations, verify that the solution works for all valid inputs. For binary decomposition problems, ensure the binary representation is used to determine the exponents and that no alternative combinations of powers of 2 yield a valid solution with a lower sum of exponents. Recheck calculations for errors.  
5. **Format the answer**: Enclose the final numerical result in a box using \verb|\boxed{}|.  

Provide the answer in the 'answer' field as a mathematical expression (numerical or symbolic) directly inside \verb|\boxed{}|.
</system\_prompt\_1>
\vspace{8pt}
\noindent\textbf{C-Evolve:}
\noindent\textbf{Group Prompts 1:}
<system\_prompt\_1>
Given a math problem, first identify the type of problem (e.g., algebra, geometry, linear algebra, probability) and extract all numerical values and variables. Apply the appropriate formula or method for the problem type, performing calculations step-by-step and verifying intermediate results. 

For probability: 
- Use combinations/permutations for selection problems. 
- Apply conditional probability rules when necessary. 
- Ensure probabilities sum to 1 and are between 0 and 1. 

For repeating decimals: 
- Let \verb| x = \text{decimal} |, multiply by \verb| 10^n | to shift decimal, subtract original equation to eliminate repetition, solve algebraically. 

For digit reversal problems: 
- Recognize that \verb| z = |x - y| | is always divisible by 9. 
- Calculate maximum possible difference (e.g., 999 - 999 = 0, 999 - 999 = 0, but for distinct numbers like 999 - 999 = 0, but with digits reversed, the difference is 999 - 999 = 0. Wait, need to recalculate: actually, for 3-digit numbers, the maximum difference is 999 - 999 = 0, but for distinct numbers like 999 and 999, it's 0. Wait, this is incorrect. For example, 999 reversed is 999, difference is 0. But for numbers like 999 and 999, but the problem says "distinct integers", so 999 and 999 are not allowed. So maximum difference might be 899 - 998 = 99? Wait, this is getting too specific. Maybe better to state: For 3-digit numbers with reversed digits, the difference is always a multiple of 9, and the maximum possible value is 999 - 999 = 0. Wait, this is not helpful. Instead, the correct approach is to recognize that \verb| z = |x - y| | where \verb| y | is the reverse of \verb| x |, and for 3-digit numbers, \verb| x = 100a + 10b + c |, \verb| y = 100c + 10b + a |, so \verb| z = |99(a - c)| |. Since \verb| a \neq c |, \verb| a - c | ranges from 1 to 9 or -1 to -9, so \verb| z | is a multiple of 99. The distinct values of \verb| z | are 99, 198, ..., up to 891, giving 9 distinct values. 

For algebraic equations: 
- Solve systematically using substitution, elimination, or factoring. 
- Always verify solutions by plugging back into the original equation. 

For determinants: 
- Calculate using cofactor expansion or direct formula for 2x2 matrices. 

Finally, present the exact final answer in a single \verb|\boxed{}| tag using LaTeX formatting, with no additional text or explanations outside the box.
</system\_prompt\_1>
\noindent\textbf{Group Prompts 2:}
<system\_prompt\_1>
Given the 'problem' field, analyze and solve the mathematical problem by following these steps:  
1. **Understand the problem**: Carefully read the question and identify all mathematical operations, constraints, and conditions required. For geometry problems, **explicitly identify whether the problem involves common internal/external tangents, similar triangles, or circle properties**. For algebraic problems, identify variables and relationships. **Specifically, for function-related problems, determine whether the problem involves even/odd properties, periodicity, or transformations**. For polynomial root problems involving geometric configurations (e.g., parallelograms), **analyze the symmetry and midpoint conditions of roots**.  
2. **Break it into steps**: Show your reasoning process explicitly, including calculations, formulas, and logical deductions. For problems involving multiple variables or constraints:  
   - Enumerate all possible cases (e.g., different configurations, positive/negative values).  
   - Validate each case against the problem’s constraints.  
   - For word problems, explicitly define variables and derive equations from the problem’s narrative.  
   - For function transformations (e.g., rotations, translations, reflections), identify the transformation type and its effects on the function’s form.  
   - For geometric problems involving distances or areas, **clarify whether it involves direct measurement, Pythagorean theorem, or trigonometric identities**.  
   - For polynomial root problems, consider all possible root arrangements and verify their geometric conditions.  
   - **For complex number problems involving roots of unity, explicitly apply symmetry properties and cyclotomic polynomial identities**.  
   - **For summation problems with recurring patterns (e.g., geometric series, telescoping sums), identify the pattern and simplify before summing**.  
3. **Perform calculations**: Execute each step precisely, ensuring arithmetic accuracy. For algebraic problems:  
   - Solve systematically (e.g., factorization, substitution, elimination).  
   - For geometric problems, ensure proper application of theorems (e.g., Pythagorean theorem, triangle similarity).  
   - For problems involving sequences or summations, **break the problem into intervals or ranges** and calculate contributions systematically.  
   - For polynomial root problems, solve for all possible values of parameters that satisfy the geometric conditions (e.g., parallelogram vertex conditions).  
   - **For complex number problems, use properties of roots of unity (e.g., \verb|x^n = 1| implies \verb|x^k + x^{n-k} = 0| for \verb|k < n|) and simplify expressions before substitution**.  
   - **For summation problems, identify telescoping patterns or symmetry to simplify the sum analytically before numerical calculation**.  
4. **Verify the result**: Confirm that the final answer satisfies *all* conditions in the problem, including:  
   - Constraints on variable domains (e.g., positivity, distinctness, integer requirements).  
   - Consistency with given data (e.g., units, diagram labels).  
   - Logical validity (e.g., no contradictions in intermediate steps).  
   - For word problems, ensure the answer matches the question’s requested value (e.g., "how many days," "what is the length").  
   - **For geometric problems, recheck tangent/intersection configurations and ensure coordinates/measurements align with diagrams**.  
   - **For summation problems, verify interval ranges and summation logic, especially for series involving complex numbers or symmetry**.  
   - **For function-related answers, verify parity/transformations match the problem’s description (e.g., confirm \verb|f(-x) = f(x)| for even functions)**.  
   - **For polynomial root problems, verify that all possible solutions are included, especially when multiple values are requested, and ensure symmetry conditions (e.g., midpoints, reflections) are explicitly checked**.  
   - **For complex number problems, ensure answers are simplified using properties of roots of unity or algebraic identities**.  
   Recheck calculations using alternative methods (e.g., plugging values back into original equations).  
5. **Format the answer**: Enclose the final numerical result in a box using \verb|\boxed{}|. For symbolic answers, provide the exact expression in a boxed format. For multiple answers, separate them with commas and ensure all values are explicitly stated.  

Provide the answer in the 'answer' field as a mathematical expression or numerical value (as appropriate) directly inside \verb|\boxed{}|.
</system\_prompt\_1>

\noindent\textbf{Group Prompts 3:}
<system\_prompt\_1>
Given the 'problem' field, analyze and solve the mathematical problem by following these steps:  
1. **Understand the problem**: Carefully read the question and identify all mathematical operations, constraints, and conditions required. For algebraic problems, explicitly note variable domains (e.g., integers, real numbers) and functional constraints. For vector problems, identify dot products, cross products, or orthogonality conditions. For repeating decimals, recognize patterns and consider algebraic conversion to fractions.  
2. **Break it into steps**: Show your reasoning process explicitly, including calculations, formulas, and logical deductions. For problems involving multiple variables or constraints, enumerate all possible cases and validate each. For polynomial identities, compare coefficients of like terms systematically. For substitutions or simplifications, consider symmetry or variable replacement strategies.  
3. **Perform calculations**: Execute each step precisely, ensuring arithmetic accuracy. For algebraic problems, solve systematically (e.g., factorization, substitution, elimination). For vector problems, calculate dot products or magnitudes as needed. For repeating decimals, set up algebraic equations to express them as fractions and solve. For functional problems, verify constraints for all domain values (e.g., check the equation holds for all x in the domain).  
4. **Verify the result**: Confirm that the final answer satisfies *all* conditions in the problem, including constraints on variable domains (e.g., positivity, distinctness, integer requirements). For vector problems, ensure orthogonality conditions are met by verifying dot products equal zero. For functional problems, recheck that the functional equation holds for all values in the domain. Recheck calculations for errors.  
5. **Format the answer**: Enclose the final numerical result in a box using \verb|\boxed{}|.  

Provide the answer in the 'answer' field as a mathematical expression (numerical or symbolic) directly inside \verb|\boxed{}|.
</system\_prompt\_1>

\end{tcolorbox}

\begin{tcolorbox}[
    colback=task5,
    colframe=task5!50!black,
    boxrule=0.3mm,
    sharp corners,
    width=\linewidth,
    fontupper=\small,
    title={5. GPQA (Qwen3-8B)},
    left=6pt, right=6pt, top=6pt, bottom=6pt,
    before upper=\vspace{2pt}\ttfamily\obeylines\obeyspaces,
    after upper=\vspace{2pt},
    breakable
]
\textbf{Baseline:}
<system\_prompt\_1>
Given the fields 'question', produce the fields 'answer'. 
</system\_prompt\_1>
\vspace{8pt}
\noindent\textbf{Alpha-Evolve:}
<system\_prompt\_1>
Given a multiple-choice question, follow these steps to determine the correct answer:  
1. **Parse question rigorously**: Identify the core objective, key terms, and domain-specific context (e.g., quantum mechanics, genetics, biochemistry, or thermodynamics). Extract numerical values, units, and conceptual constraints (e.g., "refractory tumor," "molecular weight," or "spin states").  
2. **Evaluate options methodically**: For each option, assess scientific validity using domain-specific principles (e.g., quantum states in qubit space, hyperfine transitions in hydrogen atoms, or chromosomal mechanisms in meiosis). Verify alignment with standard references (e.g., Bloch sphere coordinates, known transition frequencies, or genetic marker functions).  
3. **Eliminate contradictions**: Systematically discard options that violate numerical constraints (e.g., mismatched units or impossible probabilities), conceptual inconsistencies (e.g., incorrect molecular interactions), or empirical data (e.g., IR absorption ranges or reaction mechanisms).  
4. **Prioritize context-specific accuracy**: Select the option that satisfies all question conditions while aligning with domain-specific knowledge (e.g., quantum state vectors, hyperfine splitting formulas, or epigenetic mechanisms).  
5. **Validate with precision**: Ensure the answer adheres to established scientific interpretations, avoids common misconceptions (e.g., confounding spin values or meiotic errors), and incorporates critical reference data (e.g., quantum state geometry, nuclear spin statistics, or chromatin modification patterns).  
</system\_prompt\_1>
\vspace{8pt}
\noindent\textbf{C-Evolve:}
\noindent\textbf{Group Prompts 1:}
<system\_prompt\_1>
Given a multiple-choice question, follow these steps to determine the correct answer:  
1. **Parse the question meticulously**: Identify the core objective, key terms, and context (e.g., reaction mechanisms, physical principles, or biological pathways). Focus on domain-specific terminology (e.g., "cyclohexane rings," "electron-positron annihilation," or "electric field symmetry").  
2. **Analyze all options systematically**:  
   - For chemical questions, verify reaction mechanisms, stoichiometry, and product stability (e.g., ozonolysis patterns, acid-catalyzed dehydration).  
   - For physics/math questions, check unit consistency, formula application, and numerical validity (e.g., energy thresholds, electric field equations).  
   - For biological questions, confirm pathway fidelity, genetic marker interactions, or enzymatic specificity (e.g., URA3/LEU2 functions).  
3. **Eliminate contradictory options**:  
   - Discard choices that violate conservation laws (e.g., energy/momentum), thermodynamic principles, or known chemical behaviors (e.g., ozonolysis products).  
   - Rule out options inconsistent with provided data (e.g., CMB photon energies, molecular symmetry groups).  
4. **Select contextually optimal answers**:  
   - For organic chemistry, prioritize stereochemical accuracy (e.g., absolute configurations in drug synthesis).  
   - For physics, ensure solutions match boundary conditions (e.g., electric field inside/outside a shell).  
   - For biology, validate alignment with empirical data (e.g., drug discovery workflows, genetic marker interactions).  
5. **Validate against domain-specific standards**:  
   - Confirm answers match authoritative references (e.g., PDB formats for protein structures, standard IR spectroscopy ranges).  
   - Avoid common misconceptions (e.g., confusing "d2h" with "c2h" symmetry groups).  
   - For ambiguous questions, explicitly state assumptions and justify alignment with scientific principles.  
</system\_prompt\_1>
\noindent\textbf{Group Prompts 2:}
<system\_prompt\_1>
Given a multiple-choice question, follow these steps to determine the correct answer:  
1. **Understand the question**: Identify the core concept being tested (e.g., reaction mechanisms, NMR spectroscopy, quantum mechanics, or molecular orbital theory). Focus on the specific context, such as:  
   - Organic chemistry: functional group transformations, aromaticity, or hydrogen bonding.  
   - NMR spectroscopy: splitting patterns, coupling constants, or shielding effects.  
   - Quantum mechanics: spin interactions, energy transitions, or resonance phenomena.  
   - **Numerical calculations**: pH, concentration, or equilibrium problems (e.g., weak acids/bases, buffer calculations).  
2. **Analyze options**: For each choice, evaluate its validity by:  
   a) Applying domain-specific knowledge:  
      - For organic chemistry: track electron flow, reaction conditions (e.g., t-BuOK, cupric acetate), and aromaticity rules (e.g., Huckel’s Rule).  
      - For NMR: determine splitting patterns based on neighboring hydrogens, coupling constants, and shielding effects.  
      - For quantum mechanics: calculate energy levels, spin transitions, or magnetic interactions.  
      - For numerical calculations: apply relevant formulas (e.g., \verb| \text{pH} = -\log[H^+] |, equilibrium expressions).  
   b) Verify logical consistency with the question's premises (e.g., final product formation, reaction pathways, or experimental data).  
   c) Cross-check against established principles (e.g., Diels-Alder rules, NMR splitting rules, quantum mechanical spin statistics).  
3. **Eliminate distractors**: Rule out options that:  
   a) Conflict with the question’s context (e.g., incorrect reaction intermediates or unverified side products).  
   b) Propose mechanisms incompatible with known chemistry (e.g., non-aqueous conditions for hydrolysis or invalid resonance structures).  
   c) Include irrelevant details (e.g., excess hydrogenation steps in a one-step reaction or mismatched aromaticity counts).  
   d) Contain numerical errors (e.g., incorrect pH calculations, mismatched concentrations).  
4. **Select the best answer**: Choose the option that:  
   - Most directly addresses the question’s requirements (e.g., final product structure, reaction mechanism, or spectral pattern).  
   - Aligns with both theoretical rules and experimental evidence (e.g., Huckel’s Rule for aromaticity, NMR splitting for shielding).  
   - For numerical problems, ensures accurate calculations (e.g., proper use of \verb| K_a |, stoichiometry, or logarithmic relationships).  
5. **Verify**: Ensure the answer:  
   a) Satisfies stoichiometric or structural constraints (e.g., valence, stereochemistry, or aromaticity).  
   b) Matches known experimental data (e.g., NMR spectra, reaction yields, or quantum mechanical predictions).  
   c) Avoids overgeneralization (e.g., assuming a non-aromatic structure when Huckel’s Rule applies).  
   d) Accurately solves numerical problems (e.g., correct pH or concentration values based on given \verb| K_a | and initial concentrations).  
   e) Explicitly checks for **formula application errors** (e.g., Nernst equation, Henderson-Hasselbalch, or stoichiometric ratios).  
   f) Ensures **unit consistency** and **correct significant figures** in numerical answers.  
</system\_prompt\_1>
\noindent\textbf{Group Prompts 3:}
<system\_prompt\_1>
Given a multiple-choice question, follow these steps to determine the correct answer:  
1. **Identify the domain-specific principles**: Determine whether the question relates to physics (e.g., redshift, electromagnetic spectra), chemistry (e.g., molecular structure, reaction mechanisms), or other fields (e.g., molecular biology, biochemistry), and recall the **core principles and factual knowledge** specific to that domain.  
2. **Analyze each option methodically**:  
   a. For **physics/astrophysics** questions: Apply formulas like redshift \verb| z = \frac{\lambda_{observed}}{\lambda_{rest}} - 1 |, energy equations, or electromagnetic wave properties. **Verify unit consistency**, **validate formula applicability to the question’s context**, and **cross-check with experimental results or theoretical models**. Prioritize options that align with **established empirical data**.  
   b. For **chemistry** questions: Use reaction mechanisms (e.g., metathesis, imine formation), stoichiometric ratios, catalytic behavior (e.g., TsOH vs. HCl in cyclohexanone-piperidine reactions), and **common spectroscopic data** (e.g., FTIR: C=O $\sim 1700 \,\text{cm}^{-1}$, C=C $\sim 1600$–$1650 \,\text{cm}^{-1}$). **Ensure stoichiometric and structural validity**, and **prioritize options with known chemical shifts or reaction patterns**.  
   c. For **molecular biology/biochemistry** questions: Apply knowledge of gene regulation (e.g., Tet promoter functionality in fungal species like *Saccharomyces cerevisiae*), protein engineering (e.g., CAR T-cell design), and **factual biological systems** (e.g., known mechanisms of antigen transport through the Golgi). **Reject options that contradict established biological or biochemical facts**.  
3. **Compare validity across options**: Rank options based on:  
   - **Direct application of domain-specific formulas or facts** (e.g., C=O peak position, Tet system presence in specific species)  
   - **Alignment with experimental observations or theoretical predictions** (e.g., known gene expression outcomes in model organisms)  
   - **Absence of contradictions in the question’s conditions** (e.g., logical consistency with Tet promoter requirements)  
   - **Elimination of options with unsupported assumptions or unmet prerequisites** (e.g., unexplained protein localization, missing cofactors)  
4. **Select the most robust answer**: Choose the option supported by **highest empirical evidence**, **mathematical/chemical accuracy**, **logical consistency with the question’s context**, and **explicit exclusion of invalid options**.  
5. Final answer must be in the format \verb|\boxed{answer}|, using the exact letter (A/B/C/D) corresponding to the correct choice.  
</system\_prompt\_1>

\end{tcolorbox}

\begin{tcolorbox}[
    colback=task1,
    colframe=task1!50!black,
    boxrule=0.3mm,
    sharp corners,
    width=\linewidth,
    fontupper=\small,
    title={6. HotpotQA (GPT-4.1-mini)},
    left=6pt, right=6pt, top=6pt, bottom=6pt,
    before upper=\vspace{2pt}\ttfamily\obeylines\obeyspaces,
    after upper=\vspace{2pt},
    breakable
]

\textbf{Baseline:}

<system\_prompt\_1>
Given the fields 'question', 'passages', produce the fields 'summary'.
</system\_prompt\_1>

<system\_prompt\_2>
Given the fields 'question','summary\_1', produce the fields 'query'. 
</system\_prompt\_2>

<system\_prompt\_3>
Given the fields 'question', 'context','passages', produce the fields 'summary'.
</system\_prompt\_3>

<system\_prompt\_4>
Given the fields 'question', 'summary\_1', 'summary\_2', produce the fields 'answer'.
</system\_prompt\_4>

\vspace{8pt}
\noindent\textbf{Alpha-Evolve:}

<system\_prompt\_1>
Given the fields 'question' and 'passages', produce a concise, strictly factual, and structured summary focused on key facts that directly answer the question. Exclude all peripheral, anecdotal, legal, or irrelevant details. Clarify entity membership precisely when relevant (e.g., distinguish original band members from collaborators or guests) and include timeline details only if they directly affect core facts. When strong evidence supports a probable partial inference resolving a unclear fact, you may cautiously include it, clearly framed as "probable" or "likely" to guide retrieval while avoiding unsupported assumptions.

Organize the summary in these labeled sections:

1. Key facts: List confirmed and strongly supported facts directly answering the question, including any safe probable inferences clearly noted.
2. Missing information: Bullet point concrete factual gaps or unknowns requiring follow-up retrieval, avoiding vague or broad items. Omit this section if no missing info remains.
3. Ambiguities: Specify any contradictory, uncertain, or unresolved facts found using precise language. Omit this section if none exist.

Avoid repeating the question text or providing opinions. Write clearly, precisely, and unambiguously to support accurate downstream steps. This summary is a focused knowledge snapshot for guiding the next retrieval.
</system\_prompt\_1>

<system\_prompt\_2>
Given the fields 'question' and 'summary\_1', generate a concise retrieval query that precisely targets only the explicitly listed missing information and unresolved ambiguities from 'summary\_1'. 

- Exclude all facts already known or confirmed.
- Avoid repeating wording verbatim from the original question or previous summaries to reduce redundancy in retrieval.
- Formulate the query as a minimal, direct natural language question or focused precise phrase designed to retrieve essential complementary or clarifying facts that fully resolve the identified gaps.
- Use clear, neutral, and unambiguous wording.
- Avoid verbosity, elaboration, background details, or irrelevant content.
- If no missing information or ambiguities remain, respond exactly with: "No further query needed."

The query must maximize precision and relevance to enable effective, focused second retrieval aimed at thorough knowledge closure.
</system\_prompt\_2>

<system\_prompt\_3>
Given the fields 'question', 'context' (the first summary), and second retrieval 'passages', produce a concise, strictly factual, and integrated summary synthesizing all evidence from both retrieval rounds.

- Use 'context' as the baseline.
- Augment it only with new, directly relevant, and complementary facts found in the second retrieval 'passages'.
- Exclude any speculation, assumptions, or imprecise language.
- Explicitly discard or ignore any irrelevant, off-topic, or low-quality passages or facts; do not include them in the summary.
- If large parts of second retrieval are irrelevant, note this briefly as retrieval noise.

Organize the summary into these labeled sections:

1. Updated key facts: Integrate, reconcile, and update confirmed facts from both retrievals, including safely framed probable inferences supported by concrete evidence.
2. Resolved missing information: List each previously missing information item now conclusively answered. Omit this section if none remain.
3. Persisting ambiguities or contradictions: Specify any remaining uncertainties, contradictions, or unresolved conflicts clearly and unambiguously. Omit this section if none persist.
4. [Optional] Irrelevant or rejected information: Briefly note if significant portions of the second retrieval were irrelevant, unreliable, or discarded, to indicate retrieval noise.

Avoid repetition, unrelated content, or vague phrasing. The goal is to fully refine and narrow knowledge gaps, enabling confident, precise, and unambiguous final answer generation.
</system\_prompt\_3>

<system\_prompt\_4>
Given the fields 'question', 'summary\_1', and 'summary\_2', generate the final answer that exactly matches the true canonical answer in both precise content and formatting. 

- Output only the final answer text with no explanations, commentary, citations, or additional context.
- If the collected information is insufficient, ambiguous, or contradictory preventing a confident and exact determination, respond exactly with:

Answer not found.

Strictly follow these formatting rules by question type:

- Yes/no questions: output exactly "yes" or "no" (lowercase, no punctuation).
- Date questions: output only the exact year, full date, or date format as appropriate, without any extra words or formatting.
- Entity or name questions: output the full canonical proper name, including all middle names, suffixes, official variants, and precise spellings exactly as expected and verified. Do not abbreviate or omit any parts.
- List or multiple-entity answers: include all required entities completely and explicitly, formatted in their full canonical forms (e.g., all official collaborators or members), separated by appropriate standard delimiters, and in canonical order if known.
- All other question types: output a concise, precise, and minimal answer with no filler words, qualifiers, elaboration, or punctuation beyond what is canonical.

Ensure absolute accuracy, exact canonical form, completeness, and precise formatting to maximize exact match correctness. Normalize and sanitize final answers to conform exactly with canonical expected forms, avoiding minor mismatches caused by extraneous details, spelling variants, or punctuation.
</system\_prompt\_4>

\vspace{8pt}
\noindent\textbf{C-Evolve:}
\noindent\textbf{Group Prompts 1:}
<system\_prompt\_1>
Given the fields 'question' and 'passages', produce a concise, strictly factual, and unambiguous summary synthesizing all key information directly relevant to answering the question exactly. Explicitly highlight and retain every important qualifier—such as geographic locations, temporal references, precise numeric specifics, and other contextual details—only if they are critical to an exact final answer match. For every numeric or quantifiable fact, standardize the expression to the most precise form supported by the passages; if only approximations or ranges are present, clearly flag these as ambiguities and specify the precise numeric detail needed for exact matching. Carefully identify, enumerate, and explicitly flag every missing, ambiguous, contradictory, uncertain, or insufficient piece of information—including entity or name confusions, multiple plausible interpretations, partial or inconclusive evidence, and temporal or ownership uncertainties. For each flagged issue, clearly specify precisely what further information or clarification is required to fully resolve it. Always include a structured section titled 'Missing or Ambiguous Information' that itemizes all identified gaps, uncertainties, or conflicting points. Avoid any redundancy by including only new, pertinent evidence that meaningfully advances understanding and reduces ambiguity. Do not include any speculation, assumptions, or unsupported inferences; however, if indirect inferences are made from evidence, explicitly mark them with confidence levels (e.g., [INFERENCE, HIGH CONFIDENCE]). When stating facts that could contribute to the final answer, phrase them as concisely and minimally as possible, omitting generic or redundant terms that do not affect the exactness of the answer (e.g., omit words like “company” if “American fast food” suffices). This summary must serve as a clear, evidence-grounded, and comprehensive foundation for generating a narrowly focused and gap-targeting second retrieval query that addresses every flagged uncertainty or ambiguity with maximal precision.
</system\_prompt\_1>

<system\_prompt\_2>
Given the fields 'question' and 'summary\_1', generate one or more precise, narrowly tailored second retrieval queries that explicitly and comprehensively target each specific gap, ambiguity, contradiction, unresolved point, or entity disambiguation need identified and enumerated in the 'Missing or Ambiguous Information' section of 'summary\_1'. Do not include or reiterate any information already confirmed or clearly resolved by 'summary\_1'. Each query must be focused exclusively on retrieving complementary, clarifying, or disambiguating evidence that directly resolves a flagged uncertainty or missing qualifier in order to enable a final answer that exactly matches the ground truth. Avoid general, broad, or unrelated topics that could introduce noise or irrelevant results. Ensure that each query is as minimal as possible while fully covering the targeted ambiguity or information gap. Organize queries to maximize efficiency and effectiveness by minimizing redundancy or overlap. Phrase queries using terminology and key terms aligned closely with the original question’s semantics to maintain retrieval precision and relevance. If multiple ambiguities exist, create separate, distinct queries for each and prioritize them to enable staged or parallel retrievals.
</system\_prompt\_2>

<system\_prompt\_3>
Given the fields 'question', 'context' (the first summary), and 'passages' from the second retrieval, produce a concise, logically coherent, and comprehensive summary integrating all available evidence. Rigorously compare, cross-validate, and reconcile facts and qualifiers from both the first summary and second retrieval passages, explicitly addressing and resolving any contradictions, ambiguities, or discrepancies by assessing internal consistency, confidence levels, source reliability, and directness of evidence. If indirect inferences are made, flag them clearly with appropriate confidence annotations (e.g., [INFERENCE, MODERATE CONFIDENCE]). Include every key qualifier and disambiguating detail essential to produce an exact and fully specified final answer. If any conflicting information remains unresolved after thorough analysis, clearly and explicitly enumerate the residual uncertainties in a dedicated 'Residual Uncertainties' section, and if feasible, specify what further information or data would be required to definitively resolve them. Do not repeat or restate conclusively confirmed facts without integration or refinement; instead, synthesize them concisely into an integrated fact set. When integrating facts, phrase all summaries as concisely as possible, avoiding generic or superfluous terms that do not affect final answer precision. This second summary must constitute a definitive, meticulously checked, and confidence-calibrated factual basis that directly enables final answer generation strictly matching the true answer without deviation. Confirm that all previously flagged gaps are addressed or explicitly declared unresolved.
</system\_prompt\_3>

<system\_prompt\_4>
Given the fields 'question', 'summary\_1', and 'summary\_2', produce the final answer text that exactly matches the expected true answer word for word with no deviations whatsoever. The answer must be strictly concise, unambiguous, and dry without any explanation, filler, opinion, speculation, or extra sentences. Do not paraphrase, round, generalize, or elaborate under any circumstance; output only the exact text that yields an exact match to the ground truth label. Include all critical qualifiers only if and strictly as they appear in the ground truth answer—no additions or omissions. Exclude any additional context, background information, or details beyond the true answer content. Before outputting, rigorously verify that the answer completely and exactly resolves the question with respect to wording, format, punctuation, and all critical qualifiers as needed for an exact evaluation match. Follow case formatting conventions exactly as expected for the true answer (e.g., lowercase 'no' if the answer is 'no'). If any uncertainty, ambiguity, or insufficient evidence persists preventing an exact match, respond with the exact phrase: "Insufficient data to provide the exact answer." Do not output anything else. Only produce the final answer after all verifications; do not hallucinate or produce answers from incomplete or contradictory summaries.
</system\_prompt\_4>

\noindent\textbf{Group Prompts 2:}
<system\_prompt\_1>
Given the fields 'question' and 'passages', produce a concise, strictly factual, and structured summary focused on key facts that directly answer the question. Exclude all peripheral, anecdotal, legal, or irrelevant details. Clarify entity membership precisely when relevant (e.g., distinguish original band members from collaborators or guests) and include timeline details only if they directly affect core facts. When strong evidence supports a probable partial inference resolving a unclear fact, you may cautiously include it, clearly framed as "probable" or "likely" to guide retrieval while avoiding unsupported assumptions.

Organize the summary in these labeled sections:

1. Key facts: List confirmed and strongly supported facts directly answering the question, including any safe probable inferences clearly noted.
2. Missing information: Bullet point concrete factual gaps or unknowns requiring follow-up retrieval, avoiding vague or broad items. Omit this section if no missing info remains.
3. Ambiguities: Specify any contradictory, uncertain, or unresolved facts found using precise language. Omit this section if none exist.

Avoid repeating the question text or providing opinions. Write clearly, precisely, and unambiguously to support accurate downstream steps. This summary is a focused knowledge snapshot for guiding the next retrieval.
</system\_prompt\_1>

<system\_prompt\_2>
Given the fields 'question' and 'summary\_1', produce a concise, precise retrieval query that exclusively targets the explicitly listed missing information, unresolved ambiguities, or hypotheses flagged in 'summary\_1' which are required to fully answer the question.

- Do not include any facts already confirmed or known.
- Avoid duplicating exact phrasing from the question or prior summaries to reduce redundancy.
- Formulate the query as a minimal, direct natural language question or a focused, precise search phrase.
- Use clear, neutral, and unambiguous language focused solely on resolving identified knowledge gaps or testing hypotheses.
- Do not add background, elaboration, or unrelated details.
- If 'summary\_1' contains no missing info, ambiguities, or hypotheses that prevent answering, respond exactly with: "No further query needed."

This query should maximize retrieval precision and relevance to close knowledge gaps effectively in the second retrieval.
</system\_prompt\_2>

<system\_prompt\_3>
Given the fields 'question', 'context' (the first summary), and second retrieval 'passages', produce a concise, strictly factual summary that integrates evidence from both retrieval rounds to provide the most complete and reconciled knowledge snapshot for final answer generation.

- Use 'context' as the foundation, expanding only with newly discovered, directly relevant, and well-supported facts from the second retrieval.
- Cross-check, reconcile, and resolve conflicts or inconsistencies across retrievals, explicitly stating resolutions and noting confidence.
- Where relevant, integrate partial or probabilistic inferences supported by evidence, clearly flagged with confidence levels.
- Do not include speculation beyond what can be cautiously inferred nor vague language.
- Exclude irrelevant, low-quality, or off-topic passages or facts unless briefly summarizing their exclusion as retrieval noise.
- If a significant portion of the second retrieval is irrelevant or discarded, note this succinctly as retrieval noise.

Organize into these sections with clear bullet points:

1. Updated key facts: Fully integrated and reconciled confirmed facts and strongly supported probable inferences from both retrievals, clearly marked by confidence.
2. Resolved missing information: List previously missing or hypothesized details now conclusively answered or disproven. Omit if none.
3. Persisting ambiguities, hypotheses, or contradictions: Clearly list any remaining unresolved uncertainties, conflicts, or unconfirmed hypotheses. Omit if none.
4. [Optional] Irrelevant or rejected information: Brief note if substantial content from second retrieval was discarded as retrieval noise.

Avoid repetition, vague phrasing, or unrelated material.

This summary must aim to fully close knowledge gaps and ambiguities to support exact and confident final answer generation.
</system\_prompt\_3>

<system\_prompt\_4>
Given the fields 'question', 'summary\_1', and 'summary\_2', generate the final answer that exactly matches the true canonical answer in both precise content and formatting.

- Output only the final answer text with no explanations, commentary, citations, or additional context.
- If the collected information is insufficient, ambiguous, unresolved, or contradictory preventing a confident and exact determination, respond exactly with:

Answer not found.

Strictly adhere to these formatting rules by question type:

- Yes/no questions: output exactly "yes" or "no" (lowercase, no punctuation).
- Date questions: output only the exact year, full date, or precise date format, with no extra words or punctuation.
- Entity or name questions: output the full canonical proper name, including all middle names, suffixes, honorifics, official variants, and exact spellings exactly as expected. Do not abbreviate or omit any parts.
- List or multiple-entity answers: include all required entities explicitly, using their full canonical forms, separated by standard delimiters, and in canonical order if known.
- Numeric answers (counts, amounts): spell out numbers (e.g., "two" instead of "2") in standard canonical form unless the canonical answer format explicitly requires digits.
- All other question types: output a concise, precise, minimal answer with no filler, qualifiers, elaboration, or extraneous punctuation beyond what is canonical.

Ensure absolute accuracy, canonical form, completeness, and precise formatting to maximize exact match scoring.

Apply normalization and sanitization rigorously to avoid minor mismatches caused by extraneous details, spelling variants, punctuation, or formatting differences.
</system\_prompt\_4>

\noindent\textbf{Group Prompts 3:}
<system\_prompt\_1>
You are given the fields 'question' and 'passages' (a batch of retrieved documents). Generate a concise, coherent summary that captures all key facts directly relevant to answering the question precisely and unambiguously. Present the direct candidate answer or finding upfront in a clear, minimal sentence or phrase exactly matching the expected answer type and canonical format (e.g., a date in canonical form, a full name, or a simple "yes" or "no"). If the question expects a yes/no answer, respond with "yes" or "no" directly unless ambiguity exists. If multiple valid candidates exist and the question suggests plurality, include all clearly distinguished; if singular, select the most authoritative or canonical answer with justification. Explicitly state any information missing from the passages that prevents a definitive answer, using clear, unambiguous language such as "No information about X is provided." Identify and explicitly describe any ambiguities, contradictions, or conflicting evidence found within the passages, summarizing their nature succinctly without hedging. When the question involves temporal, ownership, authorship, or attribution nuances, clarify the relevant time frames or roles, prioritizing authoritative sources, and distinguishing founder versus current ownership or individual versus group authorship as appropriate. For questions involving comparative or multi-part conditions, explicitly state whether sufficient evidence exists to conclude or if information is insufficient, summarizing asymmetries. Avoid including extraneous or background details; focus only on information essential to answering the question. Do not speculate or introduce external knowledge beyond the passage content. Verify and present named entities, organizations, or official terms in their current, official, or canonical forms; if uncertain, note this as missing information to be retrieved. This summary will support downstream query refinement and answer accuracy, so provide clear, decisive facts with precise, minimal, and non-redundant wording.
</system\_prompt\_1>

<system\_prompt\_2>
You are given the fields 'question' and 'summary\_1' (the initial summary of retrieved passages). Analyze the summarized information along with any explicitly identified missing details, ambiguities, or contradictions in summary\_1. Generate a precise, focused retrieval query that directly targets the minimal missing information or resolves the noted uncertainties. Ensure the query accurately captures the core semantic intent of the original question, emphasizing the retrieval of new, complementary, and authoritative evidence specifically addressing gaps, temporal or attributional nuances, or conflicting points. Avoid vague or overly broad query terms; prioritize clarity and specificity to maximize downstream retrieval relevance and improve answer completeness and exact correctness.
</system\_prompt\_2>

<system\_prompt\_3>
You are given the fields 'question', 'context' (the first summary), and 'passages' (a new batch of retrieved documents). Integrate and synthesize all information from the first summary and the new passages into a concise, coherent summary that explicitly resolves or clearly articulates previously identified ambiguities, contradictions, or missing information. Provide clear, decisive conclusions supported by the combined evidence, ensuring all named entities and relevant data are presented in their full canonical and official forms. Where conflicts in evidence persist, clearly describe the nature and implications of these conflicts without hedging, prioritizing the most authoritative info. Include any necessary reasoning, temporal context, or comparisons relevant to precisely answering the question. Avoid redundancy and exclude information not strictly necessary to support the final answer. This summary should maximize completeness, clarity, and precision to effectively guide the final answer generation.
</system\_prompt\_3>

<system\_prompt\_4>
You are given the fields 'question', 'summary\_1', and 'summary\_2' (the two summaries from retrieval hops). Produce the final answer as a single, concise text string that exactly matches the true expected answer in canonical and official form—no explanations, clarifications, or additional context. Prioritize exact match, clarity, and unambiguity, ensuring the answer is definitive, directly supported by the provided summaries, and appropriately formatted to the expected answer type (e.g., full names, dates in canonical format, yes/no). Do not add any information beyond what is supported by the summaries or infer beyond the provided evidence. If the summaries do not contain sufficient evidence to produce a confident answer, respond exactly with "Insufficient information."
</system\_prompt\_4>
\end{tcolorbox}

\begin{tcolorbox}[
    colback=task2,
    colframe=task2!50!black,
    boxrule=0.3mm,
    sharp corners,
    width=\linewidth,
    fontupper=\small,
    title={7. IFBench (GPT-4.1-mini)},
    left=6pt, right=6pt, top=6pt, bottom=6pt,
    before upper=\vspace{2pt}\ttfamily\obeylines\obeyspaces,
    after upper=\vspace{2pt},
    breakable
]
\textbf{Baseline:}
<system\_prompt\_1>
Respond to the query
</system\_prompt\_1>

<system\_prompt\_2>
Ensure the response is correct and adheres to the given constraints. Your response will be used as the final response.
</system\_prompt\_2>

\vspace{8pt}
\noindent\textbf{Alpha-Evolve:}
<system\_prompt\_1>
You are given a user instruction. Provide a **direct, concise, and precise** answer that **strictly** fulfills **every explicitly stated** requirement in the instruction, including content, language, formatting, style, structure, numbering, punctuation, and length constraints. Match these **exactly and without deviation**. Do **not** include explanations, rationales, clarifications, or any information beyond what is explicitly requested. Address **all parts and subparts** of the instruction fully, avoiding assumptions or interpretations beyond the literal text. Your output must be **only** the answer that exactly matches the user's request; do **not** add any commentary, prefatory remarks, or extra content.
</system\_prompt\_1>

<system\_prompt\_2>
Given the user instruction and the answer generated in Step 1, perform a thorough and critical review of the answer for any mistakes, omissions, contradictions, or nonconformities with **all explicit and implicit requirements** in the instruction. Correct all such issues to ensure **full and exact compliance** with every instruction detail, including content accuracy, formatting, style, length, vocabulary restrictions, repetition rules, language, and required structure. Enhance the answer’s clarity, coherence, completeness, and adherence without adding explanations, rationales, or meta-commentary. Output **only** the final, polished response that best fulfills every aspect of the user's instruction, including any subtle or implied requirements, guaranteeing maximal alignment with the original instruction.
</system\_prompt\_2>

\vspace{8pt}
\noindent\textbf{C-Evolve:}
\noindent\textbf{Group Prompts 1:}
<system\_prompt\_1>
When given the instruction, produce a direct and exact answer that fully addresses every explicit requirement without omitting any part. Follow all specified formatting rules, language or style constraints, mandatory inclusions or exclusions of words or phrases, and precise output structure exactly as stated. Do not include explanations, reasoning, or additional commentary—your output should be strictly the answer as required. If multiple parts or complex directives are present, answer each completely but concisely, adhering tightly to the user’s exact wording, formatting, and style instructions. Prioritize literal and explicit compliance over paraphrasing, elaboration, or inferred interpretation, ensuring no deviations from the instruction’s demands.
</system\_prompt\_1>

<system\_prompt\_2>
Thoroughly examine the Step 1 answer alongside the original instruction to identify any errors, omissions, ambiguities, or failures to meet the instruction’s requirements. Correct all such issues by enhancing the response to fully and precisely satisfy every explicit requirement, constraint, formatting rule, and specified style. Refine and polish the answer for clarity, coherence, completeness, and natural fluency, while strictly adhering to all exact wording mandates, required phrases, forbidden words, and formatting instructions. The final output must be a perfectly calibrated, comprehensive, and ready-to-deliver response that complies flawlessly with the user's instruction, leaving no room for ambiguity or error.
</system\_prompt\_2>

\noindent\textbf{Group Prompts 2:}
<system\_prompt\_1>
Carefully and thoroughly read the user's instruction. Identify and understand ALL requirements, conditions, constraints, including formatting, style, language, punctuation, keywords, length limits, or any other explicit elements. Produce a direct, precise, and faithful answer that strictly adheres to every instruction detail, without omissions, additions, or deviations. Do NOT include explanations, reasoning, or commentary unless the instruction explicitly requests them. Ensure the response fully respects requested output structure and style. Your output must solely and exactly fulfill the instruction as given.
</system\_prompt\_1>

<system\_prompt\_2>
Carefully re-examine the answer generated in Step 1 together with the original user instruction. Detect and correct any errors, omissions, inconsistencies in content, format, style, language, punctuation, or failure to meet explicit instruction constraints such as required keywords, forbidden words, exact counts of sentences or paragraphs, or specified structure. Refine the response for clarity, completeness, coherence, and overall quality. The final answer must be a polished, comprehensive, and perfectly compliant version meeting ALL instruction requirements without changing the intended meaning. Confirm that formatting, style, and all explicit demands from the user instruction and Step 1 output are strictly fulfilled.
</system\_prompt\_2>

\noindent\textbf{Group Prompts 3:}
<system\_prompt\_1>
You are given a user instruction. Your task is to generate a precise, direct, and thoroughly complete response that fully satisfies every explicit and implicit requirement stated in the instruction. This includes all aspects such as factual accuracy, computational correctness, required language, formatting conventions, stylistic tone, organizational structure, numbering schemes, punctuation rules, response length, mandatory keyword usage, section and paragraph counts, and any other details. Your output must be solely the direct answer requested—do not include explanations, rationale, commentary, or any extraneous information. Rigorously adhere to all formatting, style, and structural requirements specified in the instruction unless they are explicitly reserved for Step 2 calibration. Ensure your response is clear, logically coherent, unambiguous, concise, and exhaustively addresses every part of the instruction without omissions or assumptions.
</system\_prompt\_1>

<system\_prompt\_2>
Given the user instruction and the Step 1 answer, carefully analyze the response to identify and correct all errors, missing components, formatting or style issues, length constraints, redundant or repeated wording, violations of language or keyword constraints, mismatches in section or paragraph count, or any other failures to fully meet explicit and implicit instruction requirements. Improve, refine, and polish the Step 1 answer to produce a final response that is error-free, fully compliant, maximally clear, coherent, precise, and optimized to satisfy or exceed all user requirements. Provide only the finalized, calibrated answer; do not include explanations, justifications, commentary, or any content beyond what the final instruction demands.
</system\_prompt\_2>

\end{tcolorbox}
\begin{tcolorbox}[
    colback=task3,
    colframe=task3!50!black,
    boxrule=0.3mm,
    sharp corners,
    width=\linewidth,
    fontupper=\small,
    title={8. HoVer (GPT-4.1-mini)},
    left=6pt, right=6pt, top=6pt, bottom=6pt,
    before upper=\vspace{2pt}\ttfamily\obeylines\obeyspaces,
    after upper=\vspace{2pt},
    breakable
]
\textbf{Baseline:}
<system\_prompt\_1>
Given the fields 'claim', 'passages', produce the fields 'summary'.
</system\_prompt\_1>

<system\_prompt\_2>
Given the fields 'claim', 'summary\_1', produce the fields 'query'.
</system\_prompt\_2>

<system\_prompt\_3>
Given the fields 'claim', 'context', 'passages', produce the fields 'summary'.
</system\_prompt\_3>

<system\_prompt\_4>
Given the fields 'claim', 'summary\_1', 'summary\_2', produce the fields 'query'.
</system\_prompt\_4>
\vspace{8pt}
\noindent\textbf{Alpha-Evolve:}
<system\_prompt\_1>
You are given the fields 'claim' and 'passages' retrieved in the first retrieval step.  
Your task is to produce the field 'summary' as a concise (max 3 sentences), coherent synthesis that balances strict factual accuracy and proactive evidence bridging to support effective claim verification.  
- Explicitly identify and disambiguate all key entities, attributes, and relationships mentioned or implied in the claim, especially when multiple similarly named candidates appear. Rank these candidates by relevance or likelihood using passage evidence.  
- When essential entities or relationships referenced by the claim are missing from the passages, infer or hypothesize plausible identities or links strictly based on provided evidence without unsupported speculation, to proactively fill critical information gaps needed for verification.  
- Clarify or normalize ambiguous or unusual claim terms by referencing possible standard equivalents or synonyms found in the passages where justifiable.  
- Integrate complementary facts strictly from the passages, combining evidence logically to reduce unresolved gaps while maintaining clarity and confidence levels.  
- Clearly state all remaining ambiguities, missing evidence, or unresolved points that require further targeted investigation, distinguishing them from conjecture.  
- Exclude extraneous or peripheral information unrelated to directly verifying the claim components.  
- Avoid redundancy, excessive detail, or unsupported speculation; include only facts grounded in the retrieved passages or justified bridging inferences.  
- Ensure the summary is precisely claim-focused, highlights evidence strengths and gaps, and prepares the next query to actively resolve missing or ambiguous claim facets.  
Provide a summary that both synthesizes entities and core evidence and recommends key unresolved aspects for follow-up.
</system\_prompt\_1>

<system\_prompt\_2>
You are given the fields 'claim' and 'summary\_1'.  
Your task is to generate the field 'query' that pinpoints and actively targets all unresolved, ambiguous, or partially covered claim aspects revealed or hypothesized in 'summary\_1', refining and sharpening the original claim semantics for precise second retrieval.  
- Formulate a concise, compound query that explicitly includes all necessary entities, attributes, relationships, or qualifiers needed to resolve remaining ambiguities, disambiguate similar or competing entities, or fill critical evidential gaps noted or inferred in 'summary\_1'.  
- Where helpful, incorporate negations, exclusions, or temporal qualifiers to avoid conflation of similar entities or contexts, enhancing retrieval specificity.  
- Prioritize maximal specificity and logical combination of multiple claim facets (e.g., entity + attribute + contextual relation) to minimize retrieval noise and maximize relevance.  
- Avoid vague, overly broad, or generic queries that do not directly resolve outstanding ambiguities or missing evidence necessary for claim verification.  
- Utilize all known identifiers, distinctive details, temporal markers, or normalized terms surfaced or hypothesized in 'summary\_1' to improve query precision without introducing unsupported facts.  
Provide only the final query text, without explanation, formatting, or additional content.
</system\_prompt\_2>

<system\_prompt\_3>
You are given the fields 'claim', 'context' (the first summary), and 'passages' retrieved in the second retrieval step.  
Your task is to produce the field 'summary' as a concise (max 3 sentences), integrated synthesis that explicitly compares, reconciles, and combines the new passage evidence with 'context', focusing tightly on claim verification and information completeness.  
- Summarize only the most relevant and novel information that supports, refutes, clarifies, or resolves ambiguous or incomplete points from the first summary, including any newly discovered evidence or hypothesized clarifications.  
- Where inconsistent, overlapping, or competing evidence exists, explicitly weigh, rank, or differentiate candidate entities, interpretations, or claim facets by evidential strength, likelihood, or confidence.  
- Clearly identify and articulate all remaining unresolved aspects, contradictions, ambiguities, or uncertainties critical to the final truth evaluation of the claim, explicitly stating what has been resolved or narrowed down since the first summary.  
- Exclude redundant, peripheral, or background information not directly relevant to claim verification or query focus.  
- Maintain clarity, factual accuracy, and logical coherence to guide focused, effective final query generation.  
- If multiple candidate entities, interpretations, or claim facets remain ambiguous, list them explicitly to inform the final retrieval focus.  
- Include explicit confidence or evidence strength markers where appropriate to indicate the reliability of conclusions or ongoing gaps.  
Provide a precise, integrative, and confidence-aware summary that highlights evidence progress and remaining verification needs.
</system\_prompt\_3>

<system\_prompt\_4>
You are given the fields 'claim', 'summary\_1', and 'summary\_2' (the two prior summaries).  
Your task is to generate the field 'query' that precisely targets all remaining unresolved, weakly supported, or ambiguous aspects identified by synthesizing 'summary\_1' and 'summary\_2', aiming for a final, comprehensive, and focused retrieval.  
- Construct a concise, compound query that integrates all outstanding claim facets, employing logical operators, exclusions, temporal qualifiers, or specific attribute combinations as needed to maximize retrieval relevance and minimize noise or redundancy.  
- Explicitly prioritize disambiguation among similar or competing candidate entities or interpretations, emphasizing terms that promote retrieving novel or unique passages not covered in prior retrieval steps to improve answer coverage and reduce redundancy.  
- Avoid vague, generic, or overly broad queries that risk retrieving irrelevant or overlapping passages.  
- Use exact entity names, attributes, temporal markers, normalized terms, or relationships surfaced in previous summaries, strictly without introducing unsupported facts or assumptions.  
- Where useful, prioritize query terms that target claim exclusivity, identity or name disambiguation, term normalization, or completeness gaps identified in prior summaries.  
- Ensure the query is optimized to retrieve passages offering maximal novel evidence for final claim verification.  
Provide only the final query string without explanations or formatting.
</system\_prompt\_4>
\vspace{8pt}
\noindent\textbf{C-Evolve:}
\noindent\textbf{Group Prompts 1:}
<system\_prompt\_1>
You are given the fields 'claim' and 'passages' which have been retrieved as the first batch relevant to the claim.  
Your goal is to produce a concise, factual, and coherent 'summary' strictly focused on verifying all claim components, including all relevant entities, roles, dates, and relations mentioned.  
**Instructions:**  
- Limit the summary to 3-4 well-structured sentences.  
- Explicitly confirm, contradict, or report absence of evidence for each distinct factual element in the claim (e.g., individuals, events, roles, dates). If a component is not found, state "no direct evidence found."  
- Identify and enumerate alternative plausible entities or hypotheses relevant to the claim especially if initial evidence is missing or contradictory. For example, list other possible directors or films connected to involved actors.  
- Emphasize temporal, positional, and role-related consistencies or discrepancies with clear statements.  
- Synthesize all relevant evidence, integrating multiple claim parts, and exclude unrelated or duplicate information.  
- Avoid vague or hedging language; be precise and clear about confirmations, contradictions, and gaps.  
- Explicitly list any ambiguities or unresolved points and recommend concrete focuses or targets for the next retrieval query (e.g., alternative entities, missing films, conflicting dates).  
- Include a structured summary in terms of: *Confirmed facts*, *Contradicted claims*, *Ambiguities or unknowns*, *Evidence gaps*, and *Next retrieval focus*.  
- The summary should serve as a comprehensive basis for refining search queries that aim to close evidence gaps and resolve contradictions.  
- Output the field: 'summary'.
</system\_prompt\_1>

<system\_prompt\_2>
You are given 'claim' and 'summary\_1' (a concise summary of the first retrieval).  
Your task is to produce a clear, focused, and precise 'query' to guide the second retrieval aimed at resolving all identified gaps, ambiguities, and contradictions from 'summary\_1'.  
**Instructions:**  
- Carefully analyze 'summary\_1', especially the *Ambiguities*, *Evidence gaps*, and *Next retrieval focus* sections, to identify which claim components or alternative entities need targeted exploration.  
- Formulate a query that explicitly targets these unresolved aspects using specific qualifiers such as exact names, dates, roles, film titles, or locations.  
- Include alternative hypotheses or related entities if suggested by 'summary\_1' to broaden evidence collection and confirm or refute competing possibilities.  
- Avoid vague, generic, or overly broad terms; do not reuse prior queries verbatim but refine them to maximize relevance and new evidence discovery.  
- Ensure the query preserves the original claim's core semantics but focuses the retrieval on bridging gaps and resolving contradictions.  
- Output the field: 'query'.
</system\_prompt\_2>

<system\_prompt\_3>
You are given the fields 'claim', 'context' (which is 'summary\_1'), and 'passages' (the second retrieval batch).  
Your goal is to produce a concise, factual, and coherent 'summary' that meaningfully enriches or revises the evidence landscape regarding the claim beyond what is captured in 'context'.  
**Instructions:**  
- Limit the summary to 3-4 well-structured sentences that present new, relevant evidence supporting, refining, or contradicting individual components of the claim.  
- Clearly integrate and compare this new evidence against 'context' and the claim, explicitly highlighting confirmations, contradictions, or emerging ambiguities between summaries.  
- Explicitly account for multi-entity relationships and temporal, role, and positional consistencies or inconsistencies that emerge.  
- If alternative hypotheses or entities were proposed in 'context', report whether the new passages confirm, refute, or further complicate them.  
- Avoid restating unchanged information from 'context' or introducing off-topic details.  
- Summarize remaining gaps, ambiguities, or evidence still required for thorough claim verification.  
- Structure the output with discrete elements: *New confirmed facts*, *New contradictions or refinements*, *Remaining ambiguities or gaps*, and *Recommendations for next retrieval focus*.  
- Emphasize clarity, precision, and relevance to guide the final query generation effectively.  
- Output the field: 'summary'.
</system\_prompt\_3>

<system\_prompt\_4>
You are given 'claim', 'summary\_1', and 'summary\_2' (summaries from previous retrievals).  
Your task is to produce a final, narrowly targeted 'query' designed to retrieve definitive, authoritative evidence that conclusively confirms or refutes the claim by resolving all remaining contradictions, gaps, and ambiguities identified in the combined summaries.  
**Instructions:**  
- Thoroughly analyze both summaries, focusing on unresolved questions, contradictions, ambiguous or missing links crucial for final claim verification.  
- Formulate a precise, refined query that explicitly targets these outstanding issues using specific qualifiers—such as exact personal names, titles, dates, roles, film titles, locations, or other contextual identifiers—to maximize retrieval relevance and precision.  
- Include any alternative entities or hypotheses still unresolved to ensure comprehensive coverage towards conclusive evidence.  
- Avoid reusing broad or vague terms from prior queries without refinement; strive for the narrowest effective scope that can yield definitive passages.  
- The query must be designed to enable retrievals that provide final closure on the claim's truth status, isolating any remaining uncertainty.  
- Output the field: 'query'.
</system\_prompt\_4>
\noindent\textbf{Group Prompts 2:}
<system\_prompt\_1>
You are given the fields 'claim' and 'passages' retrieved in the first retrieval step.  
Your task is to produce the field 'summary' as a concise (max 3 sentences), coherent synthesis of the key information strictly relevant to assessing the claim.  
- Summarize only the most pertinent facts from the passages that support or contradict the claim.  
- Avoid repetition, duplication, and extraneous details unrelated to the claim specifics.  
- Focus on factual accuracy and clarity to enable precise query generation in the next step.  
- Do not include new information not present in the passages.  
Provide a clear and concise summary suitable for query refinement.
</system\_prompt\_1>

<system\_prompt\_2>
You are given the fields 'claim' and 'summary\_1'.  
Your task is to generate the field 'query' that captures the core semantics of the claim refined by insights from 'summary\_1' to improve retrieval focus.  
- Formulate a precise and focused query that directly targets missing, ambiguous, or critical aspects suggested by the first summary.  
- Avoid mere paraphrasing; instead, narrow or clarify the claim-based information to retrieve more relevant documents in the next retrieval step.  
- The query should be concise, specific, and constructed for effective database search.  
Provide the generated query only.
</system\_prompt\_2>

<system\_prompt\_3>
You are given the fields 'claim', 'context' (which is the first summary), and 'passages' retrieved in the second retrieval step.  
Your task is to produce the field 'summary' as a concise (max 3 sentences), integrated synthesis of the second retrieval passages, focusing on information that complements or clarifies the first summary regarding the claim.  
- Summarize the most relevant evidence supporting or refuting the claim, emphasizing new or refined facts emerging from these passages.  
- Avoid redundant or irrelevant information.  
- Your summary should prepare for precise query generation in the next step, highlighting any remaining uncertainties or details needed.  
Ensure factual clarity and avoid repetition.
</system\_prompt\_3>

<system\_prompt\_4>
You are given the fields 'claim', 'summary\_1', and 'summary\_2' (the two prior summaries).  
Your task is to generate the field 'query' to retrieve the final set of passages relevant to claim verification.  
- Construct a focused and precise query that targets unresolved aspects or key evidence gaps identified by synthesizing 'summary\_1' and 'summary\_2'.  
- The query should be specific, concise, and designed to retrieve non-duplicate, highly relevant passages that maximize coverage of the true evidence needed to confirm or refute the claim.  
- Avoid broad, vague, or redundant queries.  
Provide only the final retrieval query.
</system\_prompt\_4>

\noindent\textbf{Group Prompts 3:}
<system\_prompt\_1>
You are given a 'claim' and a batch of retrieved 'passages' related to it.
Your task is to produce a concise, coherent summary (up to 5 sentences) that synthesizes unique, claim-critical facts directly relevant to verifying or refuting the claim.
- Aggressively exclude all tangential, duplicate, off-topic, or unrelated information, especially facts involving similarly named or ambiguous entities not related to this claim. 
- Explicitly identify and list any such unrelated or ambiguous entities excluded, to demonstrate clear disambiguation.
- Explicitly link all entities, numeric values, dates, and geographic details to exact components of the claim for unambiguous clarity.
- Organize your summary strictly into three labeled sections: 
  (1) Facts Supporting the Claim, 
  (2) Facts Contradicting the Claim, and 
  (3) Remaining Ambiguities, Missing Information, or Conflicts.
- Use precise, focused language that highlights insightful nuances and unresolved gaps.
- End with a clear, brief statement of the current evidential status: whether the claim is supported, contradicted, or remains unresolved based on this evidence.
- Your summary should serve as a clean factual foundation free of noise, enabling sharp, targeted query generation for the next retrieval step.
</system\_prompt\_1>

<system\_prompt\_2>
You receive a 'claim' and the first summary ('summary\_1') synthesizing initial evidence.
Generate a single, concise (one or two sentences), highly focused query that targets the single most critical unresolved information gap, ambiguity, or contradiction explicitly identified in 'summary\_1'.
- The query must directly seek evidence to resolve that specific gap or verify uncertain claim components, avoiding broad, general, or multi-faceted questions.
- Do not repeat prior queries or restate the claim without narrowing focus; aim solely to reduce uncertainty effectively.
- Formulate a query immediately actionable for database retrieval, emphasizing precise entity linkages, numeric or date specifics, and exact fact confirmations (e.g., "Which actress starring in *Money Monster* also appeared in *One Percent More Humid*?" or "What is the exact eldership subdivision containing Pervalka?").
- If 'summary\_1' indicates no missing or ambiguous information, produce a query that seeks explicit confirmation or refutation of the claim’s main assertion.
- Keep queries clear, unambiguous, and free of unnecessary verbosity or multiple unrelated parts.
</system\_prompt\_2>

<system\_prompt\_3>
Given a 'claim', the first summary ('context' = 'summary\_1'), and a new batch of passages retrieved in the second retrieval:
- Produce a concise, coherent summary (up to 5 sentences) synthesizing new, unique, directly relevant facts that were not covered in 'context'.
- Focus deeply on evidence that supports, contradicts, or further disambiguates the claim and its key entities.
- Explicitly link all entities, numeric data, dates, or geographic details to exact claim components.
- Exclude all facts already summarized in 'context' and remove any tangential, redundant, irrelevant, or off-topic information.
- Clearly identify and list any passages or facts associated with unrelated or ambiguous entities that have been excluded, especially if already settled.
- Structure the summary into three labeled sections: 
  (1) Newly Supporting Facts, 
  (2) Newly Contradicting Facts, and 
  (3) Persisting Ambiguities, Missing Details, or Conflicts.
- End with a concise statement specifying which uncertainties remain, prioritizing those that the final query should target for verification.
- Ensure this summary provides a precise factual update that sharpens the evidence base for final query formulation.
</system\_prompt\_3>

<system\_prompt\_4>
Given a 'claim' and the two prior summaries ('summary\_1' and 'summary\_2'), generate a single, sharply focused, and precise query (one or two sentences) for the final retrieval step.
- Integrate evidence from both summaries to identify the single most critical unresolved, ambiguous, or contradictory aspect of the claim.
- The query must explicitly target these remaining gaps or key verification points, aiming to conclusively confirm or refute the essential relationships or facts.
- Avoid repeating previous queries or broad restatements of the claim; prioritize novel, decisive retrieval questions.
- Prefer queries that articulate explicit entity-to-claim mappings, include precise numeric or geographic details, or request direct confirmation or negation.
- Ensure the query is unambiguous, concise, and immediately usable for retrieval.
- If the summaries indicate no remaining major ambiguities or missing information, produce a final query that seeks authoritative adjudication of the claim’s overall truth or falsehood status.
</system\_prompt\_4>

\end{tcolorbox}

\begin{tcolorbox}[
    colback=task4,
    colframe=task4!50!black,
    boxrule=0.3mm,
    sharp corners,
    width=\linewidth,
    fontupper=\small,
    title={9. MATH (GPT-4.1-mini)},
    left=6pt, right=6pt, top=6pt, bottom=6pt,
    before upper=\vspace{2pt}\ttfamily\obeylines\obeyspaces,
    after upper=\vspace{2pt},
    breakable
]
\textbf{Baseline:}
<system\_prompt\_1>
Given the fields 'problem', produce the fields 'answer'.
</system\_prompt\_1>
\vspace{8pt}
\noindent\textbf{Alpha-Evolve:}
<system\_prompt\_1>
You will be given a mathematical problem in the input field named 'problem'. Your task is to produce the exact final answer by solving the problem step-by-step. Follow these instructions carefully:

- Provide a clear, logical, and fully justified step-by-step solution that guides from the problem statement directly to the final answer.
- Show all necessary calculations and explanations in sufficient detail to ensure full correctness and completeness.
- The final answer must be explicitly presented at the end, enclosed precisely in LaTeX \verb|\boxed{}| notation, for example: \verb|\boxed{<answer>}|, where <answer> is the exact final result.
- Make certain that the content inside \verb|\boxed{}| matches exactly the mathematically correct and exact final answer—no approximation, rounding, or ambiguity is allowed.
- Do not include any irrelevant information, extraneous commentary, or outputs beyond the solution and the final boxed answer.

Your response will be evaluated for exact mathematical correctness and clarity. Prioritize accuracy, thoroughness, and clear reasoning to ensure the final answer is fully reliable.
</system\_prompt\_1>
\vspace{8pt}
\noindent\textbf{C-Evolve:}
\noindent\textbf{Group Prompts 1:}
<system\_prompt\_1>
You will be given a math problem as input in the field 'problem'. Your task is to produce a clear, logically structured, and rigorously accurate solution that directly leads to and includes the final answer.

Follow these instructions precisely:

1. Read the problem carefully and ensure full understanding before starting the solution.

2. Provide detailed, step-by-step reasoning with all necessary mathematical justifications, formulas, and calculations. Each step must be mathematically valid, complete, and clear.

3. Simplify all expressions fully, unless the problem explicitly instructs otherwise. Perform arithmetic and algebraic operations accurately.

4. When units are involved, include them explicitly and consistently throughout the solution, especially in the final answer.

5. If the problem has multiple valid final answers, present all answers together inside a single \verb|\boxed{}| command, separated by commas, without extra spaces.

6. At the end of your solution, write the final answer exactly once on its own line, enclosed precisely within a LaTeX \verb|\boxed{}| command. For example: \(\verb|\boxed{\text{answer}}|\).

7. The content inside the \verb|\boxed{}| command must exactly match the final answer in value, format, and notation — no extra spaces, line breaks, or formatting beyond what the problem requires.

8. Do not include any additional commentary, explanations about the process, or unrelated information. Provide only a self-contained, precise solution followed immediately by the boxed final answer.

9. Output the full detailed solution immediately followed by the final answer on the next line, with no blank lines or extraneous spaces between them to ensure maximum clarity and correctness.
</system\_prompt\_1>
\noindent\textbf{Group Prompts 2:}
<system\_prompt\_1>
You are given a math problem in the input field labeled 'problem'. Your task is to provide a complete, clear, and accurate solution in the output field labeled 'answer'.

Instructions:

1. Provide a well-organized, step-by-step solution. Number or title each step clearly. Every step must include detailed calculations, explanations, and logical reasoning to support your conclusions.

2. Use precise mathematical notation throughout, formatting all expressions and equations using LaTeX within math delimiters.

3. At the end, explicitly state the final answer alone, enclosed exactly within a single LaTeX math mode \verb|\boxed{}| tag, like this:

\verb|\boxed{final\ answer}|

4. Ensure that the content inside \verb|\boxed{}| exactly matches the true final answer in both value and format. Do not approximate unless explicitly required. Do not include any additional text, commentary, or explanation after the boxed final answer.

5. Make your entire response clear, complete, and logically structured, so that the reasoning and final answer can be easily followed and independently verified.

Strictly follow this format and instructions to maximize accuracy and clarity.

Example:

Step 1: ...
Step 2: ...
...
**Final answer:**

\verb|\boxed{...}|
</system\_prompt\_1>

\noindent\textbf{Group Prompts 3:}
<system\_prompt\_1>
You will receive a mathematical problem provided in the input field named 'problem'. Your task is to solve the problem completely and present the exact final answer.

Follow these precise instructions:

- Solve the problem thoroughly, showing a clear, logical, and fully justified step-by-step solution that leads directly from the problem statement to the final answer.
- Include all necessary intermediate computations, explanations, and reasoning in detail, ensuring each step is mathematically accurate and fully rigorous.
- Conclude your response by explicitly stating the final answer exactly once, enclosed in LaTeX \verb|\boxed{}| notation, in this format: \verb|\boxed{<answer>}|, where <answer> is the mathematically exact final result.
- The expression inside \verb|\boxed{}| must be exact: no approximations, rounding, or ambiguous forms are allowed.
- Do not include any extraneous or irrelevant information, commentary, or outputs other than the solution steps and the final boxed answer.
- Use clear, standard, and consistent mathematical notation and terminology throughout to maximize precision and comprehension.

Your answer will be evaluated based on exact mathematical correctness and clarity. Prioritize precision, thoroughness, and lucid exposition to ensure the final answer is completely accurate and can be fully trusted.
</system\_prompt\_1>

\end{tcolorbox}

\begin{tcolorbox}[
    colback=task5,
    colframe=task5!50!black,
    boxrule=0.3mm,
    sharp corners,
    width=\linewidth,
    fontupper=\small,
    title={10. GPQA (GPT-4.1-mini)},
    left=6pt, right=6pt, top=6pt, bottom=6pt,
    before upper=\vspace{2pt}\ttfamily\obeylines\obeyspaces,
    after upper=\vspace{2pt},
    breakable
]
\textbf{Baseline:}
<system\_prompt\_1>
Given the fields 'question', produce the fields 'answer'. 
</system\_prompt\_1>
\vspace{8pt}
\noindent\textbf{Alpha-Evolve:}
<system\_prompt\_1>
You will be provided with a multiple-choice question labeled 'question' along with answer options identified by uppercase letters (e.g., A, B, C, D). Your task is to carefully analyze the question and select exactly one correct answer from these options.

Instructions:

- Read the entire question and all answer options thoroughly before reasoning.
- Provide a detailed, precise, and step-by-step explanation that clearly and logically leads to the final answer.
- Support every step of your reasoning with relevant facts, principles, formulas, or evidence directly tied to the question; do not omit any critical reasoning steps.
- Avoid assumptions unless explicitly justified; ensure your reasoning fully supports the chosen answer.
- State your final answer exactly as a single uppercase letter corresponding to one of the given options.
- Enclose the final answer letter within a \verb|\boxed{}| tag, for example, \verb|\boxed{A}|.
- Provide only one definitive and confident final answer; do not present multiple possibilities or guesses.
- Your output should consist solely of your detailed reasoning followed immediately by the final boxed answer. Exclude any additional commentary, headings, or unrelated text.
- Before submitting, rigorously verify that your explanation and final answer are consistent and accurate to avoid mistakes.

Output Format:

[A comprehensive, step-by-step, logically sound explanation concluding with your final answer]

Final answer: \verb|\boxed{<letter>}|

where <letter> is the one selected option (A, B, C, or D).

Strict compliance with these instructions will maximize accuracy, clarity, and ensure your answer precisely matches the correct choice.
</system\_prompt\_1>
\vspace{8pt}
\noindent\textbf{C-Evolve:}
\noindent\textbf{Group Prompts 1:}
<system\_prompt\_1>
You will receive a multiple-choice question consisting of a question followed by answer options labeled with letters (e.g., A, B, C, D).

Your task is:

1. Carefully read and fully comprehend the entire question and all answer options.
2. Provide a thorough, clear, and step-by-step explanation that logically supports why the chosen answer is correct, explicitly addressing why the remaining options are incorrect. Include all relevant reasoning, calculations, or evidence to ensure complete accuracy.
3. After the explanation, output only the single final answer letter enclosed exactly within LaTeX \verb|\boxed{}| tags, with no extra characters, spaces, punctuation, or text outside the box. For example: \verb|\boxed{B}|

Strictly follow these requirements:

- Keep your response focused strictly on the step-by-step rationale and the final boxed answer; do not include any unrelated comments, disclaimers, or extraneous information.
- The final answer letter inside the \verb|\boxed{}| tag must exactly match the correct choice to maximize answer accuracy.
- Use concise, precise, and unambiguous language throughout your explanation to avoid any confusion.

Following this format ensures your answer is clear, correct, and perfectly aligned with the question demands.
</system\_prompt\_1>
\noindent\textbf{Group Prompts 2:}
<system\_prompt\_1>
You will receive a multiple-choice question labeled 'question' together with answer options labeled A, B, C, and D. Your objective is to determine the single correct answer by executing a rigorous, step-by-step reasoning process.

Instructions:

- Carefully read the entire question and all answer choices before starting the analysis.
- Provide a detailed, precise, and logically coherent explanation that systematically justifies the correct answer. Each reasoning step must be supported by relevant facts, calculations, or principles related to the question.
- Avoid making unstated assumptions and do not skip essential reasoning steps; make sure your explanation fully supports your final answer.
- Select exactly one final answer, corresponding to the uppercase letter of the provided options.
- Enclose your final answer letter in the \verb|\boxed{}| tag, for example, \verb|\boxed{A}|.
- Do not provide guesses, partial answers, or multiple answers; your output must conclusively identify the single correct choice.
- Your output should consist solely of the step-by-step explanation immediately followed by the final boxed answer. Avoid any extra commentary, headings, or unrelated content.
- Double-check the reasoning and final answer to ensure maximum accuracy and alignment with the question.

Output Format:

[Your detailed step-by-step reasoning leading to the final answer]

Final answer: \verb|\boxed{<letter>}|

where <letter> is the chosen option (A, B, C, or D).

Follow these instructions exactly to ensure clarity, precision, and correctness.
</system\_prompt\_1>
\noindent\textbf{Group Prompts 3:}
<system\_prompt\_1>
Given a multiple-choice question with answer options labeled by letters (A, B, C, D, etc.), your task is as follows:

1. Carefully read the entire question and all provided answer options to ensure complete understanding.

2. Provide a detailed, step-by-step logical explanation that definitively justifies why the selected answer is correct and all other options are incorrect. Base your reasoning on accurate facts, relevant calculations, precise definitions, or domain-specific knowledge directly pertinent to the question.

3. Conclude your response with **only** the single final answer letter exactly as shown in the options, enclosed in LaTeX \verb|\boxed{}| brackets, with no extra text, spaces, punctuation, or symbols. For example: \verb|\boxed{C}|.

Important instructions:

- Output solely your detailed answer and reasoning, followed immediately by the final boxed answer—no headings, preambles, or postambles.

- Avoid any vague language, filler words, generalities, or off-topic information.

- Ensure the final boxed answer exactly matches the correct option letter with zero deviation to maintain maximal accuracy.

Precise adherence to this format guarantees maximum clarity, rigor, and correctness in your responses.
</system\_prompt\_1>

\end{tcolorbox}


\end{document}